\documentclass{uclthesis}

\pdfoutput=1

\usepackage[english]{babel}       
\usepackage[utf8]{inputenc}       
\usepackage{hyperref}

\usepackage{makeidx}              
\usepackage[refpage]{nomencl}     
\usepackage{import}               

\usepackage{amsmath,amsthm,amssymb}

\usepackage{natbib}

\usepackage{url}            

\usepackage{multirow}
\usepackage{makecell}
\usepackage{ upgreek }
\usepackage{rotating}
\usepackage[]{algorithm}
\usepackage{algpseudocode}

\usepackage{graphicx}
\usepackage{subcaption}						
\usepackage{tabularx}
\usepackage{booktabs}       
\usepackage{standalone}						

\definecolor{light-gray}{gray}{0.85}

\theoremstyle{plain} 

\theoremstyle{definition}

\theoremstyle{remark}

\title{\textbf{Towards understanding deep learning with the natural clustering prior}}
\author{Simon Carbonnelle}
\institute{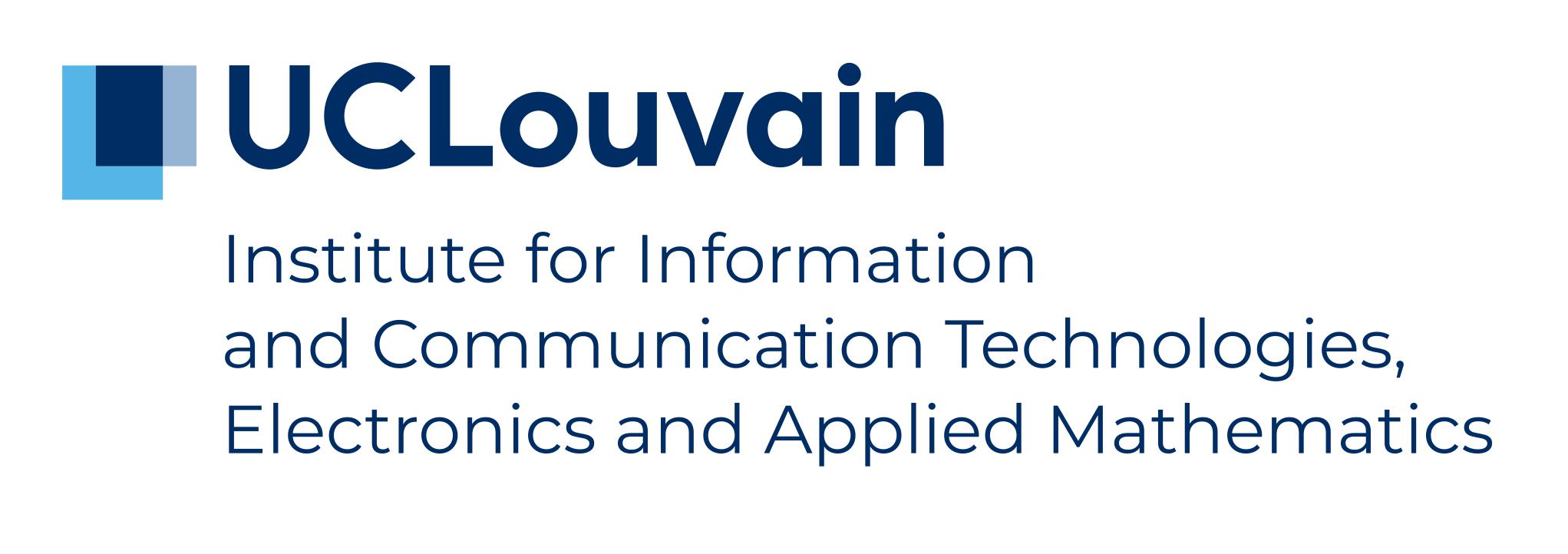} 
\grade{Doctor of Philosophy}
\date{January 2022}
\committee{Prof.\ Christophe De Vleeschouwer (UCLouvain, co-advisor)\\%
    Prof.\ Marie Van Reybroeck (UCLouvain, co-advisor)\\%
    Prof.\ Tinne Tuytelaars (K.U.Leuven)\\%
    Prof.\ Vincent François-Lavet (VU Amsterdam)\\%
    Prof.\ Benoît Macq (UCLouvain)\\%
    Prof.\ Laurent Jacques (UCLouvain)\\%
    Prof.\ David Bol (UCLouvain, chair)}

\begin{document}

\pagenumbering{gobble}
\maketitle

\renewcommand{\thechapter}{$\star$}
\renewcommand{\thesection}{$\star$}

\chapter*{Abstract}
The prior knowledge (a.k.a. \textit{priors}) integrated into the design of a machine learning system strongly influences its generalization abilities. In the specific context of deep learning, some of these priors are poorly understood as they implicitly emerge from the successful heuristics and tentative approximations of biological brains involved in deep learning design. Through the lens of supervised image classification problems, this thesis investigates the implicit integration of a natural clustering prior composed of three statements: (i) natural images exhibit a rich clustered structure, (ii) image classes are composed of multiple clusters and (iii) each cluster contains examples from a single class. The decomposition of classes into multiple clusters implies that supervised deep learning systems could benefit from \textit{unsupervised} clustering to define appropriate decision boundaries. Hence, this thesis attempts to identify implicit clustering abilities, mechanisms and hyperparameters in deep learning systems and evaluate their relevance for explaining the generalization abilities of these systems. 

Our study of implicit clustering abilities exploits hierarchical class labels to show that the subclasses (e.g., orchids, poppies, roses, sunflowers, tulips) associated to a class (e.g., flowers) are differentiated in deep neural networks that generalize well, even though only class-level supervision is provided. We then look for clustering mechanisms through the study of neuron-level training dynamics in multilayer perceptrons trained on a synthetic dataset with known clusters. Our experiments reveal a winner-take-most mechanism: training progressively increases the average pre-activation of the most activated clusters of a class and decreases the average pre-activation of the least activated clusters of the same class. Remarkably, this implicit mechanism leads neurons to differentiate some clusters from the same class more strongly than clusters from different classes. These studies indicate the emergence of a neuron-level training process that is critical for implicit clustering to occur. We propose to capture the extent by which the neurons of each layer have been effectively ``trained" during the global training process through the amount of layer rotation, i.e. the cosine distance between the initial and final flattened weight vectors of each layer. Equipped with tools to monitor and control the amount of layer rotation during training, we demonstrate that this implicit hyperparameter exhibits a consistent relationship with model generalization and training speed. Moreover, we show that the impact of layer rotation on training seems to explain the effect of several explicit hyperparameters such as the learning rate, weight decay, and the use of adaptive gradient methods.

Overall, our work thus provides a collection of experiments to support the relevance of the natural clustering prior for explaining generalization in deep learning. Additionally, it highlights the potential of using explicit clustering algorithms for training deep neural networks, as this would facilitate the integration of natural clustering-related priors into the design of deep learning systems.

\chapter*{Acknowledgements}
\dropcap{M}y deepest gratitude goes to all the people I've had the chance to live with during these six years. I don't think they know how much they have shaped my life -and still do. Thank you Sam, Ysa, Clarisse, Ol, Denis, Nouch, Oli, Bibou, Sixtine, Math, Ahmad, Faf, Delph, Steph, Radu, Gilles, Nico, Nina, Lucie, Clément, Théo, Hélène, Thomas, Arnould, Leia, Marie, Mathieu, Paloma, Fabrice, Lilas, Gregor and of course Claire, my love.\\
\\
I also want to thank my family, grandparents and friends for their constant support and loyalty despite my sometimes unconventional and confusing lifestyle.\\
\\
Thank you Christophe for your trust all along your supervision of my thesis. Thank you for the stimulating discussions and for being one of the first people with whom I dared to have an argument. Thank you for staying supportive albeit my rather unstable relationship to our shared project.\\
\\
Great thanks to all my colleagues for their precious humour, care, proofreading and support.\\ 
\\
Thanks to the reddit r/machinelearning community for helping me dive into the field of deep learning.\\
\\
Finally, I am grateful to the Université catholique de Louvain and the ICTEAM institute for providing the infrastructure necessary for this thesis. I am also grateful to the Fondation Louvain, the Université catholique de Louvain, the Fonds National de la Recherche Scientifique (F.R.S.-FNRS) of Belgium and the Walloon Region for funding this project.

\cleardoublepage  
\phantomsection 
\tableofcontents


\renewcommand{\thechapter}{\arabic{chapter}}
\renewcommand{\thesection}{\arabic{chapter}.\arabic{section}}

\chapter{Introduction and background}
\pagenumbering{arabic}
\dropcap{D}{eep learning} has lead to many technological breakthroughs since the 2010s. It has progressively substituted all other competing techniques for visual object recognition \citep{Krizhevsky}, natural  language processing \citep{Young2018}, speech recognition \citep{Graves2013b}, playing board and video games \citep{Silver2016a}, protein structure prediction \citep{Jumper2021} and many others. It is integrated in a myriad of modern applications like social network platforms, e-commerce and smartphone cameras \citep{LeCun2015}.

As the practical applications of deep learning keep flourishing, the realization that we do not really understand why and how deep learning works is growing. Renown researchers associate deep learning to ``alchemy", as current practice depends more on beliefs and intuitions than on well-established scientific facts \citep{Rahimi2017}. Specialized conference workshops are organized to make sense of an increasingly large body of observations that escape our understanding (e.g., ``Identifying and understanding deep learning phenomena" workshop organized during ICML 2019). Developing mathematical theories of deep learning has become an increasingly active area of research \citep{Arora2018, Berner2021}.

Making progress on these puzzles has the potential to facilitate the design of deep learning-based systems and widen their range of applications (e.g., safety-critical applications). Moreover, deep learning has always been tightly connected to neuroscience and biological brains \citep{Schmidhuber2014,Wang2017,Hassabis2017}. Hence, a better understanding of deep learning has the potential to bring new insights to a long-standing quest in human history: understanding our own minds. 

In order to dive into this fascinating field, we will start by introducing the basics of deep learning and machine learning. The following section provides a brief and non-technical introduction to these topics tailored for this specific thesis. Many text books are available for the readers looking for a more exhaustive overview (e.g., \citet{Bishop2006, Murphy2021, Goodfellow2016}). 

\section{Deep learning basics}
Deep learning is part of the broader field of machine learning. Machine learning is a class of techniques used to estimate an unknown function $f^*$ mapping inputs $\mathbf{x}$ to outputs $\mathbf{y}$. In the context of image classification, which is the main focus of this work, $f^*$ maps an image $\mathbf{x}$ to a class $\mathbf{y}$ (also called label or category) reflecting the image's content (e.g., ``dog", ``car" or ``house").

In order to estimate the function $f^*$, the key specificity of machine learning is to make use of knowledge contained in data. This approach reduces the need for knowledge from human experts, which is particularly useful when human expertise is costly or difficult to formalize (e.g., subjective, intuitive or unconscious expertise). Before discovering how machine learning techniques extract knowledge from data in Section \ref{sec:learning}, let's clarify what data means in the context of deep learning.

\subsection{Natural data}
While deep learning could be applied on any type of data in principle, its popularity is mostly due to its performance on natural images, sounds and language. In these cases, deep learning differs from alternative machine learning techniques by working directly on raw data, i.e. with minimal pre-processing \citep{LeCun2015}. As an example, in the context of image classification, the input to a deep learning system are typically images as represented by their pixel values. For an RGB image of size $m\times n$, we have $\mathbf{x} \in \mathbb{R}^{m\times n\times 3}$. In comparison, alternative techniques require human engineered pre-processing algorithms such as Histogram of Gradients \citep{Dalal2005} or Scale Invariant Feature Tansforms \citep{Lowe1999}.

The scenario by which data are available for a deep learning system can also vary. \textit{Supervised learning} refers to the scenario where inputs $\mathbf{x}$ are provided with their associated outputs $\mathbf{y}$. \textit{Unsupervised learning} refers to the scenario where only inputs are available\footnote{\textit{Reinforcement} and \textit{self-supervised learning} are two other scenarios that require additional formalisms which we do not introduce here.}. We also distinguish \textit{static data} that takes the form of a fixed dataset of $S$ examples $(\mathbf{x}_i,\mathbf{y}_i)$ for $i\in \left\lbrace 1,2,...,S \right\rbrace$ from a \textit{stream of data} where examples are provided sequentially during the machine learning process. This second scenario is often denoted by \textit{continuous learning}.

\begin{figure}[h]
\begin{center}
\includegraphics[width=0.9\linewidth]{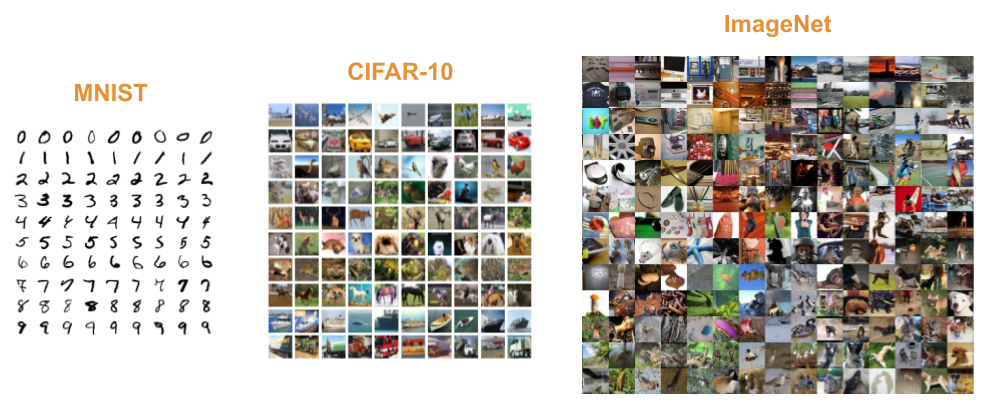}
\end{center}
\caption{Examples from three standard datasets used for deep learning research\protect\footnotemark. The images are not at scale.}
\label{fig:datasets}
\end{figure}

\textbf{This thesis focuses on supervised learning applied to image classification using static datasets.} This problem setting has been extensively used for deep learning research. In particular, four image classification datasets became the de facto standard for studying deep learning techniques: MNIST \citep{LeCun1998}, CIFAR10, CIFAR100 \citep{Krizhevsky2009} and ImageNet \citep{Deng2009}. Each of them contains more than $50.000$ images with their associated class. Visualizations and specifications of these four datasets are presented in Figure \ref{fig:datasets} and Table \ref{tab:datasets} respectively.

\footnotetext{The three dataset visualizations are taken from \url{https://en.wikipedia.org/wiki/MNIST_database}, \url{https://www.cs.toronto.edu/~kriz/cifar.html} and \url{https://cs.stanford.edu/people/karpathy/cnnembed/} respectively.}

\begin{table*}[!h]
  \caption{Specifications of four standard datasets used for deep learning research.}
  \label{tab:datasets}
  \centering
  \begin{tabular}{llll}
    \toprule
    \textbf{Dataset}     & Image size     & $\#$ of samples & $\#$ of classes \\
    \midrule
    MNIST    & $28\times 28$  & $70.000$ &  $10$  \\
    CIFAR-10 & $32\times 32 \times 3$  & $60.000$   &  $10$ \\
    CIFAR-100  & $32\times 32 \times 3$    & $60.000$ &  $100$\\
    ImageNet   & e.g., $200\times 200 \times 3$ & $>1.000.000$  & $1000$\\
    \bottomrule
  \end{tabular}
\end{table*}


\subsection{Learning from data} \label{sec:learning}
In order to extract knowledge from data, machine learning techniques need two ingredients: a hypothesis class and a training algorithm. The \textit{hypothesis class} is the set of functions that are considered as potential estimates of $f^*$. The \textit{training algorithm} is a data-driven procedure to select one estimate $\hat{f}$ from the hypothesis class. In the case of deep learning, deep neural networks (DNNs) constitute the hypothesis class and stochastic gradient descent (SGD) the most standard training algorithm. This section briefly describes these two key components.

\subsubsection{Deep neural networks}
\begin{figure}[h]
\begin{center}
\includegraphics[width=0.6\linewidth]{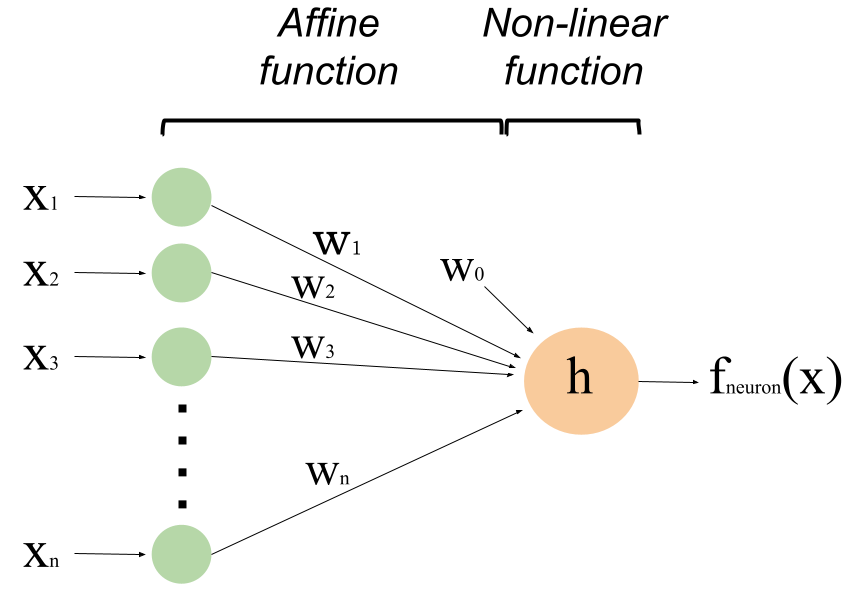}
\end{center}
\caption{Graphical representation of an artificial neuron. The weights $w_0,w_1,...,w_n$ are to be determined by a training algorithm.}
\label{fig:artificialNeuron}
\end{figure}

The fundamental building block of deep neural networks are artificial neurons. The standard artificial neuron corresponds to the composition of an affine function and a non-linear function, as represented graphically in Figure \ref{fig:artificialNeuron}. Mathematically, this corresponds to
$$ f_{\text{neuron}}\left(\mathbf{x}\right) = h\left(w_0 + \sum _{i=1}^n w_i x_i \right)$$
where $x_i$ represents the $i^{th}$ element of the input $\mathbf{x}$, $w_0, w_1, ..., w_n$ are the affine function's parameters and $h$ represents the non-linear or \textit{activation function}. As of today, the most common activation function is the rectified linear unit or ReLU \citep{Nair2010}:
$$h(x) = \text{max}(0,x).$$
The parameters $w_0, w_1, ..., w_n$ (also denoted by \textit{weights}) are unspecified, such that any parameter instantiation produces a function that is part of the hypothesis class. It is the role of the training algorithm to determine the weights to be used to estimate $f^*$. %

\begin{figure}[h]
\begin{center}
\includegraphics[width=0.75\linewidth]{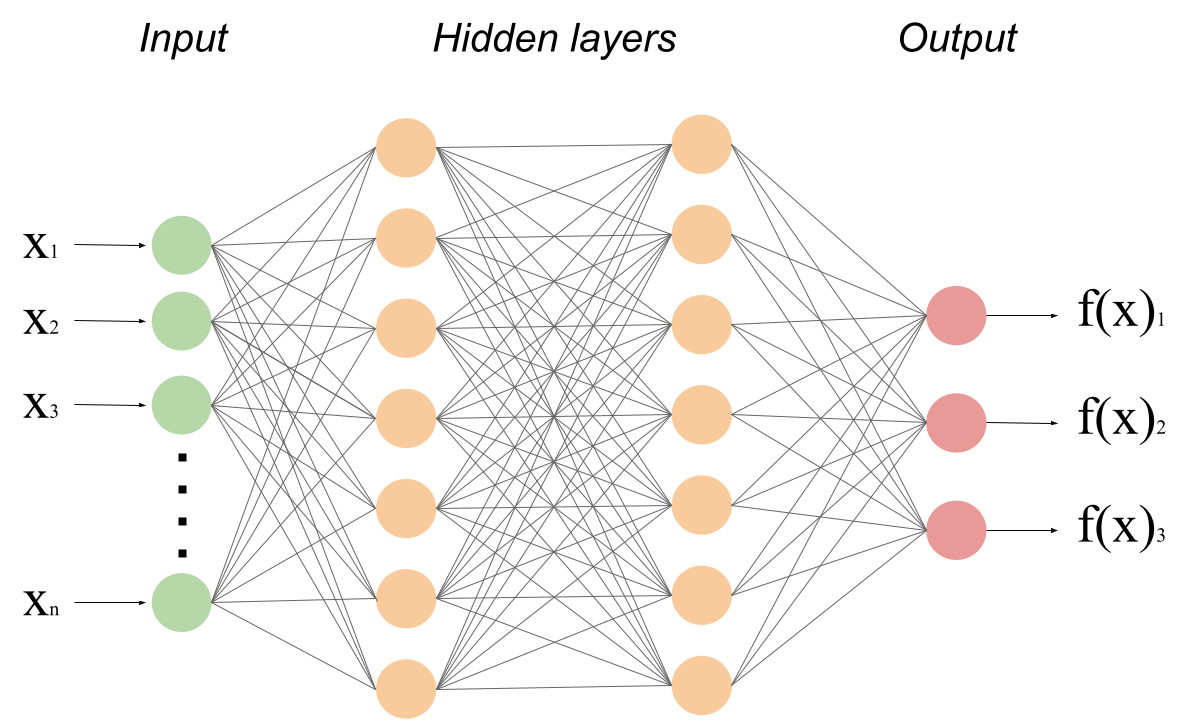}
\end{center}
\caption{Graphical representation of a MLP neural network with two hidden layers.}
\label{fig:neuralNetwork}
\end{figure}

In order to estimate complex functions, neural networks can be built by combining and connecting multiple artificial neurons. The most conceptually simple neural network is the \textit{multilayer perceptron} (MLP) represented in Figure \ref{fig:neuralNetwork}. In this network, the neurons are organized in layers, where the output of one layer becomes the input of the next. Such layered structure is a key aspect of deep neural networks, where \textit{deep} refers to the relatively many layers they contain.

In a multilayer perceptron, each neuron is connected to all the inputs of its layer (cfr. Figure \ref{fig:neuralNetwork}). We call such layers \textit{fully connected layers}. In the context of images, another connectivity pattern has been very successful: the \textit{convolutional layer} \citep{LeCun1998}. Here, the affine function becomes a convolution operation, applied on the spatial dimensions of the inputs. Each neuron is thus connected to a local neighbourhood of its layer's inputs, akin the local receptive fields of visual cortices \citep{Hubel1962}. In addition, the same affine transformation is applied on each neighbourhood, such that multiple neurons will share the same parameters $w_0,w_1,...w_n$. Neural networks that contain such layers are commonly called \textit{convolutional neural networks} (CNNs). 

Many other types of layers have been proposed besides fully connected and convolutional layers. In the context of this thesis, three other layers are regularly used: batch normalization \citep{Ioffe2015}, pooling \citep{LeCun1998} and softmax layers. Batch normalization layers are typically inserted between the affine and activation functions of a network. They normalize the (pre-)activations of each neuron to have zero mean and unit variance, based on the statistics of a subset of the entire dataset (a \textit{batch}). Pooling layers are applied between convolutional layers to reduce the spatial dimensions of a signal. They do so by aggregating the values of neighbouring pixels (typically $2\times2$ patches) through mean or max operations. Finally, softmax layers are applied at the output of the network to identify the predicted class. It is used as a differentiable alternative to the one-hot argmax operation. For the interested readers, we refer to the original papers and \citep{Goodfellow2016} for a more extensive description of all these layers.

The power of deep neural networks largely comes from their modular structure which enables a flexible and adaptative design from relatively simple components such as layers. With the years, specific design choices or \textit{architectures} gained popularity, amongst which VGG \citep{Simonyan2014}, ResNets \citep{He2016} and Wide ResNet \citep{Zagoruyko2016}. These three architectures will be regularly used in the context of this thesis.

\subsubsection{Stochastic gradient descent}
Once a deep neural network has been designed, its parameters or weights still need to be determined by the training algorithm. The optimal parameters are those that minimize the estimation error. But, because we don't have access to the function $f^*$ to be estimated, we need to use a proxy of the estimation error instead: the \textit{loss function}. The information we have about $f^*$ takes the form of data. Hence, the loss function is data-driven and typically returns large values when an estimate $\hat{f}$ does not match $f^*$ on the available data and small values when it does. In the context of image classification, the most common loss function is \textit{categorical cross-entropy} \citep{Goodfellow2016}. 

The optimization algorithm used by deep learning to minimize a loss function is stochastic gradient descent \citep{Goodfellow2016}. This method is especially compelling since (i) deep neural networks and categorical cross-entropy are differentiable almost everywhere w.r.t. the weights, (ii) the backpropagation algorithm provides an efficient way to compute the gradient \citep{Linnainmaa1970,Werbos1982,Rumelhart1986} and (iii) the loss can be approximated by a random subset (also called \textit{batch}) of data. Let $\mathcal{L}(\mathbf{x}_\text{batch},\mathbf{y}_\text{batch})$ be the average loss of a random batch of data containing $N$ samples ($N$ is also called the \textit{batch size}). Stochastic gradient descent iteratively updates each weight $w_i$ according to the following rule:
$$ w_i^{t+1} = w_i^{t} - \lambda^t \frac{\partial \mathcal{L}(\mathbf{x}_\text{batch},\mathbf{y}_\text{batch})}{\partial w_i}, $$
where $\lambda$ is the \textit{learning rate}, which is a parameter that typically evolves during training according to a pre-determined schedule. Batches are randomly sampled from the dataset without replacement. We call an \textit{epoch} the number of iterations required for all samples to be considered. A single epoch usually doesn't suffice for convergence of the algorithm, and the whole dataset is considered again after each epoch.

Despite the non-convexity of the loss function, it is empirically observed that stochastic gradient descent, provided appropriate tuning of its learning rate parameter, often converges to a global minimum of the loss function \citep{Du2019}. Hence, we are able to determine the weights of a deep neural network such that it matches the true function $f^*$ \textit{on the data used by the loss function}. But what about data that isn't considered by it?

\subsection{Generalizing to unseen data} \label{sec:generalization}
Intuitively, since the optimization of the weights targets performance on a single dataset, there's a risk that performance decreases when the model is applied on other data. The ultimate goal of machine learning techniques is to provide an estimate of $f^*$ that is also accurate on data not considered by the training algorithm. This ability is called \textit{generalization}. It is usually measured by computing the loss (or any another measure of error) on a different set of examples (the \textit{test set}) that was created independently using the same data generation process as the data used for training (the \textit{training set}). 

In addition to this empirical measurement of generalization ability, providing frameworks to predict or reason about generalization has been an important research endeavour. The most successful frameworks involve a balance between some notion of \textit{capacity} (also denoted by complexity, expressive power, richness, or flexibility) associated to the hypothesis class and the size of the training set. Informally, the capacity of a hypothesis class reflects the diversity of functions it contains. The larger the capacity, the higher the chance that the hypothesis class contains good approximations of $f^*$. However, it also augments the chance of containing functions that generalize poorly, i.e. that provide good approximations of $f^*$ \textit{on the training set only}. This risk gets mitigated by increasing the size of the training set, as the latter then becomes more representative of the data generation process.

\begin{figure}[h]
\begin{center}
\includegraphics[width=0.75\linewidth]{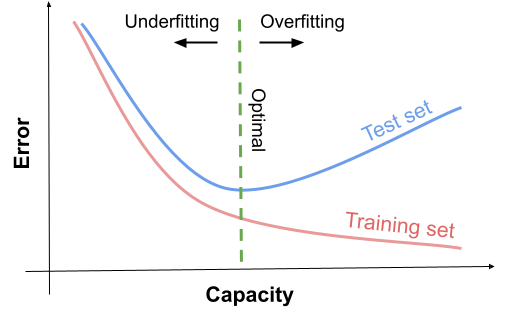}
\end{center}
\caption{Illustration of the bias-variance trade-off, a commonly adopted heuristic that, for a given training set size, postulates the existence of an optimal middle ground between too low a capacity (denoted by \textit{underfitting}) and too high a capacity (denoted by \textit{overfitting}).}
\label{fig:biasVarianceTradeoff}
\end{figure}

These intuitions have lead to the \textit{bias-variance trade-off} (cfr. Figure \ref{fig:biasVarianceTradeoff}). This is a commonly adopted heuristic that, for a given training set size, postulates the existence of an optimal middle ground between too low a capacity (denoted by \textit{underfitting}) and too high a capacity (denoted by \textit{overfitting}) \citep{Geman1992}. A more rigorous formalization of these intuitions is provided by Vapnik–Chervonenkis theory, which balances \textit{VC dimensions} (which is a measure of capacity) with the size of the training set to bound the difference between training and test errors (which reflects generalization ability) \citep{Vapnik1968, Vapnik1989}.

Capacity-based reasoning can also be useful to think about the role of training algorithms in generalization. Indeed, even if a hypothesis class has a large capacity (i.e. can represent a lot of different functions), the training algorithm doesn't necessarily search through all functions uniformly. In particular, the algorithm can be designed to favour certain types of functions, which are expected to generalize better. Aspects of the training algorithm which aim to improve generalization are commonly denoted by \textit{regularization}. The most classical example is $L_2$ regularization, which penalizes functions whose parameters have a large Euclidean norm.

\section{The generalization puzzles of deep learning} \label{sec:puzzles}
Even though generalization constitutes the ultimate purpose of machine learning systems, it largely escapes our understanding in the case of deep learning. The mystery is two-fold. First, deep neural networks generalize remarkably well from the perspective of classical theoretical frameworks and conventional wisdom (cfr. Section \ref{sec:generalization}). Second deep neural networks generalize remarkably poorly compared to us, humans. This section dives deeper into these two open questions.

\subsection{Why do deep neural networks generalize so well?} \label{sec:GoodGeneralization}
A remarkable aspect of modern deep neural networks is their gigantic size. State of the art models can contain hundreds of layers and millions of parameters \citep{He2016,Zagoruyko2016}. This implies that the capacity of the hypothesis classes used for deep learning are extremely large. A typical trend in classical theories and heuristics is that large capacity involves the risk of overfitting (cfr. Section \ref{sec:generalization}). Two pioneering works have shown that deep neural networks mysteriously mitigate this risk.

\begin{figure}[h]
\begin{center}
\includegraphics[width=0.55\linewidth]{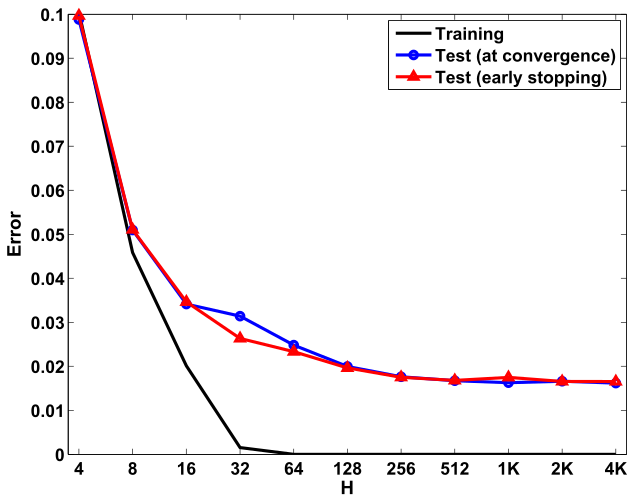}
\end{center}
\caption{This experimental result from \citet{Neyshabur2015} shows that increasing the amount of neurons $H$ in a one-hidden-layer neural network trained on MNIST does not seem to lead to an increase in the test error, even if perfect performance on the training set is already achieved. This contradicts the commonly held belief that increasing capacity should ultimately lead to overfitting (cfr. Figure \ref{fig:biasVarianceTradeoff}).}
\label{fig:biasVarianceTradeoffMNIST}
\end{figure}

First, \citet{Neyshabur2015} observed that generalization ability \textit{improves} when increasing the amount of neurons (and thus the capacity) of single-hidden-layer neural networks, even beyond what is needed to achieve zero training error (cfr. Figure \ref{fig:biasVarianceTradeoffMNIST}).  This contradicts the bias-variance trade-off, which states that increasing capacity should ultimately lead to overfitting (cfr. Figure \ref{fig:biasVarianceTradeoff}). Second, \citet{Zhang2017} observed that state of the art networks reached perfect training error even when the class labels of their training set are randomized. This implies an ability to memorize each example of the training set, and thus an hypothesis class large enough to contain many functions that cannot generalize at all. Both works thus show that \textit{the generalization abilities of deep neural network do not seem to be affected by their enormous capacity}. On the contrary, increasing the capacity of deep neural networks tends to benefit generalization and is a key component of state of the art models.

The large capacity of deep neural networks' hypothesis classes must thus be compensated by strong regularization mechanisms that steer the training algorithm towards functions that generalize well. However, both works show that their observations hold even in the absence of classical regularization techniques. In order to make sense of their experimental results, \citet{Neyshabur2015} and \citet{Zhang2017} thus conjecture the existence of \textit{an implicit form of regularization} originating from stochastic gradient descent. The characterization of this implicit regularization mechanism has become a very active, yet unsolved area of research (e.g., \citet{Zhang2021,Wu2021,Smith2021Implicit,Gradient2021,Yun2021}).


\subsection{Why do deep neural networks generalize so poorly?} \label{sec:poorGeneralization}
Deep learning is often considered as a potential candidate for human-level artificial intelligence. Hence, it makes sense to compare the performance of deep neural networks to humans. While deep neural networks can achieve super-human performance on specific datasets \citep{He2015}, their generalization ability appears to be much worse.

A first line of work demonstrated that fooling deep neural networks into wrong and yet confident image classifications was relatively easy in an adversarial setting. \citet{Szegedy2014,Goodfellow2015a} fool networks by adding small perturbations to the inputs that are invisible to the human eye, \citet{Su2019} by changing the value of a single pixel and \citet{Nguyen2015} by generating images from scratch that are unrecognizable to humans.

\begin{figure}[t]
\begin{minipage}[b]{.3\linewidth}
  \centering
  \centerline{\includegraphics[width=3.8cm]{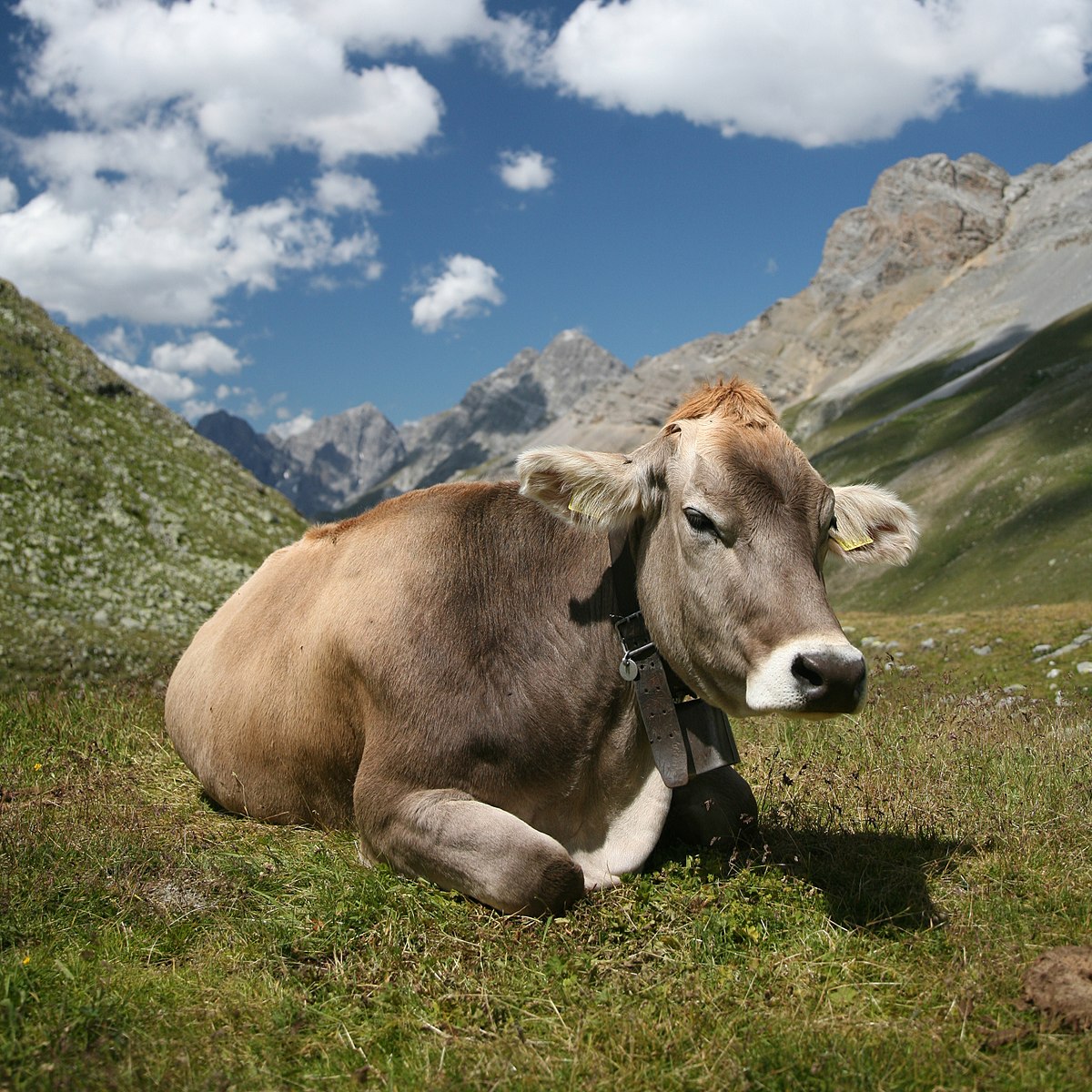}}
{\scriptsize (A) {\bf Cow: 0.99},
Pasture: 0.99,
Grass: 0.99,
No Person: 0.98,
Mammal: 0.98}

  \medskip
\end{minipage}
\hfill
\begin{minipage}[b]{0.3\linewidth}
  \centering
  \centerline{\includegraphics[width=3.8cm]{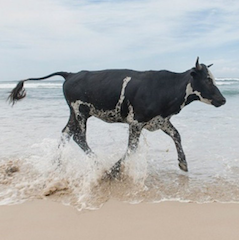}}
{\scriptsize (B) No Person: 0.99,
Water: 0.98,
Beach: 0.97,
Outdoors: 0.97,
Seashore: 0.97}

  \medskip
\end{minipage}
\hfill
\begin{minipage}[b]{0.3\linewidth}
  \centering
  \centerline{\includegraphics[width=3.8cm]{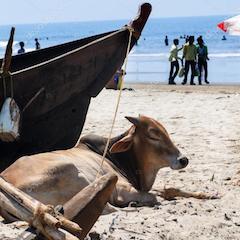}}
{\scriptsize (C) No Person: 0.97,
{\bf Mammal: 0.96},
Water: 0.94,
Beach: 0.94,
Two: 0.94}
  \medskip
\end{minipage}
\vspace{-0.15in}
\caption{This result taken from \citet{Beery2018} illustrates the poor generalization abilities of deep neural networks compared to humans. For different images of cows, the top five classes and confidence produced by a deep learning system are shown. We observe that the quality of the predictions heavily depends on the background. In particular, the cow is better recognized in a "common" background (Alpine pastures) than in unusual ones which are probably poorly represented in the training set (e.g., seashore).}
\label{fig:BadGeneralization}
\end{figure}

While in the adversarial setting data are manipulated artificially, a large body of work has shown that natural changes to the data can also dramatically affect a deep neural network's performance. \citet{Torralba2011} showed that deep neural networks do not generalize well from one image classification dataset to the other, and \citet{Recht2019,Shankar2020} observed the same behaviour even when extra care is taken to replicate the data generation process. Deep neural networks have also been shown to lack robustness to changes in the background (cfr. Figure \ref{fig:BadGeneralization}) \citep{Beery2018}, object pose \citep{Alcorn2019} or texture \citep{Geirhos2018}. Their performance also worsens when small rotations and translations are applied to the image \citep{Engstrom2019} as well as corruptions and distortions \citep{Dodge2017,Geirhos2018a,Hendrycks2019}.

Overall, there is a growing consensus that deep neural networks are very far from human-level understanding of natural data. Spurious correlations only occurring in specific datasets seem to play a crucial role in their decisions, leading to a lack of robustness to adversarial and natural changes to the data. Overcoming this crucial limitation of deep learning has become a very active and yet unsolved area of research (e.g., \citet{Arjovsky2021,Gulrajani2021,Krueger2021,Nagarajan2021}). 



\section{Solving the puzzles through the study of priors}
While machine learning leverages data to be less reliant on human expertise, the latter still plays a crucial role. In particular, machine learning practitioners specify the hypothesis class and the training algorithm, which heavily influence a machine learning system's generalization ability in practice. The practitioners' choices are typically based on some a priori knowledge they possess about the function $f^*$ to be estimated (a.k.a. \textit{priors}). This section explores the role of priors in generalization and in deep learning\footnote{The notion of prior is very related to the notion of inductive bias. We use priors as a characteristic of the problem, describing its inherent structure. An inductive bias is a characteristic of the machine learning system, describing the assumptions it makes on the problems it will be applied on. Generally, one wants the inductive biases to correspond to priors that were effectively integrated into the machine learning system's design. Hence, priors and inductive biases are often two sides of the same coin.}.

\subsection{On the role of priors in generalization}
The \textit{No Free Lunch theorem} (NFL) states that all machine learning systems (even a completely random system that does not depend on data) are equivalent in terms of generalization ability in the absence of assumptions or priors on the problem to be solved \citep{Wolpert1996, Schaffer1994}. This suggests that the priors integrated in a system are key for its performance in a specific problem setting. Intuitively, the more an algorithm integrates relevant knowledge from its designers, the less training data it requires for generalizing well. In particular, priors can lead to hypothesis classes and training algorithms which consider a more restricted set of functions while still including good estimations of the target function $f^*$.  

Including the role of priors in a general learning theory requires a formalism to represent priors and their relationship with learning problems and algorithms. The bayesian learning framework goes into this direction by expressing priors through the language of probability theory and making their role in a learning system more explicit through the use of Bayes' rule. Another more recent effort formalizes the role of priors by incorporating ``Teachers" in machine learning systems in addition to data, hypothesis classes and training algorithms \citep{Vapnik2019a}. However, these lines of work did not yet lead to theorems connecting priors and generalization in a useful and practical way. In the absence of a general theory, one can build theories for specific problem settings. In this context, assumptions concerning the relevance of priors can be made (and tested empirically). A growing body of work argues that studying the priors integrated into deep learning systems specifically is key to solve the generalization puzzles we described in Section \ref{sec:puzzles} (e.g., \citet{Arpit2017,Kawaguchi2017,Dauber2020}). But even then, producing a theory of deep learning remains a challenge. Indeed, the priors involved in deep learning appear to be quite difficult to determine and formalize.

\subsection{The difficult case of deep learning priors}
Modern deep learning is the result of a relatively long and tedious endeavour. Its development started more than 60 years ago and gathered variable amounts of popularity over time (cfr. the \textit{AI winters}). Throughout the process, biological brains have been an important source of inspiration \citep{Hassabis2017}. From the mathematical formulation of artificial neurons \citep{McCulloch1943} to their learnability \citep{Hebb1949a,Rosenblatt1958,Widrow1960}, to convolutional connectivity patterns \citep{Fukushima1980,LeCun1998} and attention mechanisms \citep{Mnih2014}, many foundational ideas of deep learning are inspired from biological brains. The origin of deep learning's most successful training algorithm (SGD) provides an exception. In contrast to many alternative training algorithms (e.g., \citet{Hebb1949a,Rosenblatt1958}), SGD is not inspired from biological brains, but is rather a very general mathematical tool whose use in deep learning stems mostly from a trick that makes it computationally efficient (the backpropagation algorithm, cfr. \citet{Linnainmaa1970,Werbos1982}). SGD's popularity greatly increased when empirical work suggested that it was capable of learning important intermediary features automatically \citep{Rumelhart1986}. But how this capability emerged from SGD was not explained. Given its empirical successes, several works attempt to discover how biological brains could in fact implement backpropagation-like algorithms after all \citep{Bengio2015,Lillicrap2020}.

While the above paragraph summarizes a long history in a few sentences (we refer to \citet{Schmidhuber2014,Wang2017,Lecun2019} for more exhaustive historical perspectives), it reveals that crucial ideas behind deep learning originate from studies of biological brains and trial and error. Since they do not stem from an understanding of natural data-related problems, they do not provide insights about the priors deep learning takes advantage of. We have little to no clue as to why deep neural networks and SGD are appropriate choices for natural data problems. Several works provide attempts to characterize the priors of deep learning. Today's most popular priors are the need for distributed representations with multiple levels of abstraction (e.g., \citet{Rumelhart1986,Hinton1987,Bengio2009,Bengio2012,LeCun2015}). These priors remain intuitive and are difficult to use in practice to solve deep learning's puzzles. For example, we are not aware of any formal way to measure the extent by which a deep neural network's representations are distributed or contain abstraction. Overall, the characterization of deep learning priors is thus far from established and complementary/alternative priors could play a critical role. 

\subsection{The natural clustering prior} \label{sec:naturalClusteringPrior}
The natural clustering prior states that natural image datasets exhibit a rich clustered structure. This means that natural images can be partitioned into different groups (or clusters) such that images inside a group are more similar to each other (according to some metric) than to images from other groups. While this remains a very high-level and quite general description, additional statements can be associated to the natural clustering prior which describe the shape of clusters, their relative density, the distance between them or their relationship with class labels. The more precision we can achieve, the more helpful the prior will be. Previous work added the statement that samples from different classes do not belong to the same cluster \citep{Chapelle2005,Bengio2012}. Hence a cluster always contains samples from one unique class. \textbf{In this thesis, we further state that there are many more clusters than classes in standard image classification datasets. This implies that a single class is divided into multiple distinct clusters, which we denote by \textit{intraclass clusters}.}

\begin{figure}[h]
\begin{center}
\includegraphics[width=0.55\linewidth]{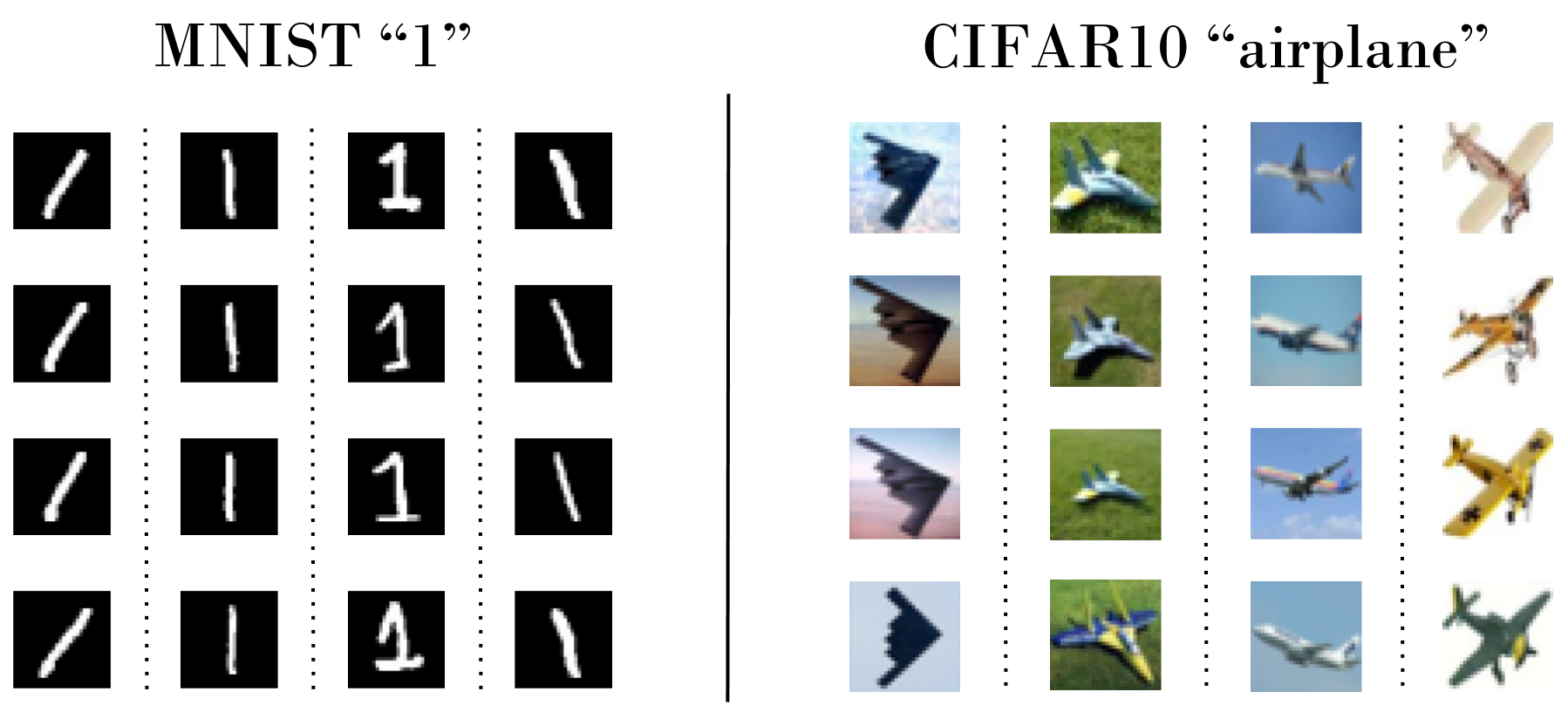}
\end{center}
\caption{In standard image classification datasets, a single class can often be decomposed in multiple groups of similarly looking images, which we interpret as \textit{intraclass} clusters. The occurrence of multiple clusters inside a class constitutes one of the statements of the natural clustering prior investigated in this thesis.}
\label{fig:intraclass_clusters}
\end{figure}

Figure \ref{fig:intraclass_clusters} provides a motivation for this prior by identifying intraclass clusters in standard image classification datasets. Another argument arises from the hierarchical structure of many class labellings. For example, CIFAR100 contains $20$ superclasses (e.g., flowers) which are further divided into $100$ subclasses (e.g., orchids, poppies, roses, sunflowers, tulips). The ImageNet class labels are also hierarchically organized with up to $6$ levels of abstraction (e.g., digital clock $\rightarrow$ clock $\rightarrow$ timepiece $\rightarrow$ measuring instrument $\rightarrow$ device $\rightarrow$ artifact). The fact that class labels can be decomposed in multiple subclasses suggests that the associated data can be grouped into multiple intraclass clusters.

\begin{figure}[h]
\begin{center}
\begin{subfigure}[h]{0.35\linewidth}
\includegraphics[width=1.\linewidth]{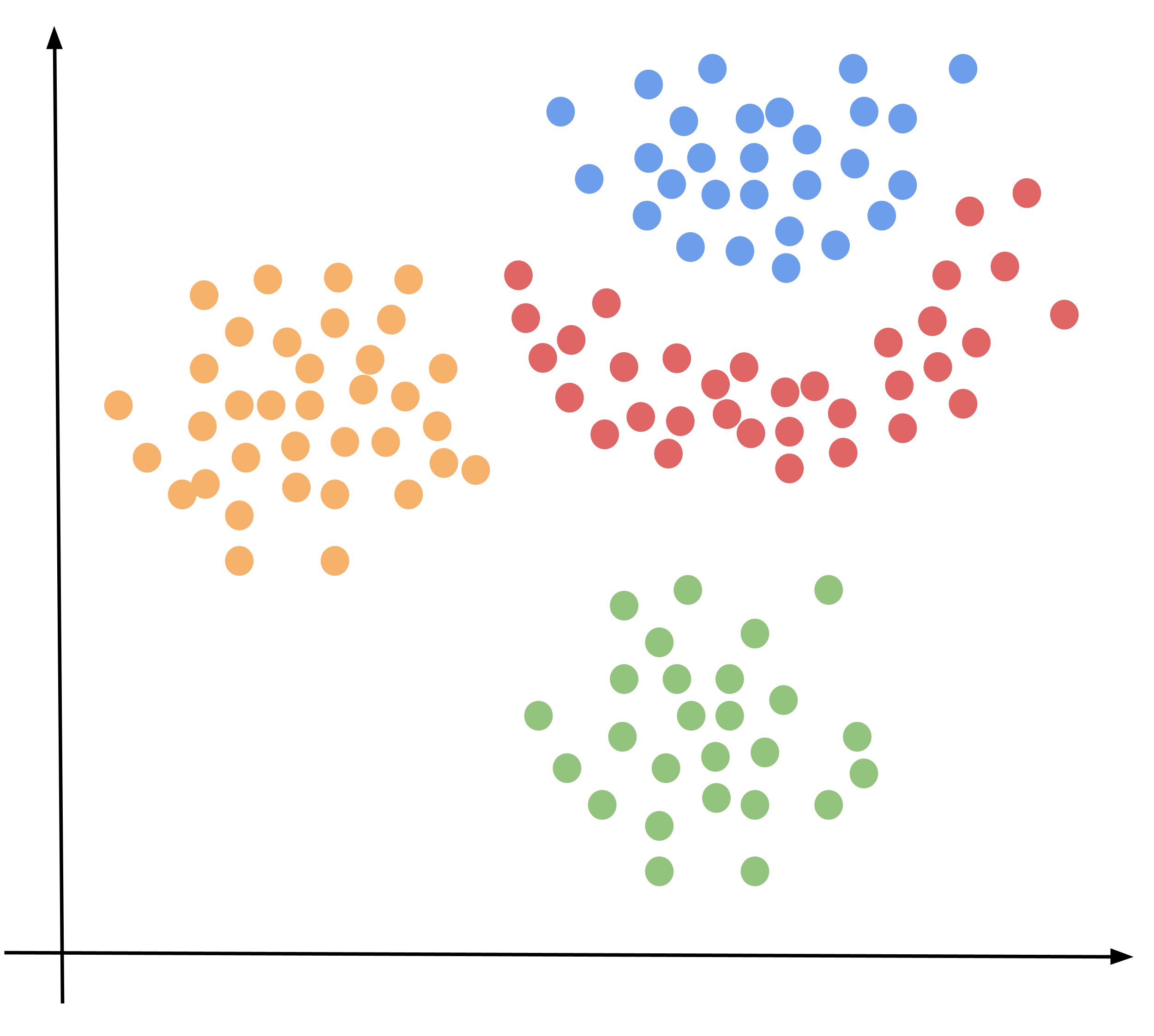}
\caption{}
\end{subfigure}
\begin{subfigure}[h]{0.35\linewidth}
\includegraphics[width=1.\linewidth]{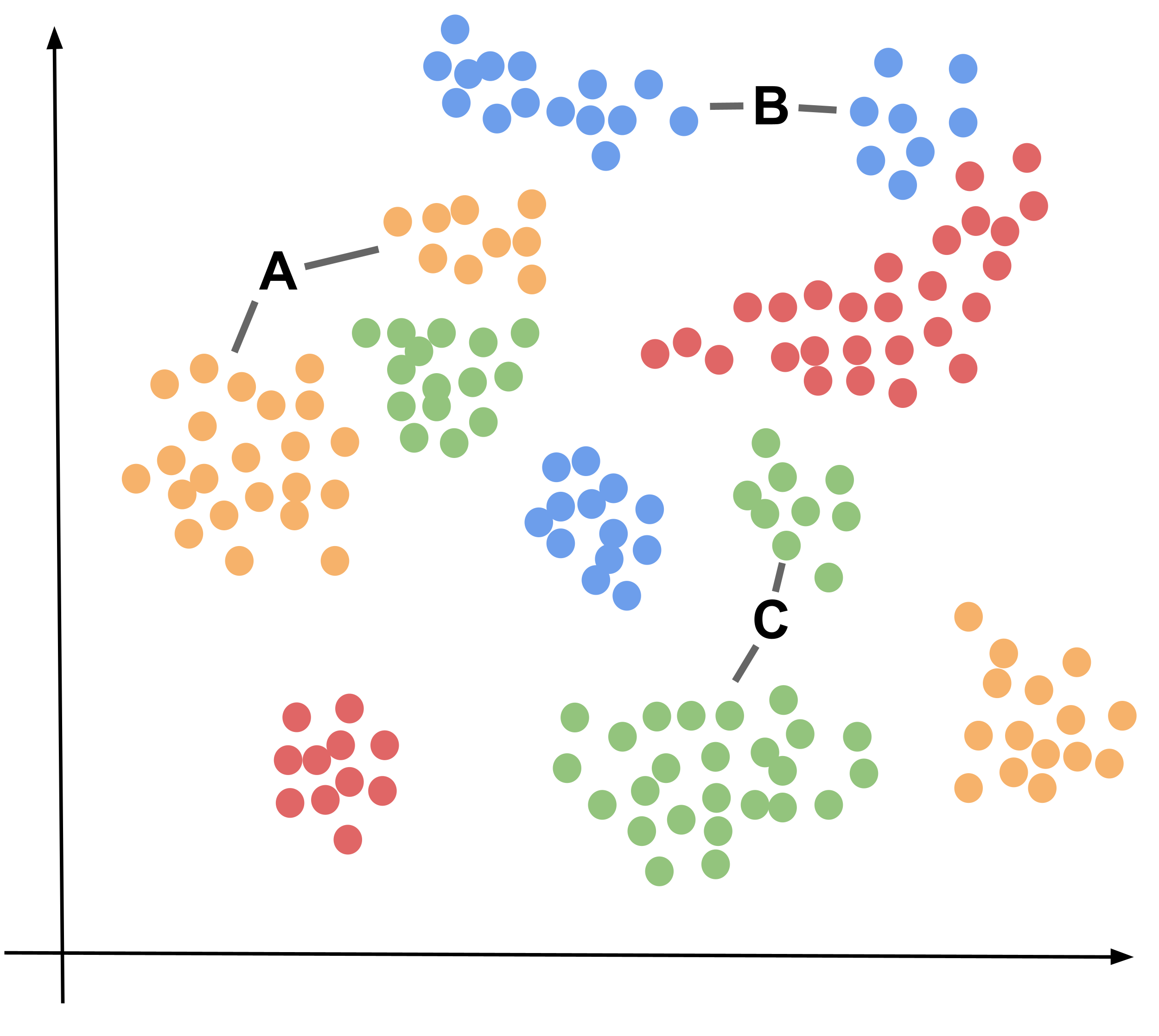}
\caption{}
\end{subfigure}
\end{center}
\caption{A simplified two-dimensional representation of the natural clustering prior as formulated by this thesis. Points refer to training examples, colors to their associated class. The natural clustering prior states that (i) data is structured into clusters, (ii) clusters contain examples from a single class (like in (a) and (b)) and (iii) classes are composed of multiple clusters (like in (b)). The latter implies that supervised image classifiers would benefit from unsupervised clustering abilities which make appropriate assumptions on the clusters' characteristics. For example, A, B and C in figure (b) constitute ambiguous data structures which could be interpreted as one unique or two distinct clusters by supervised classifiers. Each interpretation leads to different decision boundaries, and thus different generalization abilities. In this thesis, we examine whether similar phenomena occur in much higher dimensions when deep learning is applied on natural data.}
\label{fig:natural_clustering_prior}
\end{figure}

The presence of intraclass clusters implies that supervised image classifiers would benefit from unsupervised clustering and appropriate assumptions on the clusters' characteristics. Indeed, whether supervised classifiers interpret a set of data as one unique or two distinct clusters leads to different decision boundaries, and thus different generalization abilities (cfr. illustration in Figure \ref{fig:natural_clustering_prior}). The integration of unsupervised clustering in supervised image classifiers was already suggested for non-deep learning approaches \citep{Mansur2008,Hoai2013}. Could unsupervised clustering constitute a prior of deep learning systems? Even though no clustering-related components are explicitly programmed into deep neural networks or SGD, these could emerge implicitly. Such a hypothesis is especially compelling since several works conjectured the emergence of implicit forms of regularization during deep neural network training (cfr. Section \ref{sec:GoodGeneralization}).

\section{Contributions and thesis outline}
This thesis evaluates the relevance of the natural clustering prior for understanding the generalization abilities of deep learning. It does so by identifying implicit clustering in deep learning and studying its relationship with generalization. More precisely, we provide a collection of experiments suggesting the occurrence of an implicit clustering ability (Chapter \ref{chap:implicitAbility}), an implicit clustering mechanism (Chapter \ref{chap:implicitMechanism}) and an implicit clustering hyperparameter (Chapter \ref{chap:implicitHyperparam}) in deep learning. Additionally, we show that these clustering phenomena exhibit a consistent relationship with generalization ability. Our work opens many paths of investigation. Hence, we present a discussion and future perspectives in Chapter \ref{chap:discussion}. 
\chapter{An implicit clustering ability}
\label{chap:implicitAbility}
\begin{figure}[h]
\begin{center}
\includegraphics[width=0.95\linewidth]{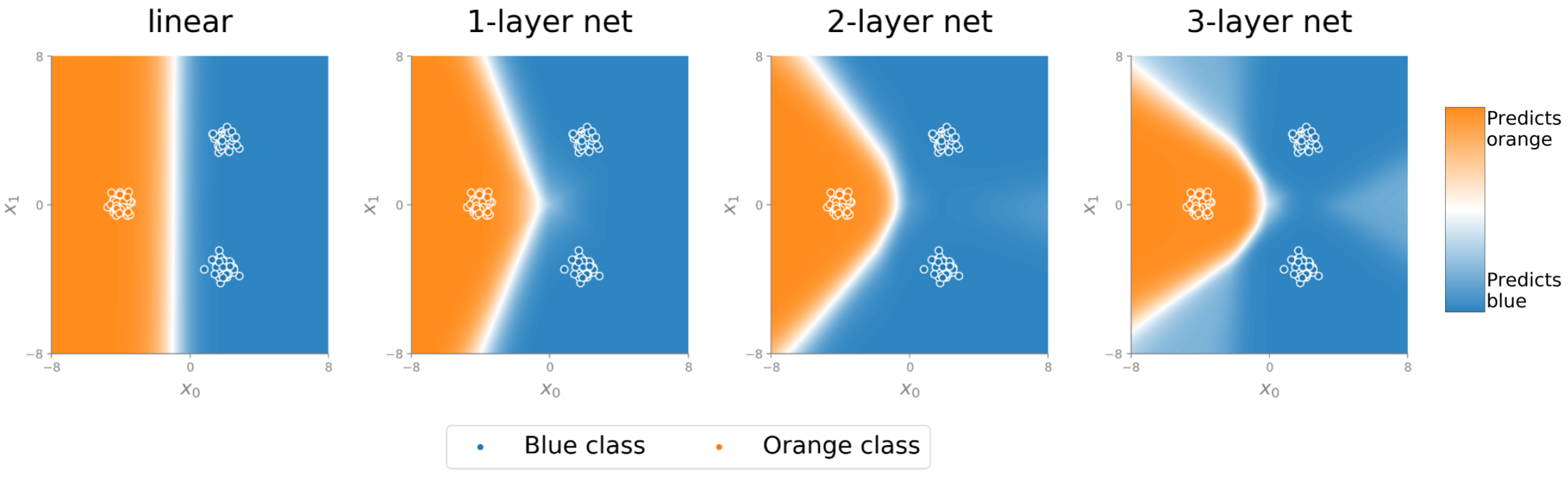}
\end{center}
\caption{Deep neural networks with different amount of layers are trained on a toy problem containing two classes (blue and orange) and two intraclass clusters associated to the blue class. For each network depth, $50$ models are trained with different initializations. The average predictions of these models is represented by the heatmaps. They reveal that the different models exhibit different intraclass clustering abilities. In particular, the linear models do not differentiate the intraclass clusters at all, while the 3-layer networks' decision boundaries tend to ``envelop" the clustered structure. In this chapter, we study the same phenomena in the high dimensions associated to real-world datasets and evaluate their relationship with generalization.}
\label{fig:ability_2D_visu}
\end{figure}

The proposed natural clustering prior suggests that unsupervised clustering abilities could benefit the generalization performance of supervised image classifiers (cfr. Section \ref{sec:naturalClusteringPrior}). While no clustering mechanisms are explicitly programmed into deep learning, these could emerge implicitly. We show in Figure \ref{fig:ability_2D_visu} that deep neural networks of sufficient depth seem to differentiate clusters belonging to the same class (i.e. \textit{intraclass clusters}) in the context of a simple 2D classification problem.

When studying standard problem settings, the main challenge resides in evaluating a model's clustering abilities without having access to the underlying mechanisms or the clusters' definitions. Hence, our work designs intraclass clustering measures based on the following three guiding principles: 
\begin{enumerate}
\item Quantify the extent by which a model differentiates examples or subclasses\footnote{Subclasses are available in datasets with hierarchical labellings, where classes (e.g., flowers) are further decomposed into multiple subclasses (e.g., orchids, poppies, roses, sunflowers, tulips).} that belong to the same class, in order to approximately capture intraclass clustering;
\item Identify measures that correlate with generalization, in order to capture phenomena that are fundamental to the learning process;
\item Study multiple measures that offer different perspectives in order to reduce the risk that the correlation with generalization is induced by phenomena independent of intraclass clustering.
\end{enumerate}
Based on these three principles, we provide five tentative measures of intraclass clustering differing in terms of representation level (black-box vs. neuron vs. layer) and the amount of knowledge about the data's inherent structure (datasets with or without hierarchical labels).

To make the link with generalization, we train more than 500 models with different generalization abilities by varying $8$ standard hyperparameters in a principled way. The measures' relationship with generalization is then evaluated qualitatively through visual inspection and quantitatively through the granulated Kendall rank-correlation coefficient introduced by \citet{Jiang}. Both evaluations reveal a tight connection between the five proposed measures and generalization ability, providing important evidence to support the occurrence and crucial role of implicit clustering abilities in deep learning. Finally, we conduct a series of experiments to provide insights on the presumed \textit{mechanisms} underlying the intraclass clustering abilities which are further studied in Chapter \ref{chap:implicitMechanism}.

\section{Measuring intraclass clustering ability}\label{sec:measures}
This section introduces the five measures of intraclass clustering ability. The measures differ in terms of representation level (black-box vs. neuron vs. layer) and the amount of knowledge about the data's inherent structure (datasets with or without hierarchical labels). An implementation of the measures based on Tensorflow \citep{Agarwal2016} and Keras \citep{chollet2015keras} is available at \url{https://github.com/Simoncarbo/Intraclass-clustering-measures}.

\subsection{Terminology and notations}
The letter $D$ denotes the training dataset and $I$ the number of classes in $D$. We denote the set of examples from class $i$ by $C_i$ with $i\in \mathcal{I}=\lbrace 1,2,...,I \rbrace$. In the case of hierarchical labels, $C_i$ denotes the samples from subclass $i$ and $S_{s(i)}$ the samples from the superclass containing subclass $i$. We denote by $\mathcal{N}=\lbrace 1,2,...,N \rbrace$ and $\mathcal{L}=\lbrace 1,2,...,L \rbrace$ the indexes of the $N$ neurons and $L$ layers of a network respectively. Neurons are considered across all the layers of a network, not a specific layer. The methodology by which indexes are assigned to neurons or layers does not matter. We further denote by $\text{mean}_{j\in \mathcal{J}}$ and $\text{median}_{j\in \mathcal{J}}$ the mean and median operations over the index $j$ respectively. Moreover, $\text{mean}^k_{j\in \mathcal{J}}$ corresponds to the mean of the top-$k$ highest values, over the index $j$.

We call pre-activations (and activations) the values preceding (respectively following) the application of the ReLU activation function \citep{Nair2010}. In our experiments, batch normalization \citep{Ioffe2015} is applied before the ReLU, and pre-activation values are collected after batch normalization. In convolutional layers, a neuron refers to an entire feature map. The spatial dimensions of such a neuron's (pre-)activations are reduced through a global max pooling operation before applying our measures.

\subsection{Measures based on label hierarchies}
The first three measures take advantage of datasets that include a hierarchy of labels. For example, CIFAR100 is organized into 20 superclasses (e.g. flowers) each comprising 5 subclasses (e.g. orchids, poppies, roses, sunflowers, tulips). We hypothesize that these hierarchical labels reflect an inherent structure of the data. In particular, we expect the subclasses to approximately correspond to different clusters amongst the samples of a superclass. Hence, measuring the extent by which a network differentiates subclasses when being trained on superclasses should reflect its ability to extract intraclass clusters during training.

\subsubsection{A black-box measure}
The first measure is black-box and is thus not restricted to deep neural networks. Motivated by the toy experiment presented in Figure \ref{fig:ability_2D_visu}, the measure is based on a model's predictions on the linear interpolation points between two training examples. We assume the model groups examples into \textit{convex} clusters. If, for two examples of a given superclass, the predicted probability of the superclass stays close to $1$ along the interpolation points, the network probably did not associate the examples to different clusters. On the contrary, a drop in the predicted probability might be reminiscent of a separation of the examples into different clusters (like in Figure \ref{fig:ability_2D_visu}). We use these intuitions to quantify the differentiation of subclasses. The measure evaluates whether the drops in predicted superclass probability are smaller when interpolating between examples from the same subclass than when interpolating between examples from different subclasses. Let $\lambda _{S_{s(i)},E_{1}\times E_{2}}$ be the average drop in the predicted probability of superclass $S_{s(i)}$ when interpolating between examples from subsets $E_{1}$ and $E_{2}$ belonging to $S_{s(i)}$. The measure is defined as:
\begin{equation} \label{eq:c0}
c_0 = \text{median}_{i\in \mathcal{I}} \text{ } \frac{\lambda _{\text{ }S_{s(i)},C_i\times C_i}}{\lambda _{\text{ }S_{s(i)},C_i\times (S_{s(i)}\setminus C_i)}}
\end{equation}
The median operation is used instead of the mean to aggregate over subclasses, as it provided a slightly better correlation with generalization. We suspect this arises from the outlier behaviour of certain subclasses observed in Section \ref{sec:visu}.

\subsubsection{Neuron-level subclass selectivity}
The second measure quantifies how selective individual neurons are for a given subclass $C_i$ with respect to the other samples of the associated superclass $S_{s(i)}$. Here, strong selectivity means that the subclass $C_i$ can be reliably discriminated from the other samples of $S_{s(i)}$ based on the neuron's pre-activations\footnote{In other words, we are interested in evaluating whether the linear projection implemented by the neuron has been effective in isolating a given subclass.}. Let $\mu _{n,E}$ and $\sigma _{n,E}$ be the mean and standard deviation of a neuron $n$'s pre-activation values taken over the samples of set $E$. The measure is defined as follows:
\begin{equation} \label{eq:c1}
c_1 = \text{median}_{i\in \mathcal{I}} \text{ mean}^k_{n\in \mathcal{N}} \text{ } \frac{\mu _{n,C_i} - \mu _{n,S_{s(i)}\setminus C_i}}{\sigma _{n,C_i} + \sigma _{n,S_{s(i)}\setminus C_i}}
\end{equation}
Since we cannot expect all neurons of a network to be selective for a given subclass, we only consider the top-$k$ most selective neurons. The measure thus relies on $k$ neurons to capture the overall network's ability to differentiate each subclass.

\subsubsection{Layer-level Silhouette score}
The third measure quantifies to what extent the samples of a subclass are close together relative to the other samples from the associated superclass \textit{in the space induced by a layer's activations}. In other words, we measure to what degree different subclasses can be associated to different clusters in the intermediate representations of a network. We quantify this by computing the pairwise cosine distances\footnote{Using cosine distances provided slightly better results than euclidean distances.} on the samples of a superclass and applying the Silhouette score \citep{Kaufman2009} to assess the clustered structure of its subclasses. This score captures the extent by which an example is close (in terms of cosine distance) to examples of its subclass compared to examples from other subclasses. Let $\textit{silhouette}(a_l,S_{s(i)},C_i)$ be the mean silhouette score of subclass $C_i$ based on the activations $a_l$ of superclass $S_{s(i)}$ in layer $l$, the measure is then defined as:
\begin{equation}
c_2 = \text{median}_{i\in \mathcal{I}} \text{ mean}^k_{l\in \mathcal{L}} \text{ } \textit{ silhouette}(a_l,S_{s(i)},C_i)
\end{equation} 

\subsection{Measures based on variance}
To establish the generality of our results, we also design two measures that can be applied in absence of hierarchical labels. We hypothesize that the discrimination of intraclass clusters should be reflected by a high variance in the representations associated to a class. If all the samples of a class are mapped to close-by points in the neuron- or layer-level representations, it is likely that the neuron/layer did not identify intraclass clusters.

\subsubsection{Variance in the neuron-level representations of the data}
\begin{figure}[h]
\begin{center}
\includegraphics[width=0.5\linewidth]{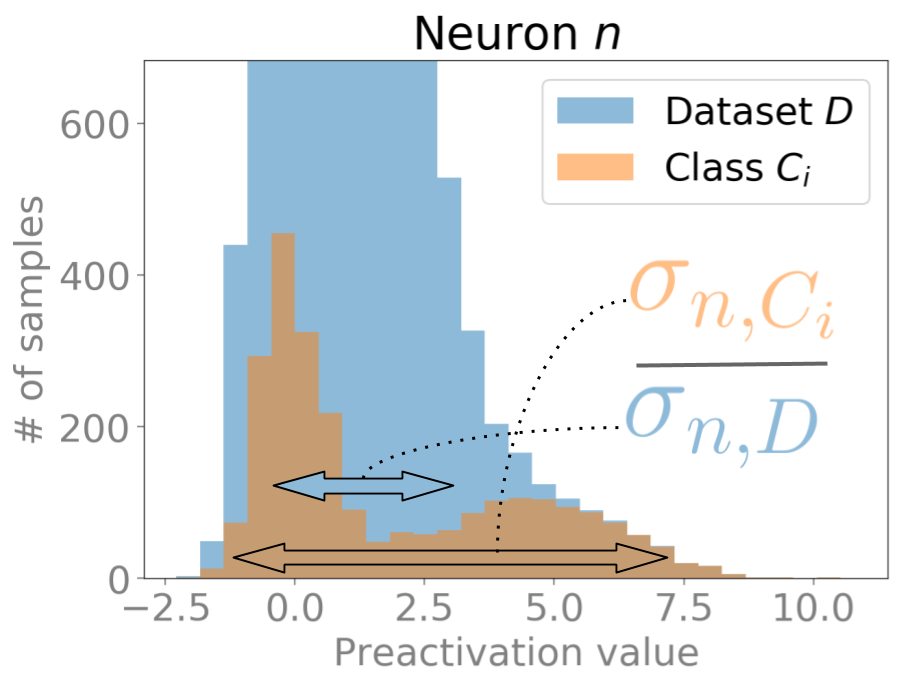}
\end{center}
\caption{Our simplest measure (denoted $c_3$) quantifies intraclass clustering through the ratio of standard deviations $\sigma _{n,C_i}$ and $\sigma _{n,D}$ associated to the class $C_i$ and the entire dataset $D$ respectively. Intuitively, a high ratio means that the neuron relies on features that \textit{differentiate} samples from $C_i$ although they belong to the same class. Despite its simplicity, our results in Section \ref{sec:results} suggest a remarkably strong connection between $c_3$ and generalization performance. This illustration of measure $c_3$ is based on a neuron from our experimental study, and the associated ratio is $2.47$.}
\label{neural_intraclass_selectivity}
\end{figure}

The first variance measure is based on standard deviations of a neuron's pre-activations. If the standard deviation computed over the samples of a class is high compared to the standard deviation computed over the entire dataset, we infer that the neuron has learned features that differentiate samples belonging to this class. The measure is defined as:
\begin{equation} \label{eq:c3}
c_3 = \text{mean}_{i\in \mathcal{I}} \text{ mean}^k_{n\in \mathcal{N}} \text{ } \frac{\sigma _{n,C_i}}{\sigma _{n,D}}
\end{equation}
A visual representation of the measure is provided in Figure \ref{neural_intraclass_selectivity}.

\subsubsection{Variance in the layer-level representations of the data}
The fifth measure transfers the neuron-level variance approach to layers by computing the standard deviations over the pairwise cosine distances calculated in the space induced by the layer's activations. Let $\Sigma _{l,E}$ be the standard deviation of the pairwise cosine distances between the samples of set $E$ in the space induced by layer $l$. The measure is defined as:
\begin{equation}
c_4 = \text{mean}_{i\in \mathcal{I}} \text{ mean}^k_{l\in \mathcal{L}} \text{ } \frac{\Sigma _{l,C_i}}{\Sigma _{l,D}}
\end{equation}
To improve this measure's correlation with generalization, we found it helpful to standardize the representations of different neurons. More precisely, we normalize each neuron's pre-activations to have zero mean and unit variance, then apply a bias and ReLU activation function such that $25\%$ of the samples are activated\footnote{Activating $25\%$ of the samples was an arbitrary choice that we did not seek to optimize.}. This makes the measure invariant to rescaling and translation of each neuron's preactivations.

\section{Experimental methodology} \label{sec:expsetup}
The purpose of our experimental endeavour is to assess the relationship between the proposed intraclass clustering measures and generalization performance. To this end, we reproduce the methodology introduced by \citet{Jiang}. First of all, this methodology puts emphasis on the scale of the experiments to improve the generality of the observations. Second, it tries to go beyond standard measures of correlation, and puts extra care to detect causal relationships between the measures and generalization performance. This is achieved through a systematic variation of multiple hyperparameters when building the set of models to be studied, combined with the application of principled correlation measures.

\subsection{Building a set of models with varying hyperparameters} \label{sec:hyperparam}
Our experiments are conducted on three datasets and two network architectures. The datasets are CIFAR10, CIFAR100 and the coarse version of CIFAR100 with 20 superclasses \citep{Krizhevsky2009}. The two network architectures are Wide ResNets \citep{He2016,Zagoruyko2016} (applied on CIFAR100 datasets) and VGG variants \citep{Simonyan2014} (applied on CIFAR10 dataset). Both architectures use batch normalization layers \citep{Ioffe2015} since they greatly facilitate the training procedure.

In order to build a set of models with a wide range of generalization performances, we vary hyperparameters that are known to be critical. Since varying multiple hyperparameters improves the identification of causal relationships, \textit{we vary 8 different hyperparameters}: learning rate, batch size, optimizer (SGD or Adam \citep{Kingma2015}), weight decay, dropout rate \citep{Srivastava2014}, data augmentation, network depth and width. A straightforward way to generate hyperparameter configurations is to specify values for each hyperparameter independently and then generate all possible combinations. However, given the amount of hyperparameters, this quickly leads to unrealistic amounts of models to be trained.

To deal with this, we decided to remove co-variations of hyperparameters whose influence on training and generalization is suspected to be related. More precisely, we use weight decay only in combination with the highest learning rate value, as recent works demonstrated a relation between weight decay and learning rate \citep{VanLaarhoven2017,Zhang2019}. We also don't combine dropout and data augmentation, as the effect of dropout is drastically reduced when data augmentation is used. Finally, we do not jointly increase width and depth, to avoid very large models that would slow down our experiments. 

The resulting hyperparameter values are as follows:
\begin{enumerate}
\item \textbf{(Learning rate, Weight decay)}: $\lbrace (0.01,0.), (0.32,0.), (0.1,0.), (0.1, 4\times 10^{-5})\rbrace$
\item \textbf{Batch size}: $\lbrace 100,300 \rbrace$
\item \textbf{Optimizer}: $\lbrace SGD, Adam \rbrace$
\item \textbf{(Dropout rate, Data augm.)}: $\lbrace (0.,\textit{true}), (0.,\textit{false}), (0.2,\textit{false}), (0.4,\textit{false})\rbrace$
\item \textbf{(Width factor, Depth factor)}: $\lbrace (\times 1.,\times 1.), (\times 1.5,\times 1.), (\times 1.,\times 1.5))\rbrace$
\end{enumerate}
We generate all possible combinations of these hyperparameter values (or pairs of values), leading to $192$ configurations. Since dropout rates of $0.4$ lead to poor training performance on VGG variants, only $144$ configurations are used in these cases.

\begin{figure}[h]
\begin{center}
\includegraphics[width=.85\linewidth]{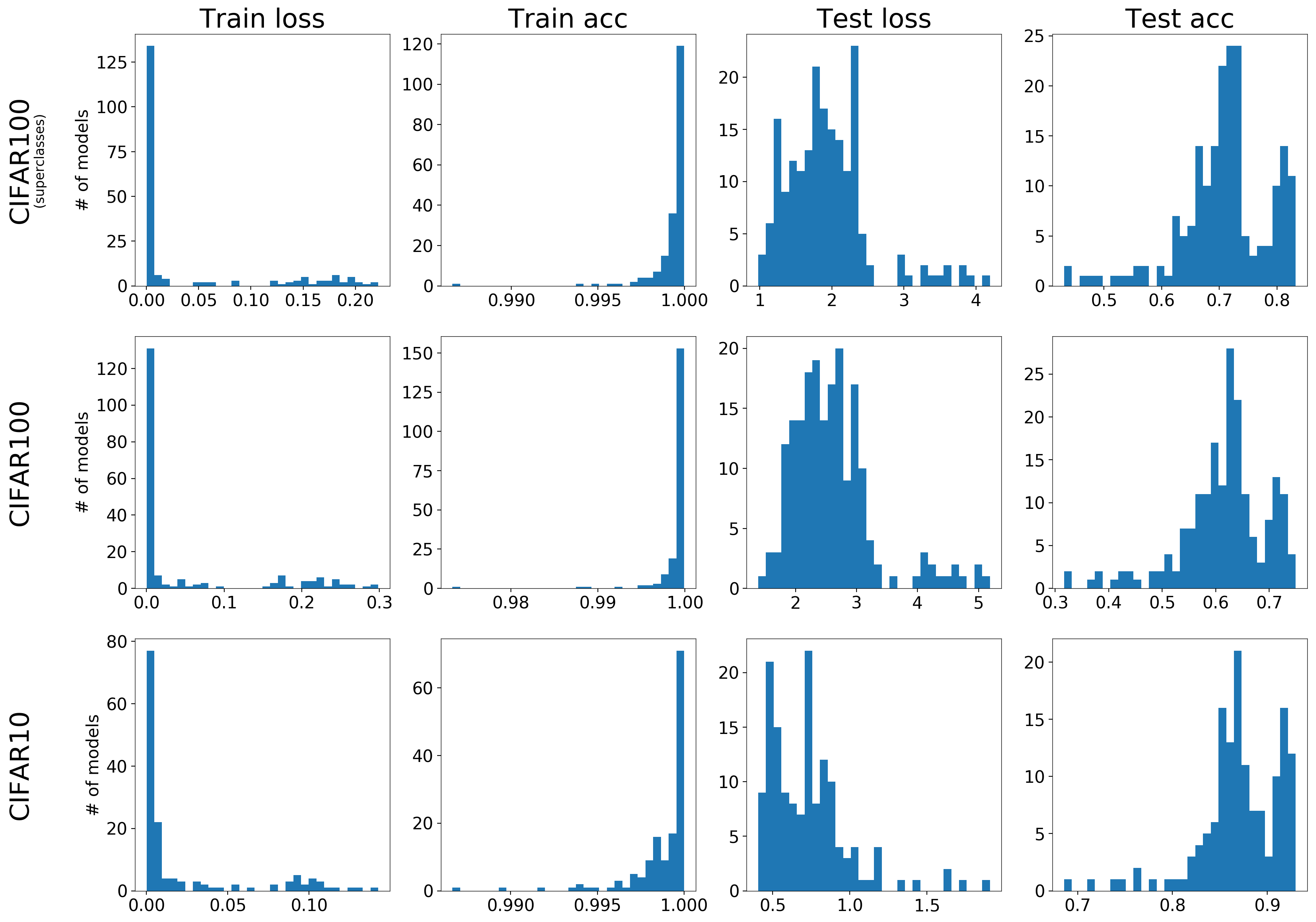}
\end{center}
\caption{Histogram of performances of the set of models used in our experiments.}
\label{fig:model_perfs}
\end{figure}

We train all the models for $250$ epochs, and reduce the learning rate by a factor $0.2$ at epochs $150,230,240$. Training stops prematurely if the training loss gets smaller than $10^{-4}$. Since different optimizers may require different learning rates for optimal performance \citep{Wilson2017}, we divide the learning rate by $100$ when using $Adam$ to improve its performance in our experiments (the same approach is used in \citet{Jiang}). Overall, all networks reach close to $100\%$ training accuracy, as reported by Figure \ref{fig:model_perfs}.

\subsection{Evaluating correlation with generalization} \label{sec:kendall}
\citet{Jiang} provides multiple criteria to evaluate the relationship between a measure and generalization. We opted for the granulated Kendall coefficient for its simplicity and intuitiveness. This coefficient compares two rankings of the models, respectively provided by (i) the measure of interest and (ii) the models' generalization performances. The Kendall coefficient is computed across variations of each hyperparameter independently. The average over all hyperparameters is then computed for the final score. The goal of this approach is to better capture causal relationships by not overvaluing measures that correlate with generalization only when specific hyperparameters are tuned.

We compare our intraclass clustering-based measures to sharpness-based measures. The latter constituted the most promising measure family from the large-scale study presented in \citet{Jiang}. Among the many different sharpness measures, we leverage the magnitude-aware versions that measure sharpness through random and worst-case perturbations of the weights (denoted by $\frac{1}{\sigma '}$ and $\frac{1}{\alpha '}$, respectively, in \citet{Jiang}). We also include the application of these measures with perturbations applied on kernels only (i.e. not on biases and batch normalization weights) with batch normalization layers in batch statistics mode (i.e. not in inference mode). We observed that these alternate versions often provided better estimations of generalization performance. We denote these measures by $\frac{1}{\sigma ''}$ and $\frac{1}{\alpha ''}$. 

\section{Results} \label{sec:results}
This section starts with a thorough evaluation of the relationship between the five proposed measures and generalization performance, using the setup described in Section \ref{sec:expsetup}. Then, it presents a series of experiments to better characterize intraclass clustering, the phenomenon we expect to be captured by the measures. These experiments include (i) an analysis of the measures' evolution across layers and training iterations, (ii) a study of the neuron-level measures' sensitivity to $k$ in the mean over top-$k$ operation, as well as (iii) visualizations of subclass extraction in individual neurons.

\subsection{The measures' relationships with generalization}
\label{sec:evaluation}
We compute all five measures on the models trained on the CIFAR100 superclasses, and only the two variance-based measures on the models trained on standard CIFAR100 and CIFAR10 -because they don't provide subclass information. We set $k=30$ for the neuron-level measures, meaning that $30$ neurons per subclass (for $c_1$) or class (for $c_3$) are used to capture intraclass clustering. For the layer-level measures, we set $k=5$ for residual networks and $k=1$ for VGG networks.

\begin{figure}[h]
\begin{center}
\includegraphics[width=1.\linewidth]{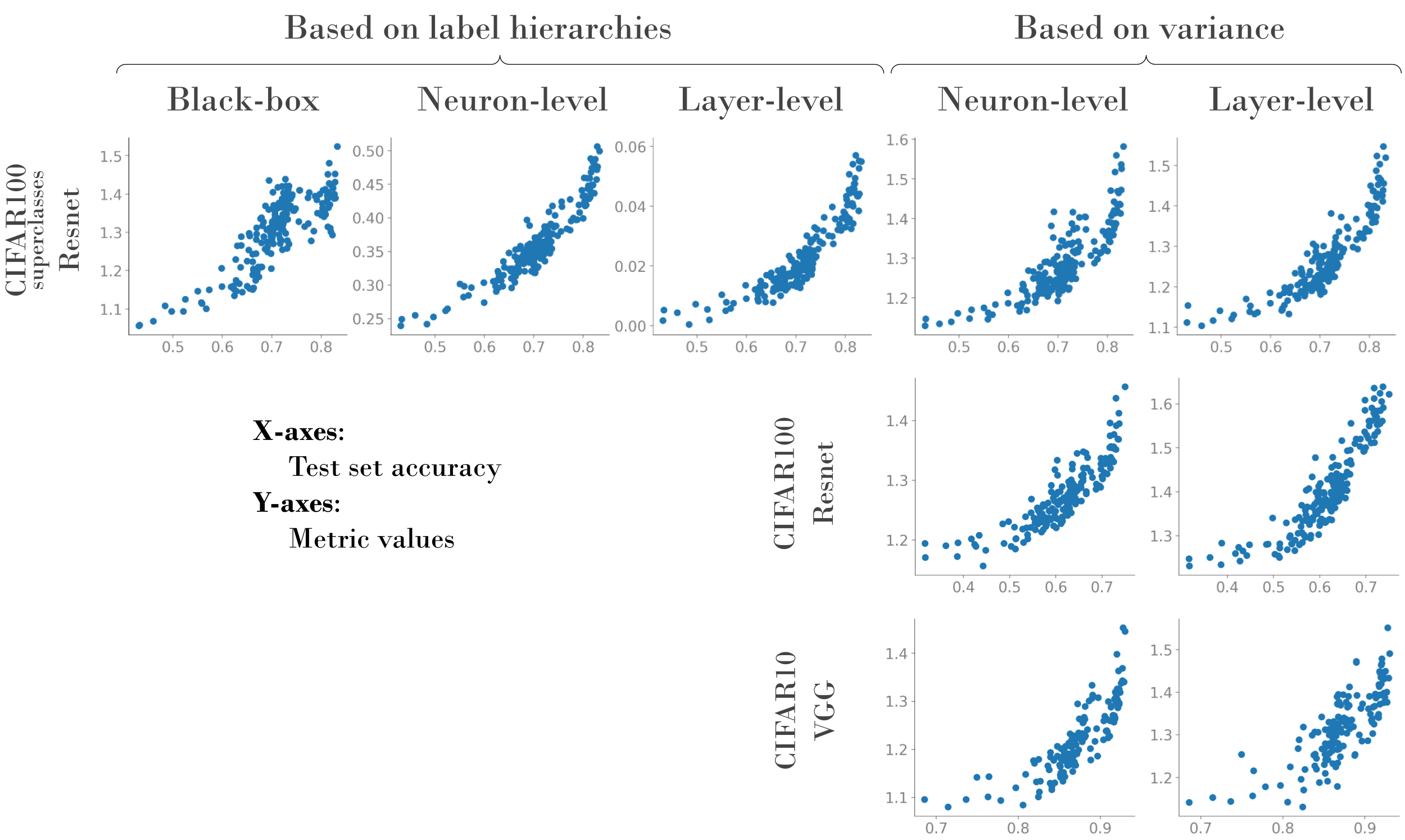}
\end{center}
\caption{Visualization of the relationship between the five proposed intraclass clustering measures and generalization performance, across datasets and network architectures. The five columns correspond to $c_0$, $c_1$, $c_2$, $c_3$ and $c_4$ measures respectively. All measures display a tight connection with generalization performance.}
\label{generalization_intraclass}
\end{figure}

We start our evaluation of the measures by visualizing their relationship with generalization performance in Figure \ref{generalization_intraclass}. \textit{We observe a clear correlation across datasets, network architectures and measures.} To further support the conclusions of our visualizations, we evaluate the measures through the granulated Kendall coefficient (cfr. Section \ref{sec:kendall}). Tables \ref{results_1}, \ref{results_2} and \ref{results_3} present the granulated Kendall rank-correlation coefficients associated with intraclass clustering and sharpness-based measures, for the three dataset-architecture pairs.

The Kendall coefficients further confirm the observations in Figure \ref{generalization_intraclass} by revealing strong correlations between intraclass clustering measures and generalization performance \textit{across all hyperparameters}. In terms of overall score, intraclass clustering measures surpass the sharpness-based measures variants by a large margin across all dataset-architecture pairs. On some specific hyperparameters, sharpness-based measures outperform intraclass clustering measures. In particular, $\frac{1}{\alpha '}$ performs remarkably well when the batch size parameter is varied, which is coherent with previous work \citep{Keskar2017}. 

\renewcommand\cellalign{cc}
\begin{table}[!h]
\caption{Kendall coefficients for resnets trained on CIFAR100 superclasses. The higher the coefficient, the stronger the correlation with generalization. Correlations are measured across variations of specific hyperparameters (cfr. $8$ first columns) or all of them (cfr. last column).}
\label{results_1}
\begin{center}
\resizebox{\textwidth}{!}{
\begin{tabular}{cc cccccccc |c}
  & & \makecell{learning \\ rate }&\makecell{batch \\ size }&\makecell{weight \\ decay } & optim. &\makecell{dropout \\ rate }&\makecell{data \\ augm. } & width &depth&\makecell{ total \\ score }\\
\hline
\multirow{5}{*}{\makecell{Intraclass \\ clustering }} & \multicolumn{1}{|c|}{$c_0$} & 0.57 & 0.5 & 0.21 & 0.27 & 0.81 &\bf 1.0 & 0.25 & 0.22 & 0.48\\
& \multicolumn{1}{|c|}{$c_1$} & 0.88 & 0.31 & 0.38 & 0.67 & 0.96 &\bf 1.0 &\bf 0.81 & 0.69 & 0.71\\
 & \multicolumn{1}{|c|}{$c_2$} & 0.86 & 0.5 &\bf 0.67 & 0.58 &\bf 0.99 &\bf 1.0 & 0.38 & 0.62 & 0.7\\
 & \multicolumn{1}{|c|}{$c_3$} & 0.88 & 0.6 & 0.46 & 0.62 & 0.81 &\bf 1.0 &\bf 0.81 & 0.66 & \bf 0.73\\
 & \multicolumn{1}{|c|}{$c_4$} &\bf 0.89 & 0.69 & 0.62 & 0.65 & 0.86 &\bf 1.0 & 0.44 & 0.69 & \bf 0.73\\
\hline
\multirow{4}{*}{\makecell{Sharpness}} & \multicolumn{1}{|c|}{$1$ / $\sigma '$} & 0.81 & 0.51 & 0.31 & 0.69 & 0.28 & -0.58 & 0.67 & 0.61 & 0.41\\
 &\multicolumn{1}{|c|}{$1$ / $\sigma ''$} & 0.86 & 0.58 & 0.17 & 0.4 & -0.05 & 0.42 & 0.69 & \bf 0.72 & 0.47\\
 & \multicolumn{1}{|c|}{$1$ / $\alpha '$} & 0.88 &\bf 0.94 & 0.29 & 0.26 & 0.6 & 0.08 & -0.03 & -0.09 & 0.37\\
 & \multicolumn{1}{|c|}{$1$ / $\alpha ''$} & 0.85 & 0.8 & 0.48 &\bf 0.71 & 0.16 & -0.08 & 0.08 & 0.34 & 0.42\\
\hline
\end{tabular}}
\end{center}
\caption{Kendall coefficients for resnets trained on CIFAR100.}
\label{results_2}
\begin{center}
\resizebox{\textwidth}{!}{
\begin{tabular}{cc cccccccc |c}
  & & \makecell{learning \\ rate }&\makecell{batch \\ size }&\makecell{weight \\ decay } & optim. &\makecell{dropout \\ rate }&\makecell{data \\ augm. } & width &depth&\makecell{ total \\ score }\\
\hline
\multirow{2}{*}{\makecell{Intraclass \\ clustering }} & \multicolumn{1}{|c|}{$c_3$} & 0.94 & 0.65 &\bf 0.62 & 0.58 &\bf 1.0 &\bf 1.0 &\bf 1.0 &\bf 0.78 &\bf 0.82\\
 & \multicolumn{1}{|c|}{$c_4$} & 0.93 & 0.62 &\bf 0.62 & 0.21 &\bf 1.0 &\bf 1.0 & 0.91 &\bf 0.78 & 0.76\\
\hline
\multirow{4}{*}{\makecell{Sharpness}} & \multicolumn{1}{|c|}{$1$ / $\sigma '$} & 0.88 & 0.68 & 0.17 &\bf 0.8 & 0.4 & -0.62 & 0.94 & 0.61 & 0.48\\
 &\multicolumn{1}{|c|}{$1$ / $\sigma ''$} & 0.92 & 0.61 & 0.12 & 0.35 & -0.06 & 0.31 & 0.94 & 0.53 & 0.47\\
 & \multicolumn{1}{|c|}{$1$ / $\alpha '$} &\bf 0.96 &\bf 0.96 & 0.17 & 0.25 & 0.54 & 0.15 & -0.16 & -0.23 & 0.33\\
 & \multicolumn{1}{|c|}{$1$ / $\alpha ''$} &\bf 0.96 & 0.91 & 0.42 & 0.64 & 0.12 & -0.25 & 0.17 & 0.14 & 0.39\\
\hline
\end{tabular}}
\end{center}
\caption{Kendall coefficients for VGG networks trained on CIFAR10.}
\label{results_3}
\begin{center}
\resizebox{\textwidth}{!}{
\begin{tabular}{cc cccccccc |c}
  & & \makecell{learning \\ rate }&\makecell{batch \\ size }&\makecell{weight \\ decay } & optim. &\makecell{dropout \\ rate }&\makecell{data \\ augm. } & width &depth&\makecell{ total \\ score }\\
\hline
\multirow{2}{*}{\makecell{Intraclass \\ clustering }} & \multicolumn{1}{|c|}{$c_3$} & 0.92 & 0.83 &\bf 0.67 & 0.51 &\bf 0.92 &\bf 1.0 &\bf 1.0 & 0.88 &\bf 0.84\\
 & \multicolumn{1}{|c|}{$c_4$} & 0.86 & 0.75 & 0.33 & 0.29 &\bf 0.92 &\bf 1.0 & 0.54 & 0.92 & 0.7\\
\hline
\multirow{4}{*}{\makecell{Sharpness}} & \multicolumn{1}{|c|}{$1$ / $\sigma '$} & 0.86 & 0.62 & -0.25 & 0.6 & -0.04 & -0.27 &\bf 1.0 & 0.85 & 0.42\\
 &\multicolumn{1}{|c|}{$1$ / $\sigma ''$} & 0.9 & 0.67 & 0.11 &\bf 0.69 &\bf 0.92 & 0.19 &\bf 1.0 &\bf 0.94 & 0.68\\
 & \multicolumn{1}{|c|}{$1$ / $\alpha '$} &\bf 0.94 &\bf 0.89 & 0.61 & 0.53 & 0.67 & 0.77 & 0.15 & 0.06 & 0.58\\
 & \multicolumn{1}{|c|}{$1$ / $\alpha ''$} & 0.93 & 0.67 & 0.36 & 0.54 & 0.69 & -0.02 & 0.15 & -0.15 & 0.4\\
\hline

\end{tabular}}
\end{center}
\end{table}

\subsection{Influence of $k$ on the Kendall coefficients} \label{sec:k_sensitivity}
\begin{figure}[h]
\begin{center}
\includegraphics[width=.85\linewidth]{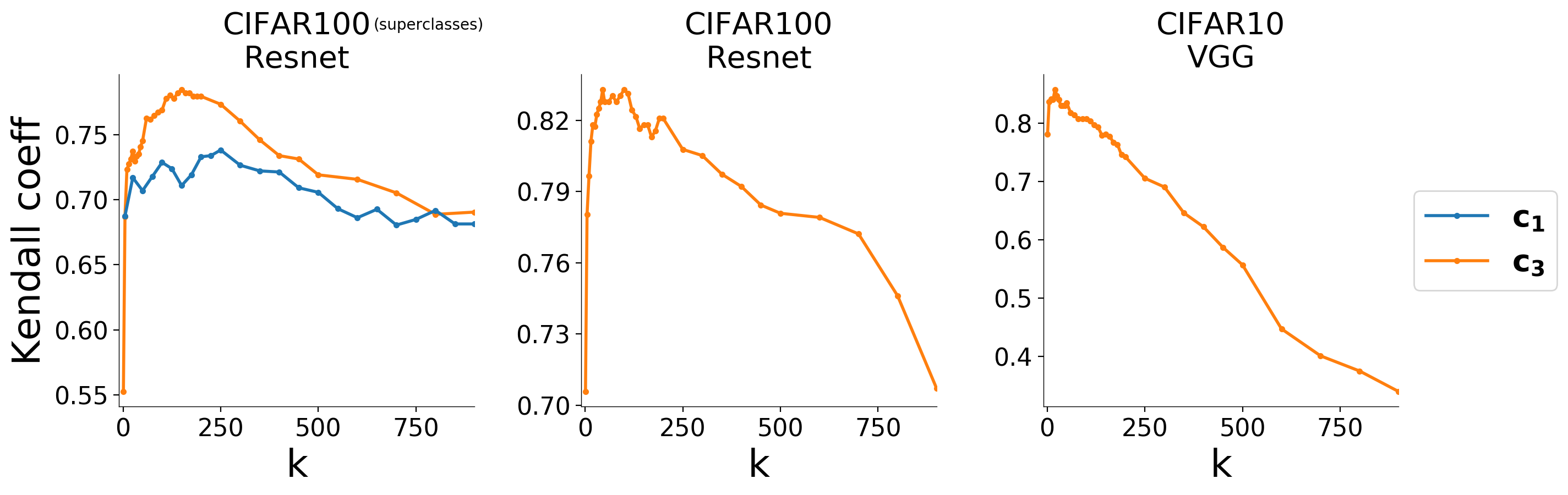}
\end{center}
\caption{Plots showing how the Kendall coefficients of $c_1$ and $c_3$ change with parameter $k$ (cfr. Equations \ref{eq:c1} and \ref{eq:c3}). The $k$ parameter associated to $c_1$ is multiplied by $5$ in the plots, to enable comparison with $c_3$ (there are $5$ subclasses in each of CIFAR100's superclasses). The total number of neurons varies from $1920$ to $2880$ in Resnets and from $960$ to $1440$ in VGGs. The plots reveal that generalization performance can be quite accurately estimated using the representations of a surprisingly small set of neurons ($k=1$, i.e. a single neuron per class, suffices in some cases).}
\label{fig:locality}
\end{figure}

In our evaluation of the measures in Section \ref{sec:evaluation}, the $k$ parameter, which controls the number of highest values considered in the mean over top-$k$ operations, was fixed quite arbitrarily. Figure \ref{fig:locality} shows how the Kendall coefficient of $c_1$ and $c_3$ changes with this parameter. We observe a relatively low sensitivity of the measures' predictive power with respect to $k$. In particular, in the case of Resnets trained on CIFAR100 the Kendall coefficient associated with $c_3$ seems to stay above $0.7$ for any $k$ in the range $[1,900]$. The optimal $k$ value changes with the considered dataset and architecture. We leave the study of this dependency as a future work.

Observing the influence of $k$ also confers insights about the phenomenon captured by the measures. Figure \ref{fig:locality} reveals that very small $k$ values work remarkably well. Using a single neuron per subclass ($k=1$ in Equation \ref{eq:c1}) confers a Kendall coefficient of $0.69$ to $c_1$. Using a single neuron per class confers a Kendall coefficient of $0.78$ to $c_3$ in the case of VGGs trained on CIFAR10. \textit{These results suggest that individual neurons play a crucial role in the extraction of intraclass clusters during training.} The fact that the Kendall coefficients monotonically decrease after some $k$ value suggests that the extraction of a given intraclass cluster takes place in a sub part of the network, indicating some form of specialization.

\subsection{Evolution of the measures across layers}
\begin{figure}[h]
\begin{center}
\includegraphics[width=.9\linewidth]{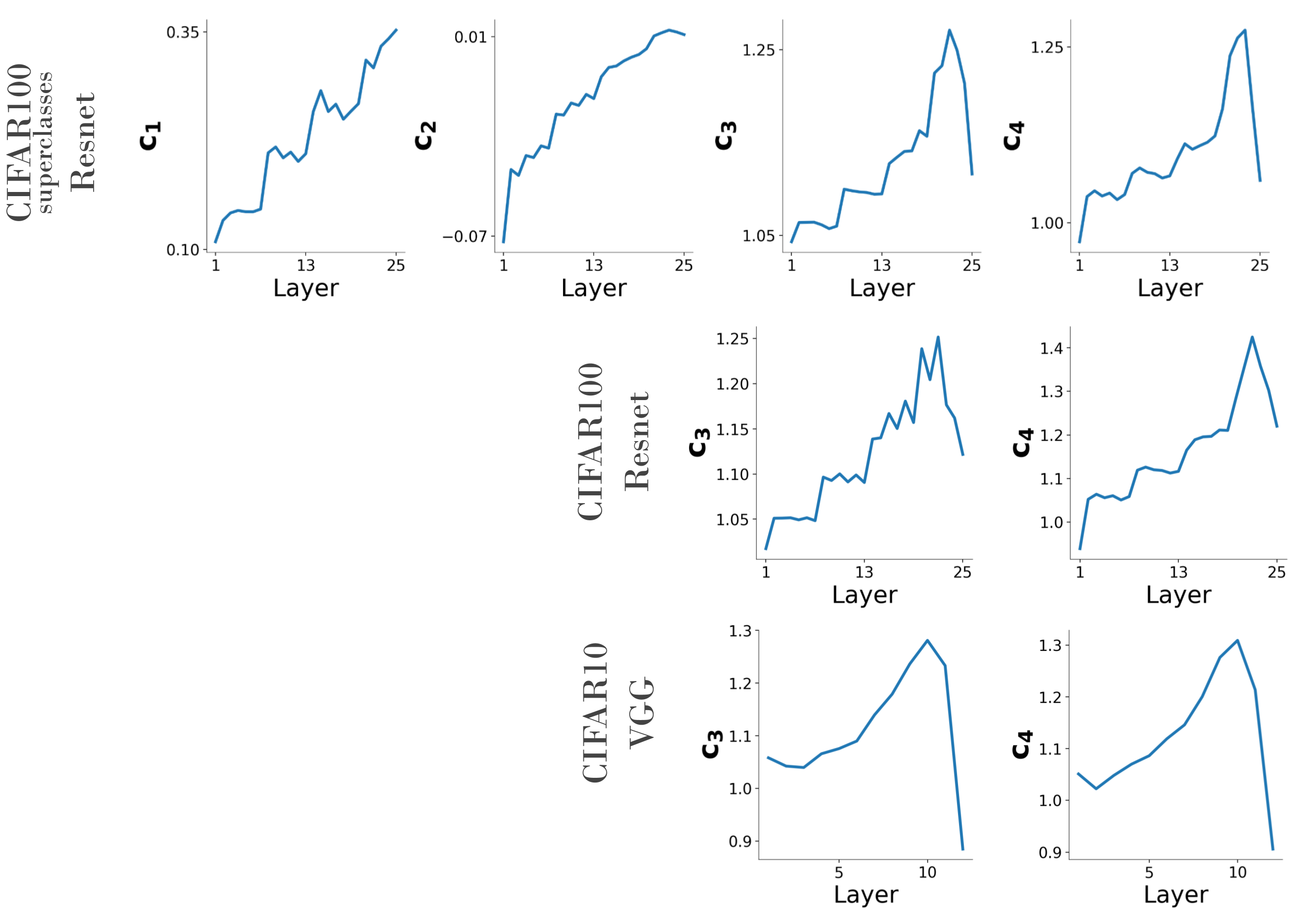}
\end{center}
\caption{Evolution of each measure (after averaging over 64 models) across layers for the three dataset-architecture pairs. The overall increase of the measures with layer depth suggests that intraclass clustering occurs even in the deepest representations of neural networks.}
\label{fig:layerwise}
\end{figure}

We pursue our experimental endeavour with an analysis of the proposed measures' evolution across layers. For each dataset-architecture pair, we select $64$ models which have the same depth hyperparameter value. We then compute the four measures on a layer-level basis (we use the top-5 neurons of each layer for the neuron-level measures) and average the resulting values over the $64$ models. Figure \ref{fig:layerwise} depicts how the average value of each measure evolves across layers for the three dataset-architecture pairs.

We observe two interesting trends. First, all four measures tend to increase with layer depth. \textit{This suggests that intraclass clustering also occurs in the deepest representations of neural networks, and not merely in the first layers, which are commonly assumed to capture generic or class-independent features.} Second, the variance based measures ($c_3$ and $c_4$) decrease drastically in the penultimate layer. We suspect this reflects the grouping of samples of a class in tight clusters in preparation for the final classification layer (such behaviour has been studied in \citet{Kamnitsas2018,Muller2019}). The measures $c_1$ and $c_2$ are robust to this phenomenon as they rely on relative distances inside a single class, irrespectively of the representations of the rest of the dataset.

\subsection{Evolution of the measures over the course of training} \label{sec:evolution_training}
\begin{figure}[h]
\begin{center}
\includegraphics[width=1.\linewidth]{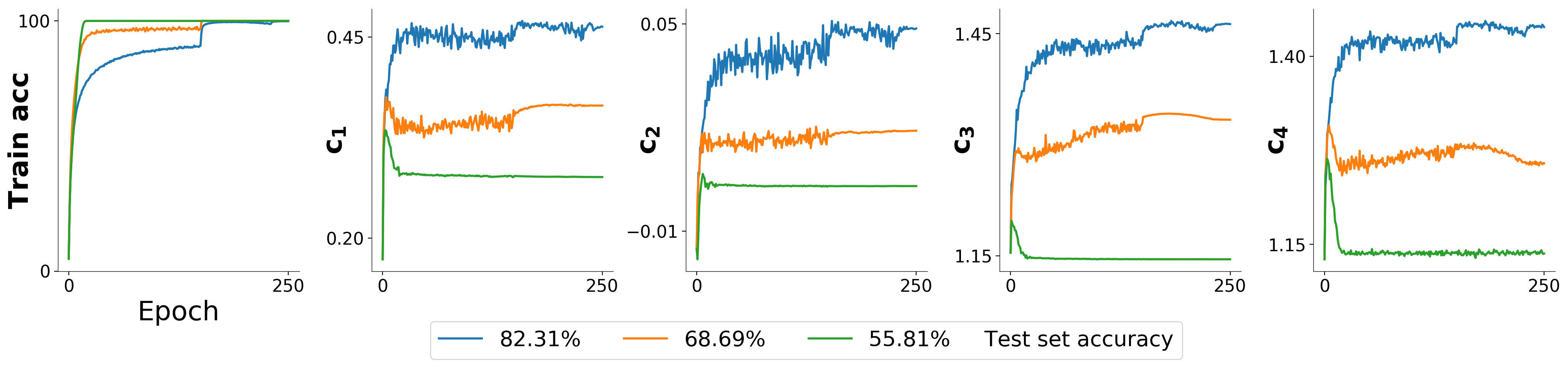}
\end{center}
\caption{Evolution of the intraclass clustering measures over the course of training for three models with different generalization performances. We observe that the differences between models with high and low generalization performance arise essentially in the early phase of training.}
\label{train_evolution}
\end{figure}

In this section, we provide a small step towards the understanding of the dynamics of the phenomenon captured by the measures. We visualize in Figure \ref{train_evolution} the evolution of the measures over the course of training of three Resnet models. The first interesting observation comes from the comparison of models with high and low generalization performances. It appears that \textit{their differences in terms of intraclass clustering measures arise essentially during the early phase of training}. The second observation is that significant increases in intraclass clustering measures systematically coincide with significant increases of the training accuracy (in the few first epochs and around epoch $150$, where the learning rate is reduced). This suggests that supervised training could act as a necessary driver for intraclass clustering ability, despite not explicitly targeting such behaviour.

\subsection{Visualization of subclass extraction in hidden neurons} \label{sec:visu}
\begin{figure}[h]
\begin{center}
\includegraphics[width=.85\linewidth]{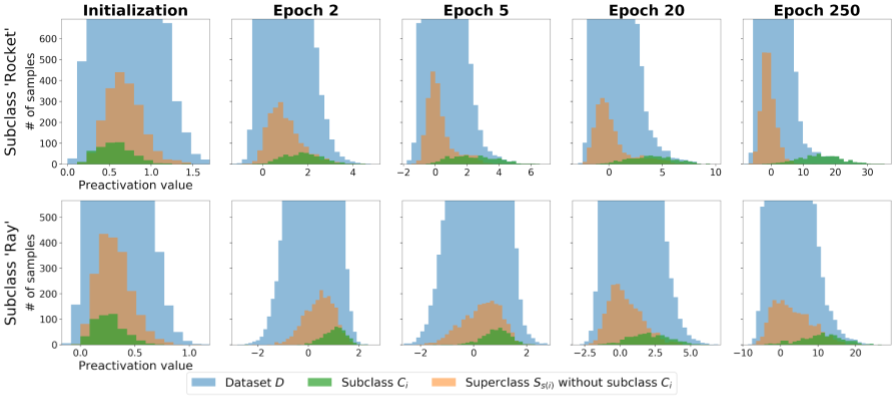}
\end{center}
\caption{Evolution along training of the preactivation distributions associated with the neurons that are the most selective (cfr. Eq. \ref{eq:c1}) for 'Rocket' and 'Ray' subclasses. The neurons behave like they were trained to identify these specific subclasses although no supervision or explicit training mechanisms were implemented to target this behaviour.}
\label{fig:selectivity_visu}
\end{figure}

We have seen in Section \ref{sec:k_sensitivity} that the measure $c_1$ reaches a Kendall coefficient of $0.69$ when considering a single neuron per subclass ($k=1$ in Eq. \ref{eq:c1}). Visualizing the training dynamics in this specific neuron should enable us to directly observe the phenomenon captured by $c_1$. We study a Resnet model trained on CIFAR100 superclasses with high generalization performance ($82.31\%$ test accuracy). For each of the $100$ subclasses, we compute the selectivity value and the index of the most selective neuron based on the part of Eq. \ref{eq:c1} to which the median operation is applied. We then rank the subclasses by their selectivity value, and display the training dynamics of the neurons associated to the subclasses with maximum and median selectivity values in Figure \ref{fig:selectivity_visu}.

\begin{figure}[h]
\begin{center}
\includegraphics[width=.4\linewidth]{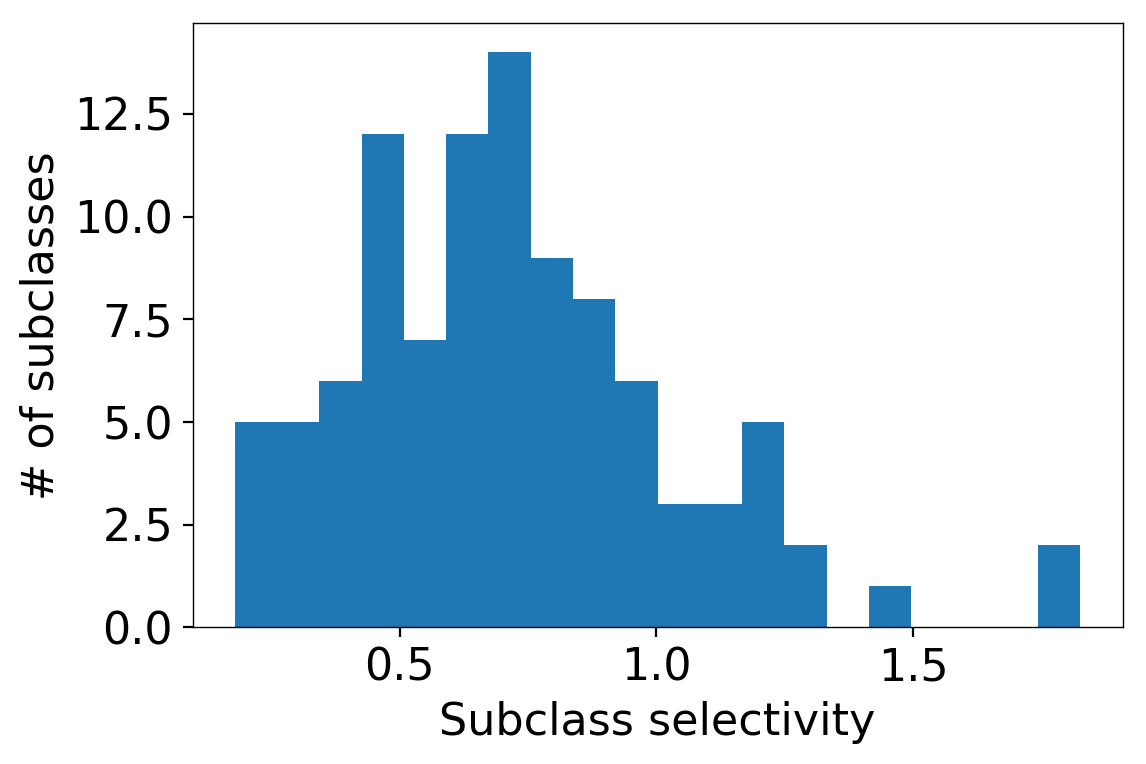}
\end{center}
\caption{Distribution of neural subclass selectivity values (cfr. measure $c_1$) over the $100$ subclasses of CIFAR100. For each subclass, neural subclass selectivity is computed based on the most selective neuron in the neural network (i.e. $k=1$). We observe that (i) only a few subclasses reach high selectivity values and (ii) the selectivity values vary much across subclasses. We suspect that the outliers with exceptionally high selectivity values cause the median operation to outperform the mean in the measures $c_1$ and $c_2$.}
\label{fig:selectivity_distrib}
\end{figure}

The evolution of the neurons' preactivation distributions along training reveals that the 'Rocket' subclass, which has the highest selectivity value, is progressively distinguished from its corresponding superclass during training. \textit{The neuron behaves like it was trained to identify this specific subclass although no supervision or explicit training mechanisms were implemented to target this behaviour.} The same phenomenon occurs to a lesser extent with the 'Ray' subclass, which has the median selectivity value. We observed that very few subclasses reached selectivity values as high as the 'Rocket' subclass (the distribution of selectivity values is provided in Figure \ref{fig:selectivity_distrib}). We suspect that the occurrence of such outliers explain why the median operation outperformed the mean in the definition of $c_1$ and $c_2$. 

\section{Related work}
Many observations made in this chapter are coherent with previous work. In the context of transfer learning, \citet{Huh2016a} shows that representations that discriminate ImageNet classes naturally emerge when training on their corresponding superclasses, suggesting the occurrence of intraclass clustering. Sections \ref{sec:k_sensitivity} and \ref{sec:visu} suggest a key role for individual neurons in the extraction of intraclass clusters. This is coherent with the large body of work that studied the emergence of interpretable features in the hidden neurons (or feature maps) of deep nets \citep{Zeiler2014,ZissermanICLR2014,Yosinski2015,Zhou2015, Bau}. In Section \ref{sec:evolution_training}, we notice that intraclass clustering occurs mostly in the early phase of training. Previous works have also highlighted the criticality of this phase of training with respect to regularization \citep{Golatkar2019}, optimization trajectories \citep{Jastrzebski2020,Fort2020}, Hessian eigenspectra \citep{Gur-Ari2016}, training data perturbations \citep{Achille2019} and weight rewinding \citep{Frankle2020a, Frankle2020}. \citet{Morcos2018a,Leavitt2020} have shown that class-selective neurons are not necessary and might be detrimental for performance. This is coherent with our observation that neurons that differentiate samples from the same class improve performance. 

\section{Discussion} \label{sec:implicitAbilityDiscussion}
Our results show that the measures proposed in Section \ref{sec:measures} (i) correlate with generalization, (ii) tend to increase with layer depth and (iii) change mostly in the early phase of training. These similarities suggest that the measures capture one unique phenomenon. Since all measures quantify to what extent a neural network differentiates samples from the same class, the captured phenomenon presumably consists in intraclass clustering. This hypothesis is further supported by the neuron-level visualizations provided in Section \ref{sec:visu}. Overall, our results thus provide empirical evidence for this thesis' hypotheses, i.e. that implicit clustering abilities emerge during standard deep neural network training and improve their generalization abilities. 

However, the assessment of the causal nature of the measures' relationships with clustering and generalization still relies on sophisticated correlation measures and informal arguments. Identifying implicit clustering \textit{mechanisms} in deep learning would further support our hypotheses by strengthening causality. Interestingly, our results provide some insights on these presumed mechanisms. In particular, the neuron-level measures correlate with generalization performance as strongly as the layer-level measures in our experiments. As suggested by Section \ref{sec:k_sensitivity}, the behaviour of some carefully selected neurons seems to quite accurately predict properties of the entire neural network they belong to. Figure \ref{fig:selectivity_visu} further suggests that individual neurons seem to possess a training routine of their own, targeting the classification of a subclass. All these results indicate that individual neurons play a crucial role in the presumed clustering mechanisms of deep learning. The next chapter of this thesis delves into these intriguing phenomena.

\chapter{An implicit clustering mechanism}
\label{chap:implicitMechanism}
Chapter \ref{chap:implicitAbility} studies five tentative measures of implicit clustering ability in deep learning that appear to strongly correlate with generalization. In order to strengthen the causal relationship between the proposed measures and clustering, this chapter tries to unveil the clustering \textit{mechanisms} that presumably underlie these abilities. This requires delving into the inner workings of deep neural network training.

The main difficulty arises from the fact that SGD is a global or end-to-end optimization algorithm. Contrary to biologically inspired alternatives, SGD is not based on the repetition of local, neuron-level mechanisms whose behavior is much simpler to study and understand (e.g. \citet{Hebb1949a,Rosenblatt1958}). However, several works have suggested that individual neurons exhibit localized behavior during SGD-based training, too. Most notably, a large body of work has observed empirically that interpretable features are captured by the hidden neurons (or feature maps) of trained deep neural networks \citep{Zeiler2014,ZissermanICLR2014,Yosinski2015,Zhou2015, Bau,Cammarata2020}. Additionally, as discussed in Section \ref{sec:implicitAbilityDiscussion}, multiple experiments of Chapter \ref{chap:implicitAbility} indicate that individual neurons play a crucial role in the presumed clustering mechanisms of deep learning. 

\begin{figure}[h!]
\begin{center}
\includegraphics[width=0.7\linewidth]{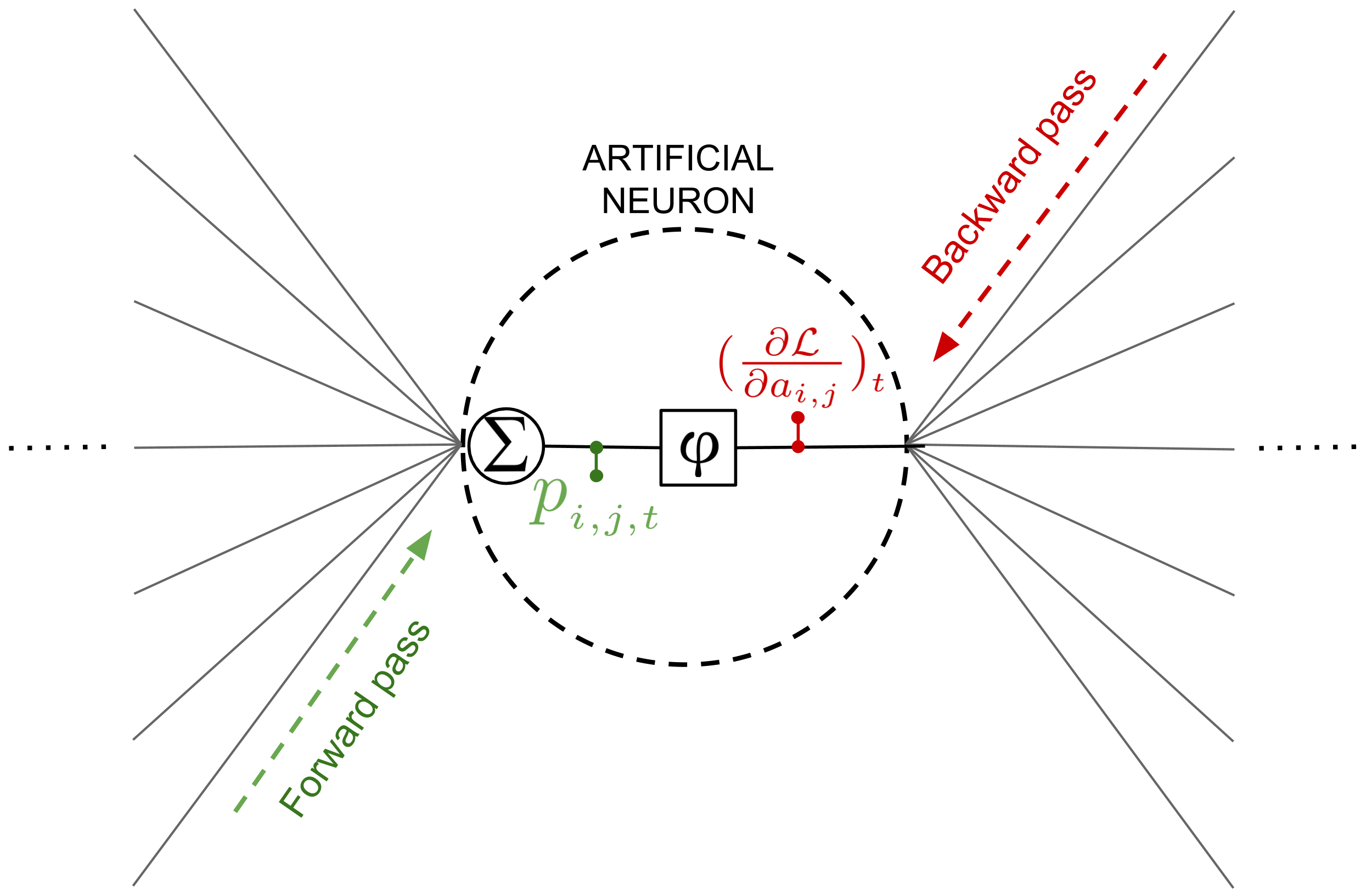}
\end{center}
\caption{Illustration of an artificial hidden neuron inside a network. $\upvarphi$ and $\mathcal{L}$ denote the activation function and the training loss respectively. The indices $i,j,t$ correspond to neuron, input example and training step indices respectively. The symbols $p_{i,j,t}$ and $a_{i,j}$ stand for the neuron's pre-activation and activation values respectively. In this paper, we monitor both the neuron's pre-activation $p_{i,j,t}$ and the partial derivative of the loss w.r.t. the neuron's activation $(\frac{\partial \mathcal{L}}{\partial a_{i,j}})_t$.}
\label{fig:neuronsignals}
\end{figure}

This chapter thus initiates the search for implicit clustering mechanisms by studying SGD from the perspective of hidden neurons. More precisely, we monitor both the pre-activations and the partial derivatives of the loss w.r.t the activations in hidden neurons. These two signals are illustrated in Figure \ref{fig:neuronsignals}. 

We study MLP networks trained on a synthetic dataset with known intraclass clusters and on a two-class version of the MNIST dataset obtained by aggregating the original classes into two superclasses. Our experiments reveal a behavior similar to the winner-take-most approach of several clustering algorithms (e.g. \citet{Martinetz1991,Fritzke1997}). Indeed, we observe that the training process progressively increases the average pre-activation of the most activated clusters of a class and decreases the average pre-activation of the least activated clusters of the same class in each neuron. Remarkably, this sometimes leads neurons to differentiate clusters belonging to the same class more strongly than clusters from different classes (cfr. Section \ref{sec:discovery}). In order to solve the classification problem, the network thus seems to apply a divide-and-conquer strategy, where different neurons specialize for the classification of different clusters of a class.

To better understand our observations, we provide an empirical investigation of the phenomenon and an intuitive explanation inspired by the Coherent Gradients Hypothesis introduced by \citet{Chatterjee2020} and the training dynamics w.r.t. example difficulty studied by \citet{Arpit2017} (cfr. Section \ref{sec:understanding}). In order to support the generality of our observations, we further show in Section \ref{sec:connections} that despite its simplicity, our setup exhibits and provides insights on many phenomena occurring in state-of-the-art models such as the regularizing effects of depth, pre-training, data augmentation, large learning rates and, importantly, the implicit clustering abilities studied in Chapter \ref{chap:implicitAbility}.

\section{Experimental setup}
\subsection{Datasets}
Our work is mainly based on a synthetic dataset whose clustered structure is exactly known. We denote this dataset by \textit{SynthClust} in the rest of the chapter. SynthClust is composed of vectors of $500$ elements, such that $\mathbf{x} \in \mathbb{R}^{500}$. Each cluster's centroid is a binary pattern with exactly five elements set to $1$. $30$ centroids with non-overlapping patterns are generated and split into two classes, such that each class contains $15$ intraclass clusters. For each cluster, $500$ training examples and $200$ testing examples are generated by adding Gaussian noise with zero mean and $0.4$ standard deviation on each of the $500$ components of the cluster's centroid. 

In order to improve the generality of our results, Section \ref{sec:discovery} also studies an MNIST variant where the first and last five digits are grouped into two distinct classes (class 0: $0\rightarrow 4$, class 1: $5\rightarrow 9$). As in Chapter \ref{chap:implicitAbility}, the five digits of a class are assumed to approximately correspond to five intraclass clusters. This constitutes an approximation, as Figure \ref{fig:intraclass_clusters} suggests that single digits are themselves composed of multiple clusters.

\subsection{Neural networks}
We train MLP networks with a single hidden layer on MNIST and SynthClust, with and without batch normalization respectively. The hidden layer is composed of $1000$ neurons without additive weights (i.e. biases). The output layer is composed of one sigmoid neuron associated to the binary cross-entropy loss. In the case of SynthClust, the model can be very simply expressed as follows:
$$f\left(\mathbf{x}\right) = \sigma \left( W_2 \cdot \text{ReLU}\left(W_1\cdot \mathbf{x} \right) + b_2 \right)$$
where ReLU and $\sigma$ denote the ReLU and sigmoid functions respectively, and $W_1$ and ($W_2,b_2$) denote the weights of the hidden and output layer respectively. In Section \ref{sec:connections}, an MLP with multiple hidden layers is trained on SynthClust. Compared to the single layer model, this multi-layer network applies batch normalization before each ReLU layer as this stabilizes training. Each hidden layer is also composed of $1000$ neurons without biases.

\subsection{Training process}
We use Layca (an SGD variant introduced in Chapter \ref{chap:implicitHyperparam}) for training, as this greatly facilitated the design and the hyperparameter tuning involved in our experiments. This design choice is more extensively discussed and investigated in Chapter \ref{chap:implicitHyperparam}. The SynthClust models are trained in full-batch mode whereas the MNIST models use a batch size of $20000$. We use large batch sizes in order to avoid the sampling noise inherent to small-batch training. While small batch sizes have been considered as a determining factor for generalization in the past \citep{Keskar2017}, this view has now been relativized by several works \citep{Hoffer2017,Goyal2017,Ginsburg2018,Geiping2021}. In our context, the use of large batch sizes did not prevent our models from exhibiting good generalization performances. We trained the models for $100$ and $400$ epochs on MNIST and SynthClust respectively, using a learning rate of $3^{-3}$ which is reduced by a factor of $5$ at epochs $85$ or $375$ respectively. The $5$ hidden layer MLP we study in Section \ref{sec:benefitsDepth} is trained for $1000$ epochs, with a reduction of the learning rate by a factor of $5$ at epoch $950$.

\section{A winner-take-most mechanism} \label{sec:discovery}
We trained the one hidden layer MLPs on SynthClust and MNIST. The resulting models reach $96.1\%$ and $98\%$ test accuracy respectively. After each training iteration, we recorded the two neuron-level training signals represented in Figure \ref{fig:neuronsignals} in $50$ hidden neurons and for each training example. After training, we selected $10$ hidden neurons amongst the $50$ monitored ones for our visualizations. This selection targets the neurons with the strongest influence on the model's predictions. More precisely, we select the neurons associated to the largest weights in the output layer (in absolute value).

\begin{sidewaysfigure}
    \includegraphics[width=\textwidth]{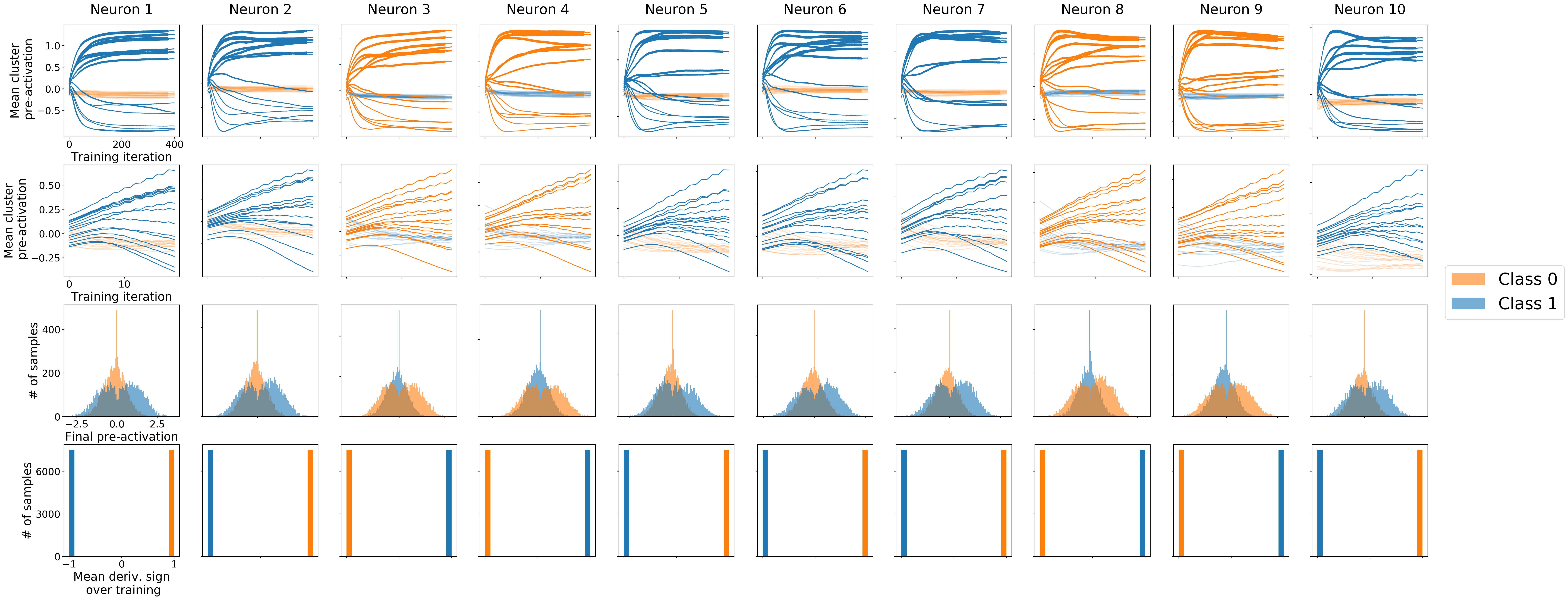}
    \caption{We train one hidden layer MLPs on SynthClust and monitor the pre-activations and the derivative of the loss w.r.t. the activations of $10$ hidden neurons after each training iteration and for each training example. The first two rows represent the evolution of each cluster's average pre-activation during training. We observe that neurons consistently differentiate the clusters of one class according to a winner-take-most mechanism: training progressively increases the average pre-activation of its most activated clusters and decreases the average pre-activation of its least activated clusters. The third row displays the resulting pre-activation distributions at the end of training. The fourth row displays a histogram of the \textit{sign} of the derivative of the loss w.r.t. the neuron activation, averaged over all training steps. This value tells us whether an increased activation generally benefits (negative average) or penalizes (positive average) the classification of the example. The results suggest that the examples of one class should always be pushed towards larger activations, and the other towards smaller activations. This seems contradictory to the observed winner-take-most mechanism. We provide insights on this apparent contradiction in Section \ref{sec:understanding}.} 
    \label{fig:synthetic_1layer}
\end{sidewaysfigure}

\begin{sidewaysfigure}
    \includegraphics[width=\textwidth]{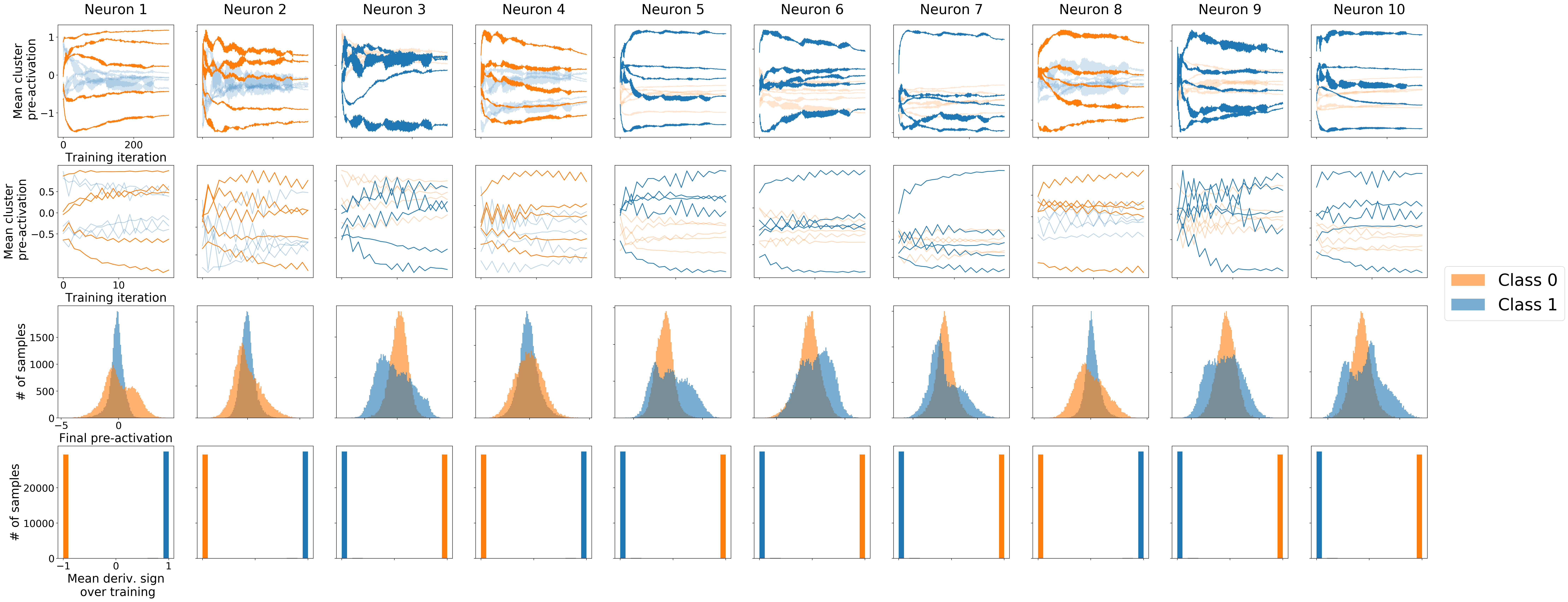}
    \caption{The same visualizations as Figure \ref{fig:synthetic_1layer}, but corresponding to training on our two-class version of the MNIST dataset (class 0: $0\rightarrow 4$, class 1: $5\rightarrow 9$). We also observe a winner-take-most mechanism that differentiates digits (i.e. the original MNIST classes) which are associated to the same superclass.}
    \label{fig:MNIST_1layer}
\end{sidewaysfigure}

Figures \ref{fig:synthetic_1layer} and \ref{fig:MNIST_1layer} display our results for both datasets. The first two rows of plots represent the evolution of each cluster's average pre-activation in the $10$ hidden neurons. The curves are colored according to the clusters' associated class. We observe that each neuron consistently differentiates the clusters of one class during training according to a winner-take-most mechanism. The clusters with larger average pre-activation are pushed towards even larger pre-activations, while the clusters with smaller average pre-activation are pushed towards even smaller pre-activations. Astonishingly, this unsupervised mechanism can be more impactful than the supervised learning process from the perspective of a single neuron. Indeed, \textit{neurons sometimes differentiate clusters belonging to the same class more strongly than clusters from different classes}.

The third row represents the histogram of the final pre-activations associated to each class. It is coherent with the first two rows: one class consistently exhibits a bimodal distribution, reflecting the differentiation of intraclass clusters. The fourth row visualizes the derivative of the loss with respect to each neuron's activation. More precisely, for each example, we compute the average \textit{sign} of the derivative across all steps of the training process. This value tells us whether an increased activation generally benefits (negative average) or penalizes (positive average) the classification of a given example. We observe that the sign is correlated with the example's class across the whole training process: the examples of one class should always be pushed towards larger activations (because their average derivative sign is $-1$), and the other to smaller activations (because their average derivative sign is $1$). We further notice that the winner-take-most mechanism always concerns the class with negative derivatives. We explore in the next section why some of the clusters of this class are pushed towards smaller pre-activations despite being associated to negative derivative signs.

\clearpage
\clearpage
\clearpage

\section{Towards understanding the mechanism} \label{sec:understanding}
To better understand why a winner-take-most mechanism emerges in our experiments, we start by performing an ablation study that identifies necessary ingredients for the phenomenon to occur. We then provide intuitions to explain the phenomenon based on difficult training examples and gradient coherence in ReLU neurons.

\subsection{An ablation study} \label{sec:ablation}
\begin{figure}[h!]
\begin{center}
\includegraphics[width=0.8\linewidth]{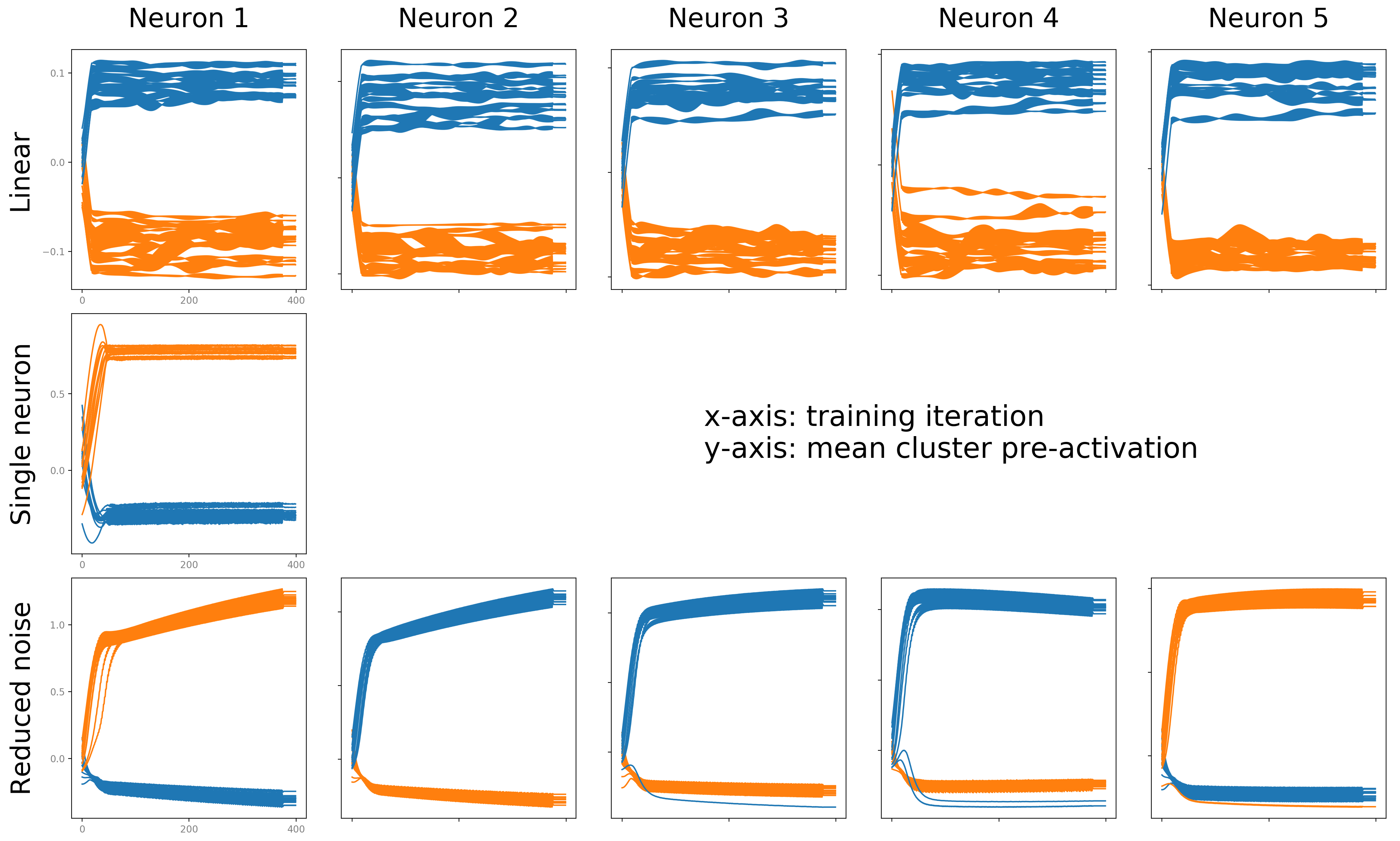}
\end{center}
\caption{We display the average pre-activation of each cluster across training for a model without ReLU activation layer (first row), with a single hidden neuron (second row) or trained on a less noisy dataset (third row). For each scenario, we observe that the winner-take-most mechanism does not occur.}
\label{fig:ablation}
\end{figure}

Figure \ref{fig:ablation} shows the average pre-activation of each cluster across training on SynthClust for a model without ReLU activation layer (first row), with a single hidden neuron (second row) and trained on a less noisy dataset\footnote{We apply Gaussian noise with a $0.1$ standard deviation instead of $0.4$ when generating the data.} (third row). For each scenario, we observe that the winner-take-most mechanism does not occur. Hidden neurons behave like the output neuron: they classify the data according to the two classes, without consideration for intraclass clusters. This results in a decrease in performance: the models achieve test accuracies ranging from $80\%$ to $84\%$. The necessity of ReLU layers, multiple hidden neurons and sufficient noise on the training examples gives rise to the intuitions we describe in the next sections.

\subsection{On the role of difficult training examples} \label{sec:difficultExamples}
At the neuron level, a mysterious force pushes some clusters in the same direction as clusters of the opposite class. This phenomenon starts around the $6^{th}$ iteration (cfr. row $2$ of Figure \ref{fig:synthetic_1layer}), and concerns the ``losing" clusters of the class subject to the winner-take-most mechanism. At first sight, this local behavior seems to be contrary to the global objective, which is to differentiate examples from their opposite class. In particular, the derivatives associated to these clusters are negative (cfr. row $4$ of Figure \ref{fig:synthetic_1layer}), aiming for the opposite direction to where they actually go.

To make sense of this apparent contradiction, we suggest considering the role of difficult training examples. Some examples can be difficult to classify because the associated noise (i) decreases their correlation with their associated class and (ii) increases their correlation with the opposite class. Therefore, these examples can lead to gradients that are contradictory with the ones of regular examples. At the beginning of training, such a contradictory force is negligible, since these examples constitute exceptions. However, once the more regular examples start being correctly classified, the share of difficult examples in the total loss increases, potentially surpassing the regular examples' share. This would lead regular examples to be pushed in a direction opposite to their associated gradient.

\begin{figure}[h!]
\begin{center}
\includegraphics[width=0.4\linewidth]{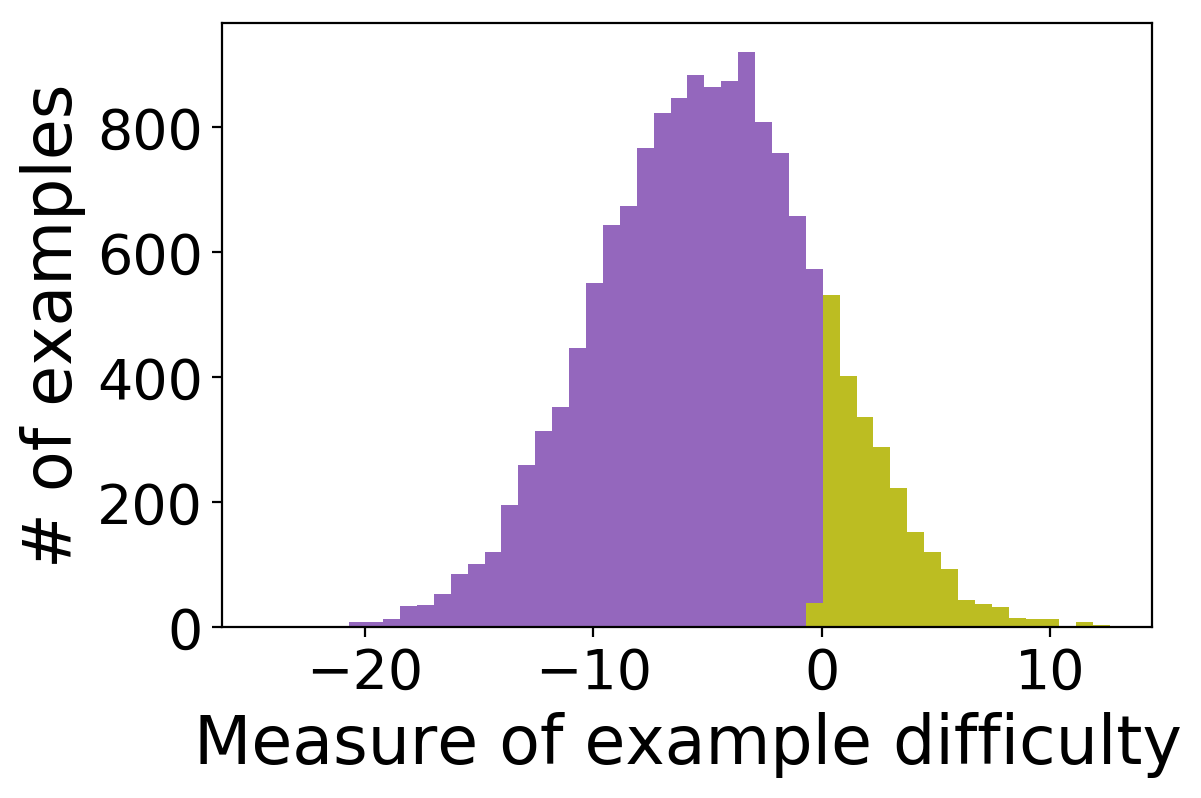} \hspace{10pt}
\includegraphics[width=0.4\linewidth]{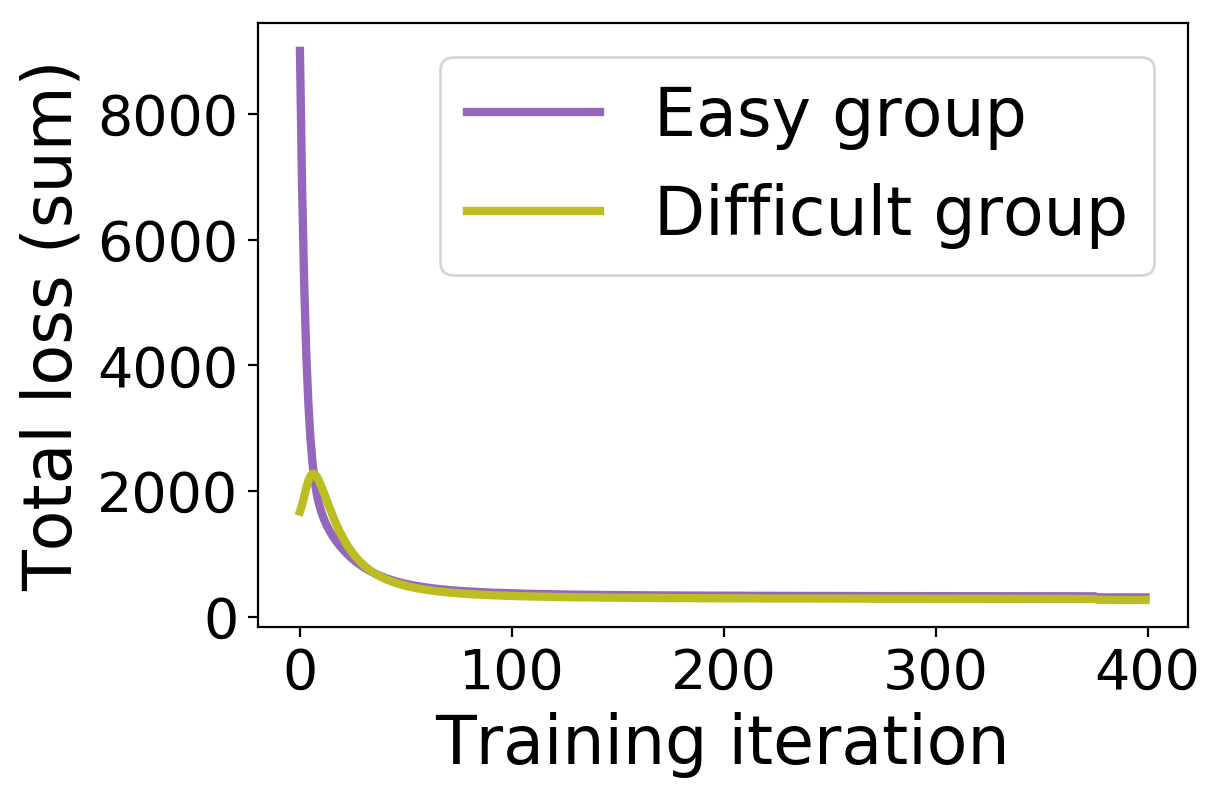}
\end{center}
\caption{\textit{Left.} We represent each class by the sum of its $15$ associated cluster centroids, and define the difficulty of an example by its scalar product with the representation of the opposite class minus its scalar product with its true class's representation. We divide training examples into easy and difficult categories depending on whether the resulting value is negative or positive. \textit{Right.} Displays the evolution of the easy and difficult groups' share in the total loss during training. We notice that (i) the loss associated to the difficult group \textit{increases} during the first iterations, indicating the occurrence of contradictory gradients and (ii) the share associated to the difficult group matches the share of the easy group around the $6^{th}$ iteration, which corresponds to the appearance of the winner-take-most mechanism (cfr. the $2^{nd}$ row of Figure \ref{fig:synthetic_1layer}).}
\label{fig:difficultExamples}
\end{figure}

We observe these exact dynamics in Figure \ref{fig:difficultExamples}. We quantify the correlation between an example and a class as the scalar product between the example and the sum of the $15$ cluster centroids associated to the class. We divide training examples into easy and difficult groups depending on whether they correlate more with their own class or with the opposite class\footnote{This can be interpreted as whether the training examples would be (in-)correctly classified by a linear classifier.}. We monitor the total loss associated to each group during training and observe that (i) the loss associated to the difficult group \textit{increases} during the first iterations, indicating the occurrence of contradictory gradients and (ii) the share associated to the difficult group matches the share of the easy group around the $6^{th}$ iteration, which corresponds to the appearance of the winner-take-most mechanism (cfr. the $2^{nd}$ row of Figure \ref{fig:synthetic_1layer}). The role of difficult training examples is further supported by our ablation study (cfr. Section \ref{sec:ablation}), which shows that reducing the noise during the data generation process, and thus the amount of difficult training examples, prevents the winner-take-most mechanism from occurring.

\subsection{On the role of ReLU} \label{sec:gradCoherence}
Since the overall gradient at a given training iteration is the sum of the per-example gradients, the directions that are coherent across multiple training examples are reinforced (as highlighted by \citet{Chatterjee2020}). At the neuron level, the ReLU activation function affects the coherence of gradients in a very specific way: since the derivative of the ReLU function is zero for negative inputs, the training examples that do not activate the neuron (i.e. have negative pre-activations) do not contribute to the gradient associated to the neuron's weights. Hence, for a given group of examples that share a common pattern, only the examples that activate the neuron reinforce each other. 

In each neuron of our single hidden layer MLP, the examples associated to each cluster share a common pattern in the input signal (by construction of SynthClust) and a common pattern in the backpropagated error signal (because they belong to the same class, cfr. row 4 of Figure \ref{fig:synthetic_1layer}). Hence, the number of examples that activate the neuron affects the relative share of each cluster in the gradient associated to the weights of the neuron. In particular, the clusters with small average pre-activations will tend to be associated with smaller shares than clusters with large average pre-activations. This could explain why clusters with smaller average pre-activations are the most impacted by the difficult training examples (cfr. Section \ref{sec:difficultExamples}), and hence ``lose" the competition. On the contrary, clusters with larger average pre-activations influence a larger share of the gradient associated to the neuron weights and will be less sensitive to difficult training examples, enabling them to ``win" the competition. The idea that ReLU layers are key for differentiating the ``winning" clusters from the ``losing" ones is also coherent with our ablation study of Section \ref{sec:ablation}, which shows that the winner-take-most mechanism does not occur in a linear model without ReLU activation layers. 

\subsection{A divide-and-conquer strategy}
Finally, our ablation study of Section \ref{sec:ablation} demonstrates that a neural network with a single hidden neuron does not exhibit a winner-take-most mechanism. This makes sense intuitively: the winner-take-most mechanism locally leads to misclassification of multiple clusters. This local misclassification is possible only if it is counter-balanced by a correct classification in other neurons. A divide-and-conquer approach, where different neurons focus on the classification of different clusters, is perfectly compatible with the winner-take-most phenomenon. Indeed, the winning clusters are determined by their initial average pre-activation, which varies from one neuron to the other because their weights are randomly initialized.

\subsection{Why does the mechanism affect a single class?}
Jointly considering the role of difficult training examples, gradient coherence and divide-and-conquer strategies also shines light on the fact that the winner-take-most mechanism only applies to the class associated to negative derivatives, i.e., whose activations should increase during training. Indeed, for this class, pushing a cluster in its ``opposite" direction \textit{simultaneously leads to a deactivation} of some of its examples. The cluster's share in this neuron's gradients is thus diminished, further promoting the correct classification of the associated \textit{difficult training examples} in this specific neuron. On the contrary, pushing clusters of the class associated to positive derivatives in their opposite direction \textit{increases} the amount of their examples that activate the neuron, promoting the correct classification of these \textit{clusters} in this specific neuron.

\section{Connections with standard deep learning settings} \label{sec:connections}
Our study discloses the emergence of a winner-take-most mechanism and provides intuitions and experiments to understand it. But these contributions are limited to relatively simple and shallow neural network architectures trained on a synthetic dataset. It is not clear whether our empirical observations and intuitions still hold in standard deep learning settings. In order to support the generality of our results, we first discuss several works that studied difficult training examples and gradient coherence in standard deep learning settings, highlighting the connections with the intuitions described in Sections \ref{sec:difficultExamples} and \ref{sec:gradCoherence}. Second, we demonstrate that our simple setup exhibits many phenomena occurring in standard settings and provide empirical evidence that these phenomena are reminiscent of winner-take-most mechanisms.

\subsection{Training dynamics w.r.t. example difficulty} \label{sec:exampleDiffStandard}
Multiple works have studied how different notions of example difficulty related to the speed at which examples are learned. \citet{Arpit2017} showed across $100$ different initializations and permutations of the training data that many examples are consistently classified (in)correctly after a single epoch of training. This observation leads them to conjecture that ``\textit{deep learning learns simple patterns first, before memorizing}". \citet{Mangalam2019} provides empirical evidence that deep neural networks learn shallow-learnable examples first, where shallow-learnable refers to being correctly classified by non-deep learning approaches. \citet{Jiang2021a} characterizes examples by their consistency score, defined by their expected accuracy as a held-out example given training sets of varying size. Figure 10 of this paper displays the training curves associated to the training examples grouped by consistency score. It reveals that examples with higher scores are learned before those with lower scores. While this aspect is not discussed in the original paper, Figure 10 also reveals that the accuracy of examples with a low consistency score \textit{decreases} in the first epochs of training. This suggests that the gradients of low-scoring examples are ``contradictory" to the ones of high-scoring examples.

In Section \ref{sec:difficultExamples}, we define the difficulty of training examples by their correlation with the opposite class relative to their correlation with their true class. In accordance with the observations conducted in standard settings, we observe in Figure \ref{sec:difficultExamples} that in our simple setup, (i) easy examples are learned before the difficult ones and (ii) the loss of difficult examples \textit{increases} in the first training iterations, suggesting the presence of contradictory gradients.

\subsection{The Coherent Gradient Hypothesis} \label{sec:CGH}
\citet{Chatterjee2020} recently introduced the Coherent Gradient Hypothesis, which states that gradient coherence plays a crucial role in the generalization abilities of deep neural networks. \citet{Zielinski2020} provides multiple experiments to support this hypothesis in the context of standard deep learning settings involving the ImageNet dataset and ResNet models with $18$ layers. These works justify the role of gradient coherence with the following intuition: because gradients are the sum of per-example gradients, it is stronger in the directions where the per-example gradients are more similar. Hence, the changes to the network during training are biased towards those that simultaneously benefit multiple examples. They further argue that such bias is beneficial for generalization based on algorithmic stability theory.

However, the previous intuition only holds at the very beginning of training, when most examples are misclassified by the model. As we've seen in Section \ref{sec:difficultExamples}, a small set of difficult training examples strongly influence the overall gradient once regular examples are correctly classified. \citet{Chatterjee2020a} provide a more extensive analysis of the evolution of gradient coherence during training. They conclude the following: ``\textit{our experiments provide additional evidence for the connection between the alignment of per-example gradients and generalization. But as our data shows this connection is complicated.}" We believe that the winner-take-most mechanisms disclosed by our work and the intuitions described in Section \ref{sec:gradCoherence} offer a promising path towards a better understanding the relationship between gradient coherence and generalization.

\subsection{The benefits of data augmentation} \label{sec:benefitsDataAugm}
\textbf{Observations in standard settings.} Data augmentation is a key component of state-of-the-art models \citep{Cubuk2019}. These techniques are motivated by the fact that applying \textit{plausible} transformations on the training data virtually increases the amount of data available for learning. Surprisingly, data augmentation techniques that apply \textit{unrealistic} transformations, such as Cutout \citep{DeVries2017} and Mixup \citep{Zhang2018} appear to be quite effective for regularizing deep neural networks as well.

\textbf{Observation in our simple setup and connection with the winner-take-most mechanisms.} We trained the single layer MLP on a reduced SynthClust dataset: $1500$ examples are randomly selected for training (instead of $15000$). As expected, training on less data resulted in a decreased test accuracy: the model reaches $80.98\%$ accuracy instead of the $96.1\%$ obtained when training on the complete dataset. We applied Dropout \citep{Srivastava2014} on the inputs of the network to augment the training data. More precisely, we randomly set input components to $0$ with a $50\%$ probability. While this transformation is fundamentally different from the Gaussian noise inherent to the data generation process, Dropout provided a huge gain in terms of test accuracy, reaching $88.57\%$. 

\begin{figure}[h!]
\begin{center}
\includegraphics[width=0.95\linewidth]{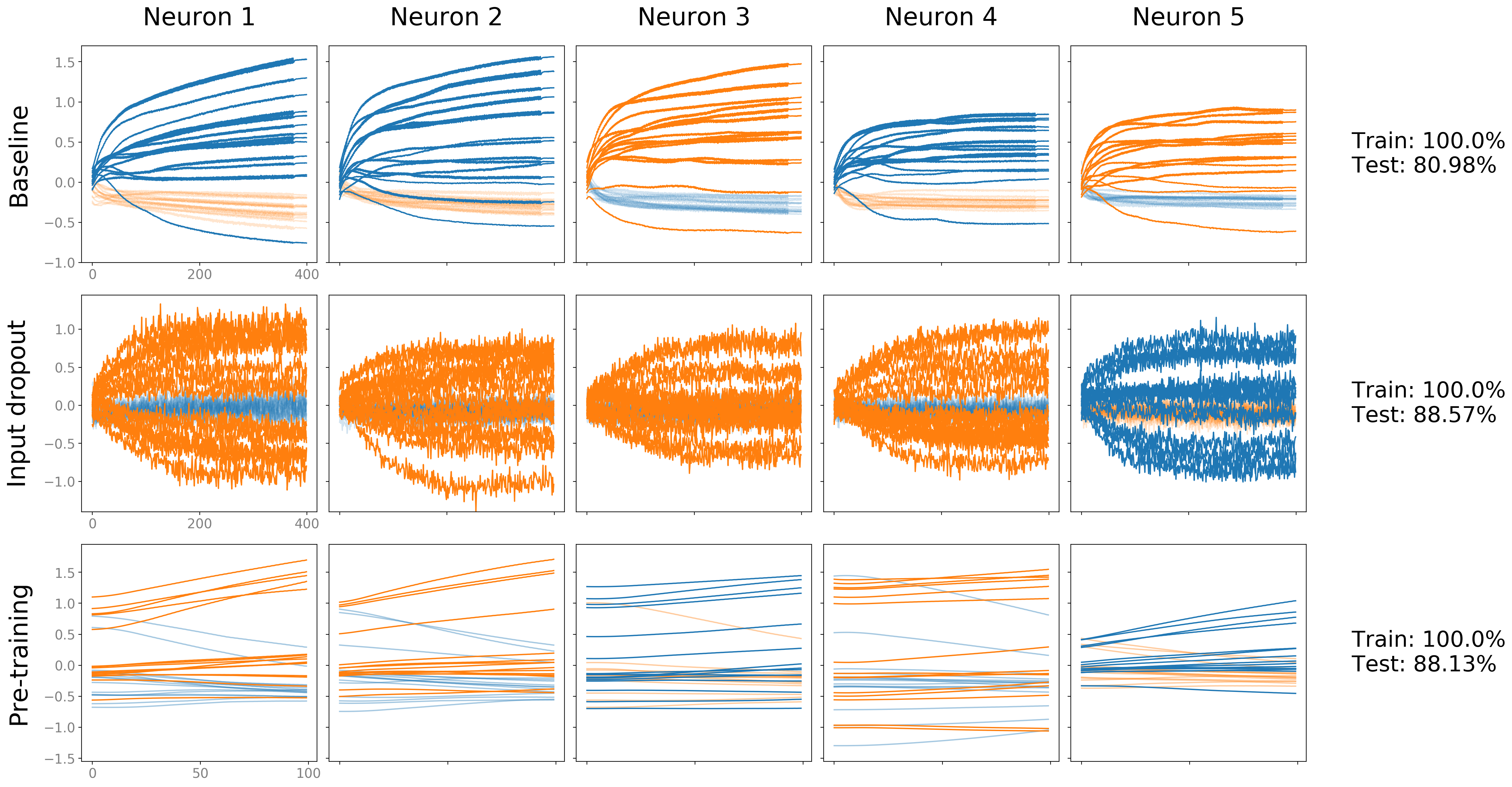}
\end{center}
\caption{The firs row displays the average pre-activation curves of each cluster (cfr. $1^{st}$ line of Figure \ref{fig:synthetic_1layer}) when training the single hidden layer MLP on a reduced SynthClust dataset ($1500$ training examples instead of $15000$). We observe that the strength of the winner-take-most mechanism decreases: clusters from the same class are less differentiated compared to clusters from different classes. Rows $2$ and $3$ provide the same visualization when applying Dropout to the inputs of the network (row $2$) or when pre-training the network on a different dataset exhibiting the same clusters and dataset size as the original SynthClust dataset but different cluster-class associations (row $3$). Both lead to stronger differentiation of intraclass clusters and improved test accuracies.}
\label{fig:connections}
\end{figure}

The first row of Figure \ref{fig:connections} displays the average pre-activation curves of each cluster for the model trained without Dropout. The visualization reveals that training on a reduced training dataset decreases the strength of the winner-take-most mechanism: clusters from the same class are less differentiated compared to clusters from different classes. Interestingly, this issue gets mitigated by the application of Dropout, as revealed by the second row of Figure \ref{fig:connections}. This observation suggests that data augmentation techniques improve generalization by generating difficult training examples, promoting the occurrence of winner-take-most mechanisms (cfr. Section \ref{sec:difficultExamples}) and improving the model's clustering abilities (cfr. Chapter \ref{chap:implicitAbility}).

\subsection{The benefits of pre-training} \label{sec:benefitsPretrain}
\textbf{Observations in standard settings.} Pre-training is a long-standing technique in deep learning \citep{Hinton2006,Oquab2014,Donahue2014}. It consists in training a network on a related task for which large amounts of data are available and fine-tuning the resulting model on the target task. It can be interpreted as a parameter initialization strategy for SGD training. Surprisingly, researchers observed empirically that pre-training is very effective, even when the pre-training tasks and target tasks are quite different. For example, contrastive learning techniques use unsupervised pretext tasks to pre-train supervised image classification networks and recently gained a lot of popularity \citep{He2020,Zhao2021a}.

\textbf{Observation in our simple setup and connection with the winner-take-most mechanisms.} We pre-trained the single layer MLP on a dataset containing the same input data as the original SynthClust dataset -and hence the same clusters and number of training examples-, but different, randomly generated cluster-class associations. We then fine-tuned the model on the reduced SynthClust dataset introduced in Section \ref{sec:benefitsDataAugm}. Despite both classification tasks being different, we observe an improvement in test accuracy compared to no pre-training: the model reaches an accuracy of $88.13\%$ instead of $80.98\%$. The study of the cluster's average pre-activation curves in Figure \ref{fig:connections} reveals that because both pre-training and target datasets share the same clustered structure, the fine-tuning process benefits from the winner-take-most mechanisms that occurred during pre-training. Indeed, clusters from the same class are already strongly differentiated at initialization.

\subsection{The benefits of depth} \label{sec:benefitsDepth}
\textbf{Observations in standard settings.} State-of-the-art deep learning models contain many hidden layers. In the context of image classification, the amount of layers typically ranges from $16$ to more than a hundred \citep{Simonyan2014,He2016}. Many works provide results concerning the benefits of depth in terms of expressivity (e.g., \citet{Telgarsky2016,Eldan2016,Lin2017,Liang2017}). Its benefits in terms of generalization ability, however, remain unexplained.

\begin{figure}[h!]
\begin{center}
\includegraphics[width=0.85\linewidth]{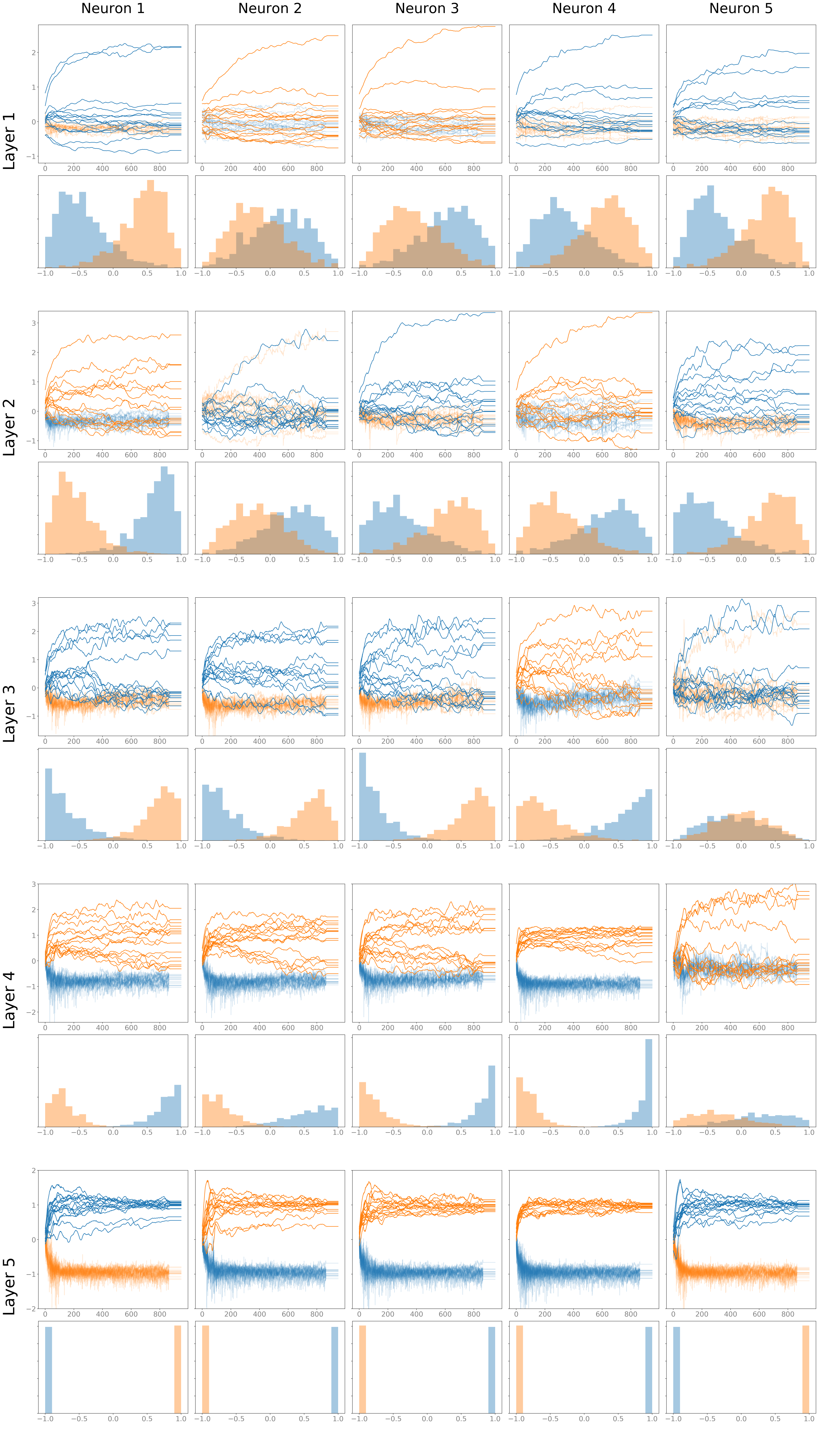}
\end{center}
\caption{Discussed in section \ref{sec:benefitsDepth}}
\label{fig:synthetic_5layers}
\end{figure}

\textbf{Observation in our simple setup and connection with the winner-take-most mechanisms.} We trained an MLP with $5$ hidden layers on SynthClust. Despite the simplicity of the SynthClust dataset, using a deeper neural network improved the test accuracy ($97.68\%$ test accuracy compared to $96.1\%$). In order to study the impact of depth on the emergence of winner-take-most mechanisms, Figure \ref{fig:synthetic_5layers} displays the average pre-activation curves of each cluster and the histograms of the average derivative signs (cfr. rows $1$ and $4$ of Figure \ref{fig:synthetic_1layer}) in $5$ neurons of each hidden layer\footnote{The $5$ neurons are selected based on the norm of the neurons' associated weight vector in the next layer (the larger the norm, the better).}. 

The visualizations reveal that winner-take-most mechanisms occur in the first three layers of the network as well as in some neurons of the $4^{th}$ layer. Interestingly, the mechanism leads \textit{a single cluster} to be differentiated from the other examples of the dataset in multiple neurons (e.g., neurons $3$ and $4$ of layer $2$). This change in behavior compared to the single layer MLP studied in Figure \ref{fig:synthetic_1layer} could be induced by the fact that derivative signs correlate less with the examples' class as they propagate through layers (cfr. the histograms of average derivative signs in Figure \ref{fig:synthetic_5layers}). The differentiation of \textit{single} clusters by hidden neurons could improve the network's clustering abilities, offering an interesting research direction for explaining the benefits of depth in terms of generalization. 

\clearpage
\clearpage

\subsection{The benefits of large learning rates} \label{sec:syntheticLargeLr}
\textbf{Observations in standard settings.} The influence of SGD's learning rate on generalization has been highlighted by many works \citep{Jastrz2017,SmithSam2017,Smith2017,Hoffer2017,masters2018revisiting}. Empirically, we observe that using larger learning rates benefits generalization -as long as convergence remains possible.

\begin{figure}
\begin{center}
\includegraphics[width=0.95\linewidth]{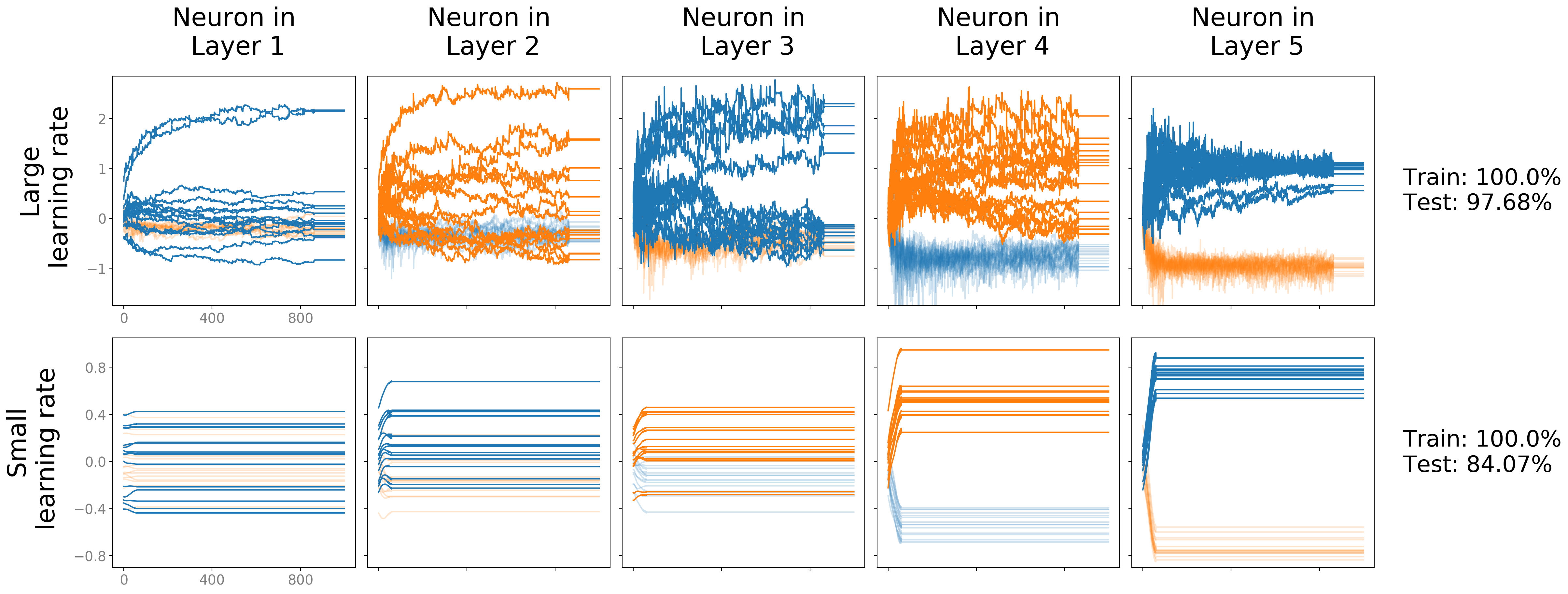}
\end{center}
\caption{Test accuracies and the average pre-activation curves of each cluster when training an MLP with $5$ hidden layers on SynthClust with large (row $1$) and small (row $2$) learning rates. The average pre-activation curves correspond to $1$ neuron in each layer, selected based on the norm of its outgoing weights. We observe that (i) large learning rates drastically improve generalization in our simple setup and (ii) small learning rates induce negligible changes to the cluster's average pre-activations in the first layers and thus no winner-take-most mechanisms.}
\label{fig:synthetic_LargeLr}
\end{figure}

\textbf{Observation in our simple setup and connection with the winner-take-most mechanisms.} We trained the 5-hidden-layers MLP on SynthClust with a learning rate reduced by a factor $27$. The test accuracy strongly decreased: the model reaches an accuracy of $84.07\%$ instead of $97.68\%$. We display the average pre-activation curves corresponding to training with large and small learning rates in Figure \ref{fig:synthetic_LargeLr} for $1$ neuron in each layer. As usual, the neuron is selected based on the norm of the associated weights in the next layer. We observe that when using small learning rates, the cluster's average pre-activations do not change much at all in the first layers, leading to the absence of winner-take-most mechanisms. On the contrary, training with large learning rates leads to significant changes in the cluster's average pre-activations, enabling the emergence of winner-take-most mechanisms.

\subsection{The benefits of implicit clustering abilities} \label{sec:benefitsAbilities}
\textbf{Observations in standard settings.} In Chapter \ref{chap:implicitAbility}, we show that five tentative measures of intraclass clustering correlate with generalization in standard deep learning settings. These correlations occur across variations of $8$ standard hyperparameters, amongst which data augmentation, depth and the learning rate. Two measures ($c_1$ and $c_3$ defined in Section \ref{sec:measures}) are applied at the neuron-level, capturing the extent by which examples or subclasses from the same class are differentiated in a neuron's pre-activations.

\textbf{Observation in our simple setup and connection with the winner-take-most mechanisms.} The winner-take-most mechanism studied in our simple setup leads to the differentiation of clusters from the same class in a neuron's pre-activations, and is thus closely related to measures $c_1$ and $c_3$. We further show that data augmentation, depth and learning rate influence the winner-take-most mechanism and the test accuracies of the studied MLP networks in Sections \ref{sec:benefitsDataAugm}, \ref{sec:benefitsDepth} and \ref{sec:syntheticLargeLr}. The observations are consistent with our experiments of Chapter \ref{chap:implicitAbility}: the better the clusters of a class are differentiated, the better the performance on the test set. Hence, our studies of both implicit clustering abilities and mechanisms constitute a coherent framework supporting the crucial role of implicit clustering in deep learning systems.
\chapter{An implicit clustering hyperparameter}
\label{chap:implicitHyperparam}
Designing and training deep learning systems requires manual tuning of many hyperparameters (network depth, width, learning rate, weight decay, optimizer, etc.). Hyperparameter tuning is usually based on successful heuristics and intuitions that practitioners gain with experience. In practice, this still translates into a lot of trial and error, greatly increasing the time and energy consumption associated to the development of deep learning systems. 

The opaque behaviour of standard hyperparameters becomes less surprising when one presumes that implicit mechanisms play a role in deep learning. Indeed, these mechanisms could be indirectly affected by the explicit hyperparameters in potentially complex and coupled ways. Chapter \ref{chap:implicitAbility} suggests that $8$ standard hyperparameters indirectly influence the implicit clustering abilities of deep learning. Hence, the following chapter looks for implicit hyperparameters that control clustering more directly. 

Chapters \ref{chap:implicitAbility} and \ref{chap:implicitMechanism} suggest the emergence of a neuron-level training process that is critical for implicit clustering to occur. However, the training process of the entire network might succeed without fully accomplishing each individual neuron's training, as suggested by our observations in Section \ref{sec:syntheticLargeLr}. Hence, the extent by which each neuron has been ``trained" potentially constitutes an implicit clustering hyperparameter which we propose to capture through \textit{layer rotation}, i.e., the evolution across training of the cosine distance between each layer's flattened weight vector and its initialization. Monitoring the \textit{rotation} of weight vectors is motivated by the fact that batch normalization renders the scale of a layer's transformation obsolete. Monitoring training on a \textit{per-layer} basis is motivated by the works on vanishing and exploding gradients, which suggest that the training dynamics can vary drastically across the layers of a network \citep{Bengio1994,Hochreiter1998,Glorot2010}.

We design tools to monitor and control layer rotations and show across a diverse set of experiments that larger layer rotations (and thus presumably more accomplished neuron-level training routines) consistently translate into better generalization. Moreover, we show that the impact of learning rates, weight decay, learning rate warmups and adaptive gradient methods on generalization and training speed seems to result from their indirect influence on layer rotations. Finally, we illustrate on a single hidden layer MLP trained on MNIST that layer rotation correlates with the degree to which the features of individual neurons have been learned, connecting our results with our initial hypothesis.  An implementation of this chapter's tools and experiments based on Tensorflow \citep{Agarwal2016} and Keras \citep{chollet2015keras} is available at \url{https://github.com/ispgroupucl/layer-rotation-tools} and \url{https://github.com/ispgroupucl/layer-rotation-paper-experiments} respectively.

\section{Tools for monitoring and controlling layer rotation} \label{sec:tools}
This section describes the tools for monitoring and controlling layer rotation during training, such that its relationship with generalization, training speed and explicit hyperparameters can be studied in Sections \ref{sec:Exploration} and \ref{sec:CommonPracticesAnalysis}.

\subsection{Monitoring layer rotation with layer rotation curves} \label{sec:monitoring}
Layer rotation is defined as the evolution of the cosine distance between each layer's weight vector and its initialization during training. The cosine distance is defined as:
\begin{equation} \label{eq:cosdis}
d(u,v) = 1- \frac{u\cdot v}{\parallel u\parallel_2 \parallel v\parallel_2}.
\end{equation}
Let $w_l^t$ be the flattened weight tensor of the $l^{th}$ layer at optimization step $t$ ($t_0$ corresponding to initialization), then the rotation of layer $l$ at training step $t$ is defined as $d(w_l^{t_0},w_l^{t})$\footnote{It is worth noting that our study focuses on weights that multiply the inputs of a layer (\textit{e.g.} kernels of fully connected and convolutional layers). Studying the training of additive weights (biases) is left as future work.}. In order to visualize the evolution of layer rotation during training, we plot how the cosine distance between each layer's current weight vector and its initialization evolves across training steps. We denote this visualization tool by \textit{layer rotation curves} hereafter.

\subsection{Controlling layer rotation with Layca} \label{sec:layca}
The ability to control layer rotations during training would enable a systematic study of their relationship with generalization and training speed. Therefore, we present Layca (LAYer-level Controlled Amount of weight rotation), an algorithm where the layer-wise learning rates directly determine the amount of rotation performed by each layer's weight vector during each training step (the \textit{layer rotation rates}), in a direction specified by an optimizer (SGD being the default choice). Inspired by techniques for optimization on manifolds \citep{absil2010}, and on spheres in particular, Layca applies layer-wise orthogonal projection and normalization operations on SGD's updates, as detailed in Algorithm \ref{alg:layca}. These operations induce the following simple relationship between the learning rate $\rho_l (t)$ of layer $l$ at training step $t$ and the angle $\theta_l (t)$ between $w_l^t$ and $w_l^{t-1}$: $\rho_l (t) = tan(\theta_l (t))$. 

Our controlling tool is based on a strong assumption: that controlling the amount of rotation performed during each individual training step (i.e. the layer rotation rate) enables control of the cumulative amount of rotation performed since the start of training (i.e. layer rotation). This assumption is not trivial since the aggregated rotation is a priori very dependent on the shape of the loss landscape. For example, for an identical layer rotation rate, the layer rotation will be much smaller if iterates oscillate around a minimum instead of following a stable downward slope. Our assumption however appeared to be sufficiently valid, and the control of layer rotation was effective in our experiments.

\begin{algorithm*}
   \caption{Layca, an algorithm that enables control over the amount of weight rotation per step for each layer through its learning rate parameter (cfr. Section \ref{sec:layca}).}
   \label{alg:layca}
\begin{algorithmic}
   \State {\bfseries Require:} $o$, an optimizer (SGD is the default choice)
   \State {\bfseries Require:} $T$, the number of training steps
   \State $L$ is the number of layers in the network
   \For{$l=0$ {\bfseries to} $L-1$}
   \State {\bfseries Require:} $\rho _l (t)$, a layer's learning rate schedule
   \State {\bfseries Require:} $w^l_0$, the initial multiplicative weights of layer $l$
   \EndFor
   \For{$t=0$ {\bfseries to} $T$}
   \State $s^0_t,..., s^{L-1}_t  = \text{getStep}(o,w^0_t,..., w^{L-1}_t)$ \noindent\hspace{6pt} (get the updates of the selected optimizer)
   \For{$l=0$ {\bfseries to} $L-1$}
   \State $s^l_t \leftarrow s^l_t - \frac{(s^l_t \cdot w^l_t) w^l_t}{w^l_t \cdot w^l_t}$ \noindent\hspace{24pt} (project step on space orthogonal to $w^l_t$)
   \State $s^l_t \leftarrow \frac{s^l_t \parallel w^l_t\parallel _2}{\parallel s^l_t\parallel _2}$                 \noindent\hspace{50pt} (rotation-based normalization)
   \State $w^l_{t+1} \leftarrow w^l_t + \rho_l(t) s^l_t$ \noindent\hspace{18pt} (perform update)
   \State $w^l_{t+1} \leftarrow w^l_{t+1} \frac{\parallel w^l_{0}\parallel _2}{\parallel w^l_{t+1}\parallel _2}$ \noindent\hspace{16pt} (project weights back on sphere)
   \EndFor
   \EndFor
\end{algorithmic}
\end{algorithm*}

\newpage
\section{Experimental setup}\label{sec:LaycaExperimentalSetup}
The experiments are conducted on five different tasks which vary in network architecture and dataset complexity, and are further described in Table \ref{tab:experiments}.

\begin{table*}[!h]
  \caption{Summary of the tasks used for our experiments\protect\footnotemark}
  \label{tab:experiments}
  \centering
  \resizebox{\textwidth}{!}{
  \begin{tabular}{lll}
    \toprule
    Name     & Architecture     & Dataset \\
    \midrule
    C10-CNN1     & VGG-style 25 layers deep CNN  & CIFAR-10    \\
    C100-resnet  & ResNet-32  & CIFAR-100      \\
    tiny-CNN     & VGG-style 11 layers deep CNN       & Tiny ImageNet  \\
    C10-CNN2     & deep CNN from torch blog         & CIFAR-10 + data augm.  \\
    C100-WRN     & Wide ResNet 28-10 with 0.3 dropout        & CIFAR-100 + data augm.  \\
    \bottomrule
  \end{tabular}}
\end{table*}
\footnotetext{References: VGG \citep{Simonyan2014}, ResNet \citep{He2016}, torch blog \citep{torchCNN}, Wide ResNet \citep{Zagoruyko2016}, CIFAR-10 \citep{Krizhevsky2009}, Tiny ImageNet \citep{Deng2009,CS231N}. Dropout layers were removed from the torch blog CNN to enable perfect classification on the training set ($100\%$ accuracy).}

We use the same amount of training epochs and the same learning rate decay scheme across layer rotation and explicit hyperparameter configurations:\\
\begin{itemize}
\item C10-CNN1: 100 epochs and a reduction of the learning rate by a factor 5 at epochs 80, 90 and 97
\item C100-resnet: 100 epochs and a reduction of the learning rate by a factor 10 at epochs 70, 90 and 97
\item tiny-CNN: 80 epochs and a reduction of the learning rate by a factor 5 at epoch 70
\item C10-CNN2: 250 epochs and a reduction of the learning rate by a factor 5 at epochs 100, 170, 220
\item C100-WRN: 250 epochs and a reduction of the learning rate by a factor 5 at epochs 100, 170, 220
\end{itemize}
The only exceptions are C10-CNN2 and C100-WRN trained with SGD+weight decay and with adaptive methods (cfr. Sections \ref{sec:WDanalysis} and \ref{sec:ADGanalysis}), where the learning rate decay schemes are the ones used in their original implementation or in \citep{Wilson2017} respectively. \textit{Training} accuracy is close to optimal in most cases, as revealed by the Tables \ref{tab:train_exploration}, \ref{tab:train_lr}, \ref{tab:train_SGD}, \ref{tab:train_warmup} and \ref{tab:train_AGM}.

\begin{table*}[!h]
  \caption{Train accuracies of models used in Figure \ref{fig:exploration_curves}}
  \label{tab:train_exploration}
  \centering
  \resizebox{\textwidth}{!}{
  \begin{tabular}{l|ccccc|}
     & $\alpha = 0.6$     & $\alpha = -0.6$  & $\rho (0) = 3^{-5}$ &  $\rho (0) = 3^{-4}$ & Best\\ \hline 
    C10-CNN1     & \rule{0pt}{2.6ex}$100\%$  & $99.99\%$   & $100\%$ & $100\%$ &  $99.99\%$ \\[3pt]
    C100-resnet  & $82.09\%$  & $99.54\%$ &$99.87\%$ & $99.99\%$  &$99.75\%$  \\[3pt]
    tiny-CNN     & $99.98\%$ & $99.95\%$ &$99.97\%$ &$99.97\%$ &$98.91\%$ \\[3pt]
    C10-CNN2  & $100\%$  & $99.94\%$ &$99.99\%$ & $99.99\%$  &$99.97\%$  \\[3pt]
    C100-WRN  & $99.88\%$  & $99.91\%$ &$99.97\%$ & $99.99\%$  &$99.96\%$  \\[2pt]
    \hline
  \end{tabular}}
\end{table*}

\begin{table*}[!h]
  \caption{Train accuracies of models used in Figure \ref{fig:lr_analysis}}
  \label{tab:train_lr}
  \centering
  \resizebox{\textwidth}{!}{
  \begin{tabular}{l|ccccc|}
     & $lr = 3^{-4}$     & $lr = 3^{-3}$  & $lr = 3^{-2}$ &  $lr = 3^{-1}$ & $lr = 3^{0}$\\ \hline 
    C10-CNN1     & \rule{0pt}{2.6ex}$100\%$  & $100\%$   & $100\%$ & $100\%$ &  $100\%$ \\[3pt]
    C100-resnet  & $87.8\%$  & $100\%$ &$100\%$ & $100\%$  &$99.7\%$  \\[3pt]
    tiny-CNN     & $100\%$ & $100\%$ &$100\%$ &$100\%$ &$100\%$ \\[3pt]
    C10-CNN2  & $99.8\%$  & $99.9\%$ &$100\%$ & $100\%$  &$83.7\%$  \\[3pt]
    C100-WRN  & $100\%$  & $100\%$ &$100\%$ & $100\%$  &$57.4\%$  \\[2pt]
    \hline
  \end{tabular}}
\end{table*}

\begin{table*}[!h]
  \caption{Train accuracies of models used in Figure \ref{fig:SGD_analysis}}
  \label{tab:train_SGD}
  \centering
  \resizebox{\textwidth}{!}{
  \begin{tabular}{l|ccccc|}
     & C10-CNN1 & C100-resnet  & tiny-CNN &  C10-CNN2 & C100-WRN \\ \hline
    SGD + $L_2$    & \rule{0pt}{2.6ex}$100\%$  & $100\%$   & $100\%$ & $100\%$ &  $100\%$ \\[2pt]
    \hline
  \end{tabular}}
\end{table*}

\begin{table*}[!h]
  \caption{Train accuracies of models used in Figure \ref{fig:warmup_analysis}}
  \label{tab:train_warmup}
  \centering
  \resizebox{\textwidth}{!}{
  \begin{tabular}{|ccccc|}
     No warmup & 5 epochs  & 10 epochs &  15 epochs & Layca-No warmup \\ \hline
        \rule{0pt}{2.6ex}$96.67\%$  & $99.76\%$   & $99.85\%$ & $99.68\%$ &  $99.85\%$ \\[2pt]
    \hline
  \end{tabular}}
\end{table*}

\begin{table*}[!h]
  \caption{Train accuracies of models used in Figure \ref{fig:AGM_analysis_curves}}
  \label{tab:train_AGM}
  \centering
  \resizebox{\textwidth}{!}{
  \begin{tabular}{l|ccccc|}
     & C10-CNN1 & C100-resnet  & tiny-CNN &  C10-CNN2 & C100-WRN \\ \hline 
    Adaptive methods    & \rule{0pt}{2.6ex}$100\%$  & $100\%$   & $100\%$ & $100\%$ &  $99.9\%$ \\[3pt]
    Adaptive + Layca  & $100\%$  & $99.7\%$ &$99.2\%$ & $100\%$  &$100\%$ \\[2pt]
    \hline
  \end{tabular}}
\end{table*}

\section{A systematic study of layer rotation configurations} \label{sec:Exploration}
Section \ref{sec:tools} provides tools to monitor and control layer rotation. The purpose of this section is to use these tools to conduct a systematic experimental study of layer rotation configurations. We adopt SGD as default optimizer, but use Layca (cfr. Algorithm \ref{alg:layca}) to vary the relative rotation rates (faster rotation for first layers, last layers, or no prioritization) and the global rotation rate value (high or low rate, for all layers). 

\subsection{Layer rotation rate configurations}
Layca enables us to specify layer rotation rate configurations, i.e. the amount of rotation performed by each layer's weight vector during one optimization step, by setting the layer-wise learning rates. To explore the large space of possible layer rotation rate configurations, our study restricts itself to two directions of variation. First, we vary the initial global learning rate $\rho (0)$, which affects the layer rotation rate of all the layers. During training, the global learning rate $\rho (t)$ drops following a fixed decay scheme, hence the dependence on $t$. The second direction of variation tunes the relative rotation rates between different layers. More precisely, we apply static, layer-wise learning rate multipliers that exponentially increase/decrease with layer depth (which is typically encountered with exploding/vanishing gradients, cfr. \citet{Bengio1994,Hochreiter1998,Glorot2010}). The multipliers are parametrized by the layer index $l$ (in forward pass ordering) and a parameter $\alpha \in [-1,1]$ such that the learning rate of layer $l$ becomes:
\begin{equation} \label{eq:alpha}
\rho_l(t) =
\left\lbrace
\begin{array}{lll}
(1-\alpha)^{5\frac{(L-1-l)}{L-1}} \rho (t)  & \mbox{if} & \alpha>0\\
(1+\alpha)^{5\frac{l}{L-1}} \rho (t) & \mbox{if} & \alpha \leq 0
\end{array}\right.
\end{equation}
Values of $\alpha$ close to $-1$ correspond to faster rotation of first layers, $0$ corresponds to uniform rotation rates, and values close to $1$ to faster rotation of last layers. 

\subsection{Layer rotation's relationship with generalization} \label{subsec:generalization}
\begin{figure}[!h]
\begin{center}
\includegraphics[width=0.92\linewidth]{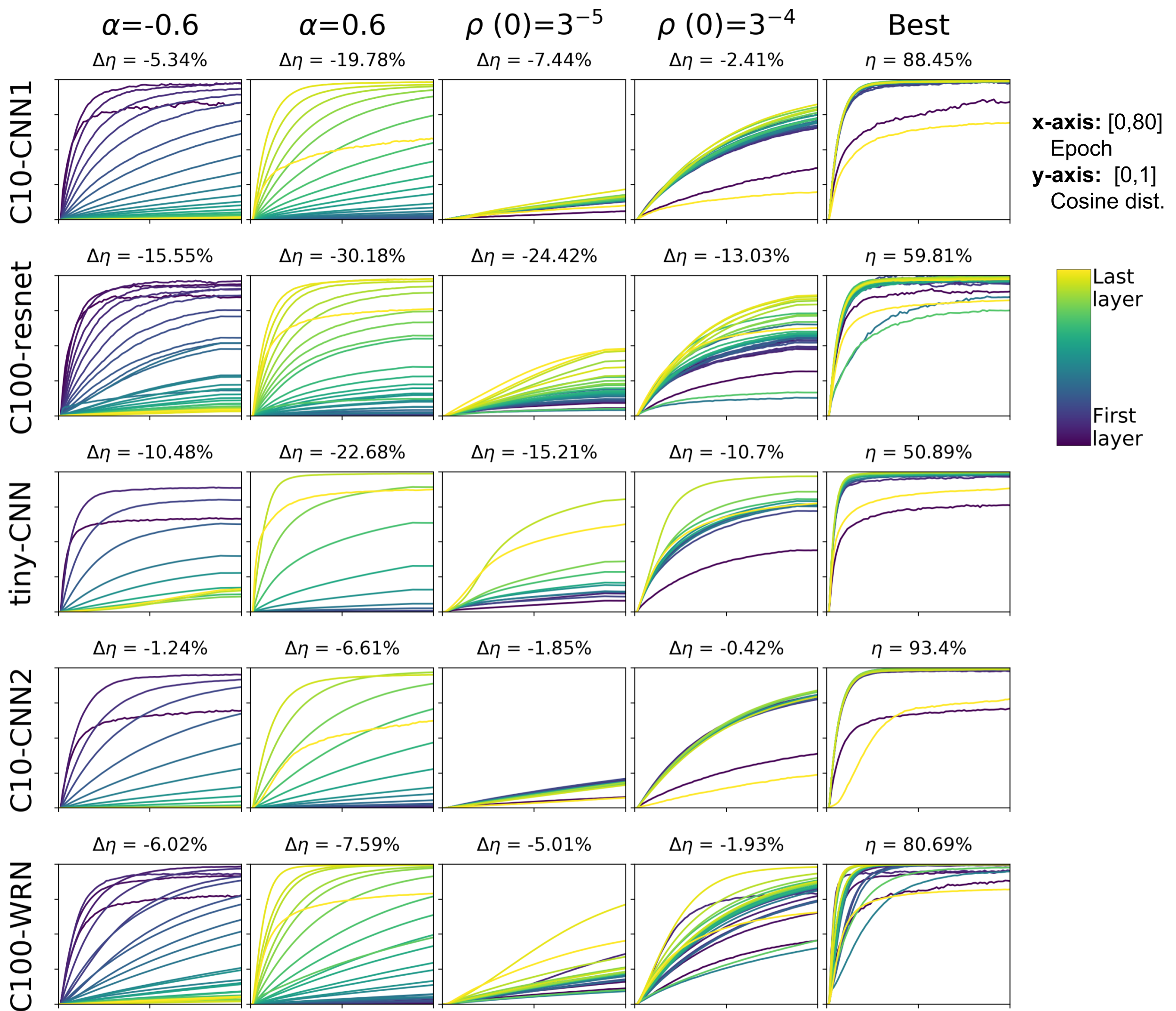}
\end{center}
\caption{Analysis of the layer rotation curves (cfr. Section \ref{sec:monitoring}) and test accuracies ($\eta$) induced by different layer rotation rate configurations (using Layca for training) on the five tasks of Table \ref{tab:experiments}. The configurations are parametrized by $\alpha$, that controls which layers have the highest rotation rates (first layers for $\alpha <0$, last layers for $\alpha >0$, or no prioritization for $\alpha=0$), and $\rho (0)$, the initial global rotation rate value shared by all layers. $\Delta\eta$ is computed with respect to the best configuration (last column), which corresponds to $\alpha = 0$ and $\rho (0) = 3^{-3}$ for the five tasks. This visualization unveils large differences in generalization ability across configurations which seem to follow a simple yet consistent rule of thumb: the larger the layer rotation for each layer, the better the generalization performance. Training accuracies are provided in Section \ref{sec:LaycaExperimentalSetup} ($\approx 100\% $ in all configurations).}
\label{fig:exploration_curves}
\end{figure}

Figure \ref{fig:exploration_curves} depicts the layer rotation curves (cfr. Section \ref{sec:monitoring}) and the corresponding test accuracies obtained with different layer rotation rate configurations. While each configuration achieves $\approx 100\%$ training accuracy (cfr. Section \ref{sec:LaycaExperimentalSetup}), we observe huge differences in generalization ability (differences of up to $30\%$ test accuracy). These differences seem to be tightly connected to differences in layer rotations. In particular, we extract the following rule of thumb that is applicable across the five considered tasks: the larger the layer rotations, the better the generalization performance. The best performance is consistently obtained when nearly all layers reach a cosine distance of $1$, which corresponds to final weights that are orthogonal to their initialization (cfr. fifth column of Figure \ref{fig:exploration_curves}). This observation would have limited value if many configurations (amongst which the best one) lead to cosine distances of $1$. However, we notice that most configurations do not. In particular, rotating the layers weights very slightly is sufficient for the network to achieve $100\%$ training accuracy (cfr. third column of Figure \ref{fig:exploration_curves}).

We also observe that layer rotation rates (rotation with respect to the previous iterate) translate remarkably well into layer rotations (rotation with respect to the initialization). For example, the $\alpha = 0$ setting used in the fifth column indeed leads all layers to rotate quasi synchronously. As discussed in Section \ref{sec:layca}, this is not self-evident. Understanding why this happens (and why the first and last layers seem to be less tameable) is an interesting direction of research resulting from our work.

\subsection{Layer rotation's relationship with training speed} \label{sec:convergence}
\begin{figure}[!h]
\begin{center}
\includegraphics[width=.95\linewidth]{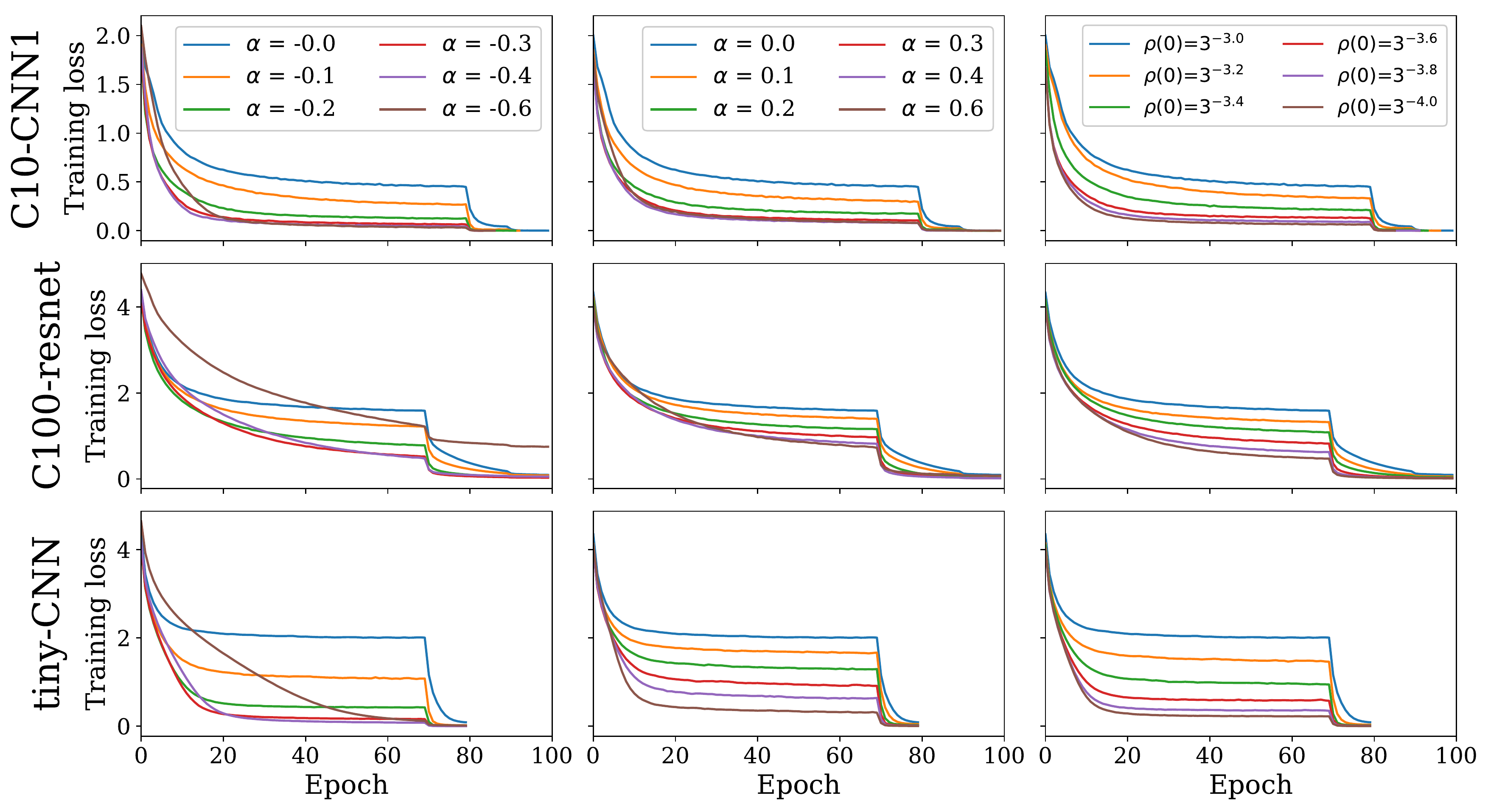} 
\end{center}
\caption{Loss curves obtained for different $\alpha$ and $\rho (0)$ values on the first three tasks of Table \ref{tab:experiments}, using Layca for training. The $\alpha$ and $\rho (0)$ configurations are specified for each column in the associated legend. The visualizations unveil a remarkable phenomenon: the more uniform or the larger the layer rotation rates, the higher the plateaus in which the loss gets stuck into. The sudden drops correspond to a reduction of the global learning rate as specified in \ref{sec:LaycaExperimentalSetup}.}
\label{fig:convergence}
\end{figure}

While generalization is the main focus of our work, we observed through our experiments that layer rotation rates also influenced the training speed of our models in a remarkable way. Figure \ref{fig:convergence} depicts the loss curves obtained for different values of $\alpha$ and $\rho (0)$ on the first three tasks of Table \ref{tab:experiments}. It appears that the larger or the more uniform the layer rotation rates are, the higher the plateaus in which loss curves get stuck into. The plateaus might be due to a form of interference between the different neurons' training processes. The precision with which the height of plateaus can be controlled through the $\alpha$ and $\rho (0)$ parameters is striking and further supports the idea that layer rotation controls a fundamental yet implicit mechanism in deep learning. Following our rule of thumb, this result also suggests that high plateaus are additional indicators of good generalization performance. This is consistent with the systematic occurrence of high plateaus in the loss curves of state of the art networks (e.g., \citet{He2016,Zagoruyko2016}).

\section{A study of layer rotation in standard training settings} \label{sec:CommonPracticesAnalysis}
Section \ref{sec:Exploration} uses Layca to study the relation between layer rotations and generalization or training speed in a controlled setting. This section investigates the layer rotation configurations that naturally emerge when using SGD, weight decay or adaptive gradient methods for training. First of all, these experiments will provide supplementary evidence for the rule of thumb proposed in Section \ref{sec:Exploration}. Second, we'll see that studying training methods from the perspective of layer rotation can provide useful insights to explain their behaviour.

The experiments are performed on the five tasks of Table \ref{tab:experiments}. The learning rate parameter is tuned independently for each training setting through grid search over 10 logarithmically spaced values ($3^{-7},3^{-6},...,3^2$), except for C10-CNN2 and C100-WRN where learning rates are taken from their original implementations when using SGD + weight decay, and from \citep{Wilson2017} when using adaptive gradient methods for training. The test accuracies obtained in standard settings will often be compared to the best results obtained with Layca, which are provided in the 5th column of Figure \ref{fig:exploration_curves}.

\subsection{Analysis of SGD's learning rate} \label{sec:SGDanalysis}
The influence of SGD's learning rate on generalization has been highlighted by several works \citep{Jastrz2017,SmithSam2017,Smith2017,Hoffer2017,masters2018revisiting}. The learning rate parameter directly affects layer rotation rates, since it changes the size of the updates. In this section, we verify if the learning rate's impact on generalization is coherent with our rules of thumb.

\begin{figure}[!h]
\begin{center}
\includegraphics[width=.92\linewidth]{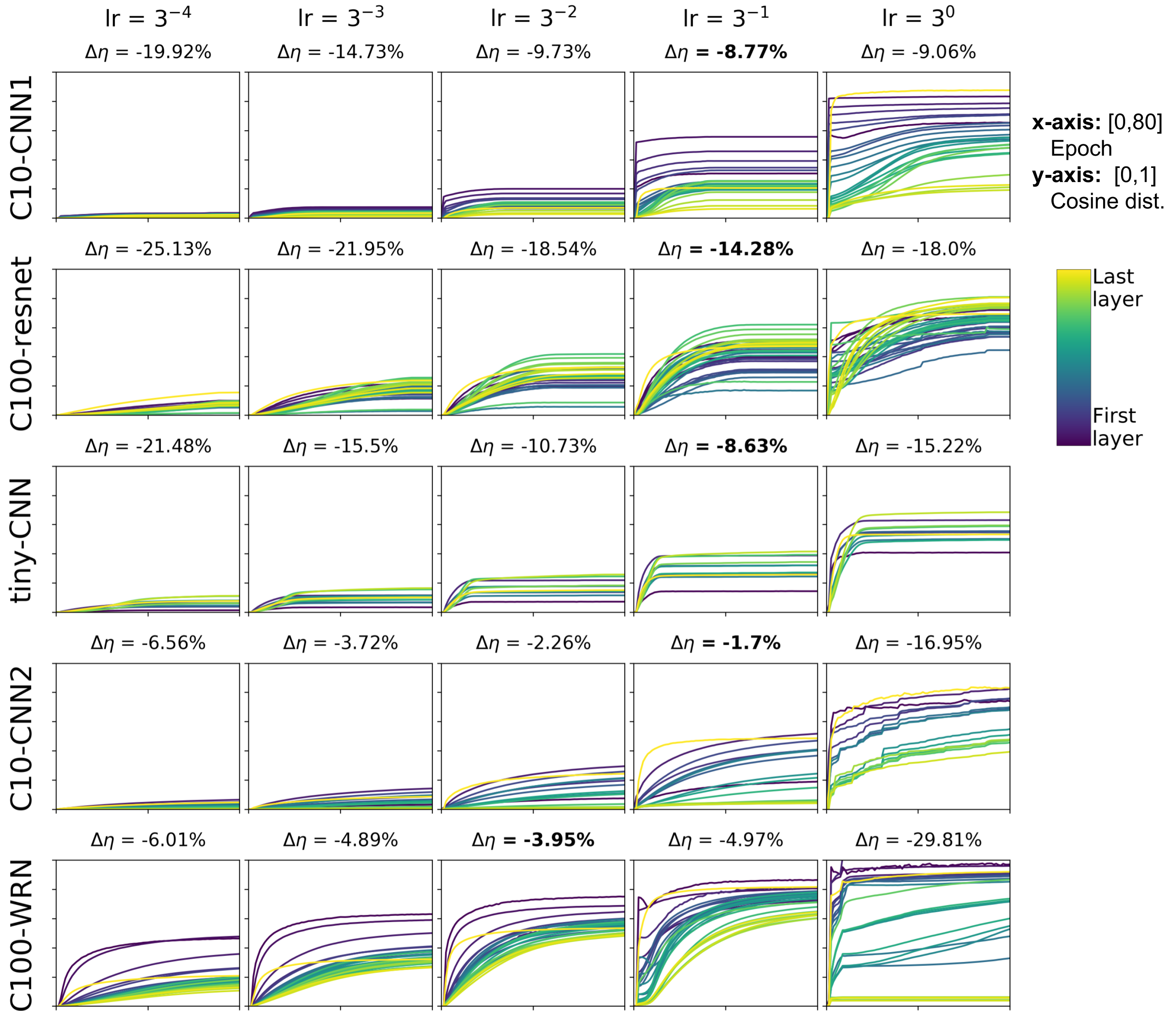}
\end{center}
\caption{Layer rotation curves and the corresponding test accuracies generated by vanilla SGD with different learning rates. Colour code, axes and $\Delta\eta$ computation are the same as in Figure \ref{fig:exploration_curves}. The influence of the learning rate parameter on generalization is consistent with our rule of thumb (larger layer rotations $\rightarrow$ better generalization), except for cases with abrupt layer rotations. We further show in Figure \ref{fig:lr_further_analysis} that these abrupt layer rotations do not translate in improvements of the loss. Moreover, despite extensive learning rate tuning, SGD induces test performances that are significantly below Layca's optimal configuration (cfr. $5^{th}$ column of Figure \ref{fig:exploration_curves}). This is also in accordance with our rules of thumb, since SGD does not seem to be able to generate layer rotations that reach a cosine distance of 1.}
\label{fig:lr_analysis}
\end{figure}

\begin{figure}[!h]
\begin{center}
\includegraphics[width=.95\linewidth]{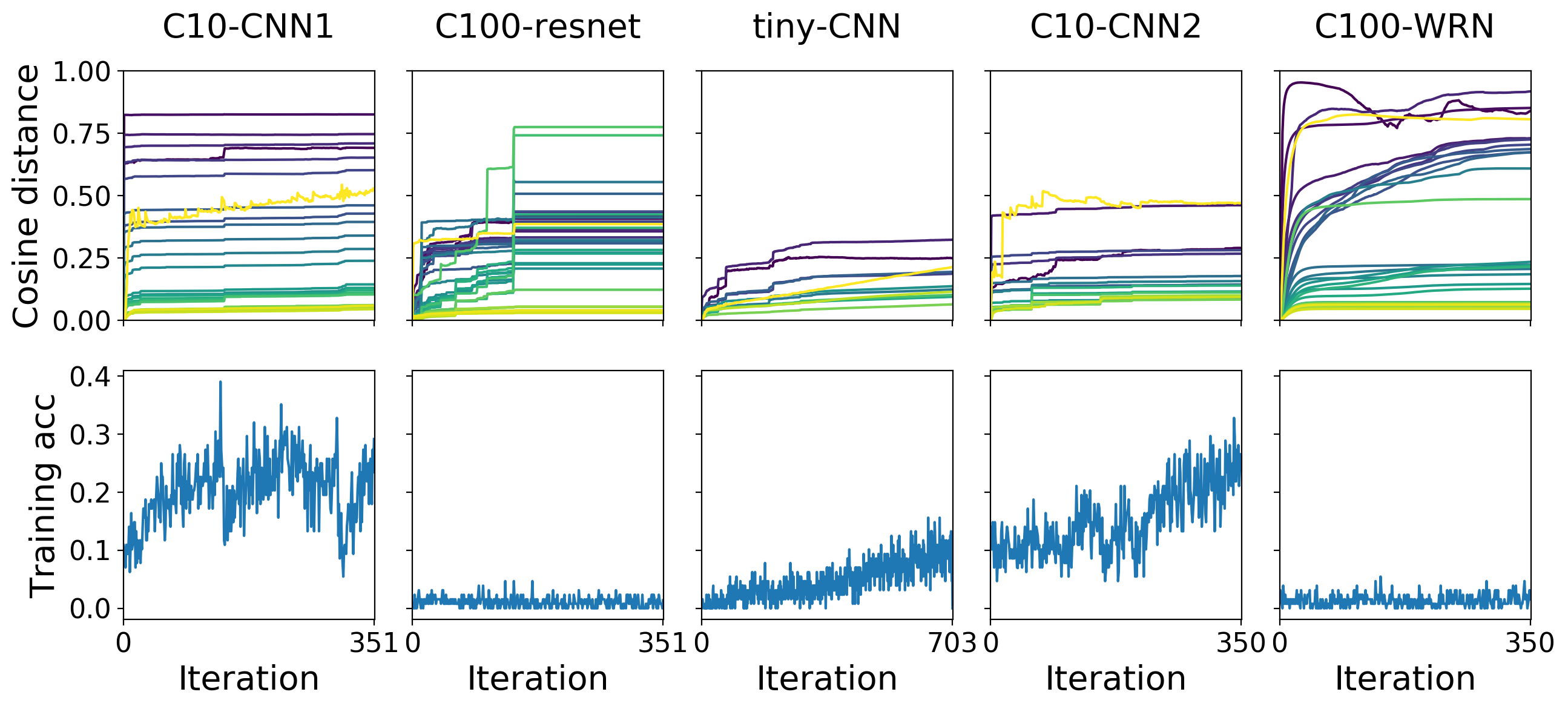}
\end{center}
\caption{Layer rotation and training curves during the first epoch of SGD training with high learning rates (cfr. Figure \ref{fig:lr_analysis}). The visualization reveals large layer rotations that are sometimes performed in a single iteration. Importantly, these iterations do not induce improvements in training accuracy, which probably explains why these configurations escape the scope of our rule of thumb.}
\label{fig:lr_further_analysis}
\end{figure}

Figure \ref{fig:lr_analysis} displays the layer rotation curves and test accuracies generated by different learning rate configurations during vanilla SGD training on the five tasks of table \ref{tab:experiments}. We observe that test accuracy increases for larger layer rotations (consistent with our rule of thumb) until a tipping point where it starts to decrease (inconsistent with our rule of thumb). We show in Figure \ref{fig:lr_further_analysis} that these problematic cases involve extremely abrupt layer rotations that do not translate in improvements of the training loss. These observations thus highlight an important condition for our rules of thumb to hold true: the monitored layer rotations should coincide with actual training (i.e. improvements on the loss function). 

A second interesting observation is that the layer rotation curves obtained with vanilla SGD are far from the ideal scenario disclosed in Section \ref{sec:Exploration}, where the majority of the layers' weights reached a cosine distance of 1 from their initialization. In accordance with our rules of thumb, SGD also reaches considerably lower test performances than Layca. A more extensive tuning of the learning rate (over 10 logarithmically spaced values) did not help SGD to solve its two systematic problems: 1) layer rotations are not uniform and 2) the layers' weights stop rotating before reaching a cosine distance of 1.

\subsection{Analysis of SGD and weight decay} \label{sec:WDanalysis}
\begin{figure}[!h]
\begin{center}
\includegraphics[width=.9\linewidth]{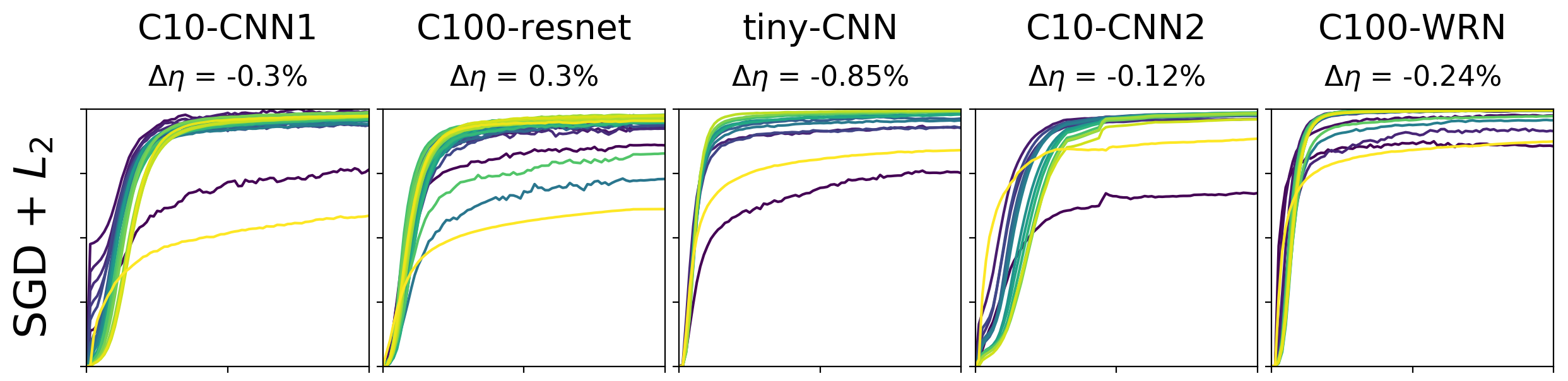}
\end{center}
\caption[Caption for LOF]{Layer rotation curves and the corresponding test accuracies generated by SGD with weight decay. Colour code, axes and $\Delta\eta$ computation are the same as in Figure \ref{fig:exploration_curves}. The application of weight decay during SGD training enables layer rotations that reach a cosine distance of 1 and leads to test performances comparable to Layca's optimal configuration (cfr. $5^{th}$ column of Figure \ref{fig:exploration_curves}). These results thus provide supplementary evidence for our rule of thumb and a new perspective on weight decay regularization.}
\label{fig:SGD_analysis}
\end{figure}

Several papers have recently shown that, in batch normalized networks, the regularization effect of weight decay was caused by an increase of the effective learning rate \citep{VanLaarhoven2017,Hoffer2018,Zhang2019}. More generally, reducing the norm of weights increases the amount of rotation induced by a given training step. It is thus interesting to see how weight decay affects layer rotations, and if its impact on generalization is coherent with our rule of thumb. Figure \ref{fig:SGD_analysis} displays, for the 5 tasks, the layer rotation curves generated by SGD when combined with weight decay (in this case, equivalent to $L_2$-regularization). We observe that weight decay solves SGD's problems ( cfr. Section \ref{sec:SGDanalysis}): all layers' weights reach a cosine distance of 1 from their initialization, and the resulting test performances are on par with the ones obtained with Layca.

This experiment not only provides important supplementary evidence for our rules of thumb, but also novel insights around weight decay's regularization ability in deep nets: weight decay seems to be key for enabling large layer rotations (weights reaching a cosine distance of 1 from their initialization) during SGD training. Since the same behaviour can be achieved with tools that control layer rotation rates (cfr. Layca), without an extra parameter to tune, our results could potentially lead weight decay to disappear from the standard deep learning toolkit. 

\subsection{Analysis of learning rate warmups} \label{sec:WarmUpanalysis}
We've seen in Section \ref{sec:SGDanalysis} that during SGD training, high learning rates could generate abrupt layer rotations at the very beginning of training that do not improve the training loss. In this section, we investigate if these unstable layer rotations could be the reason why learning rate warmups are sometimes necessary when using high learning rates \cite{He2016,Goyal2018}. For this experiment, we use a deeper network that notoriously requires warmups for training: ResNet-110 \cite{He2016}. The network is trained on the CIFAR-10 dataset with standard data augmentation techniques. We use a warmup strategy that starts at a 10 times smaller learning rate and linearly increases to reach the final learning rate in a specified number of epochs.

\begin{figure}[!h]
\begin{center}
\includegraphics[width=.95\linewidth]{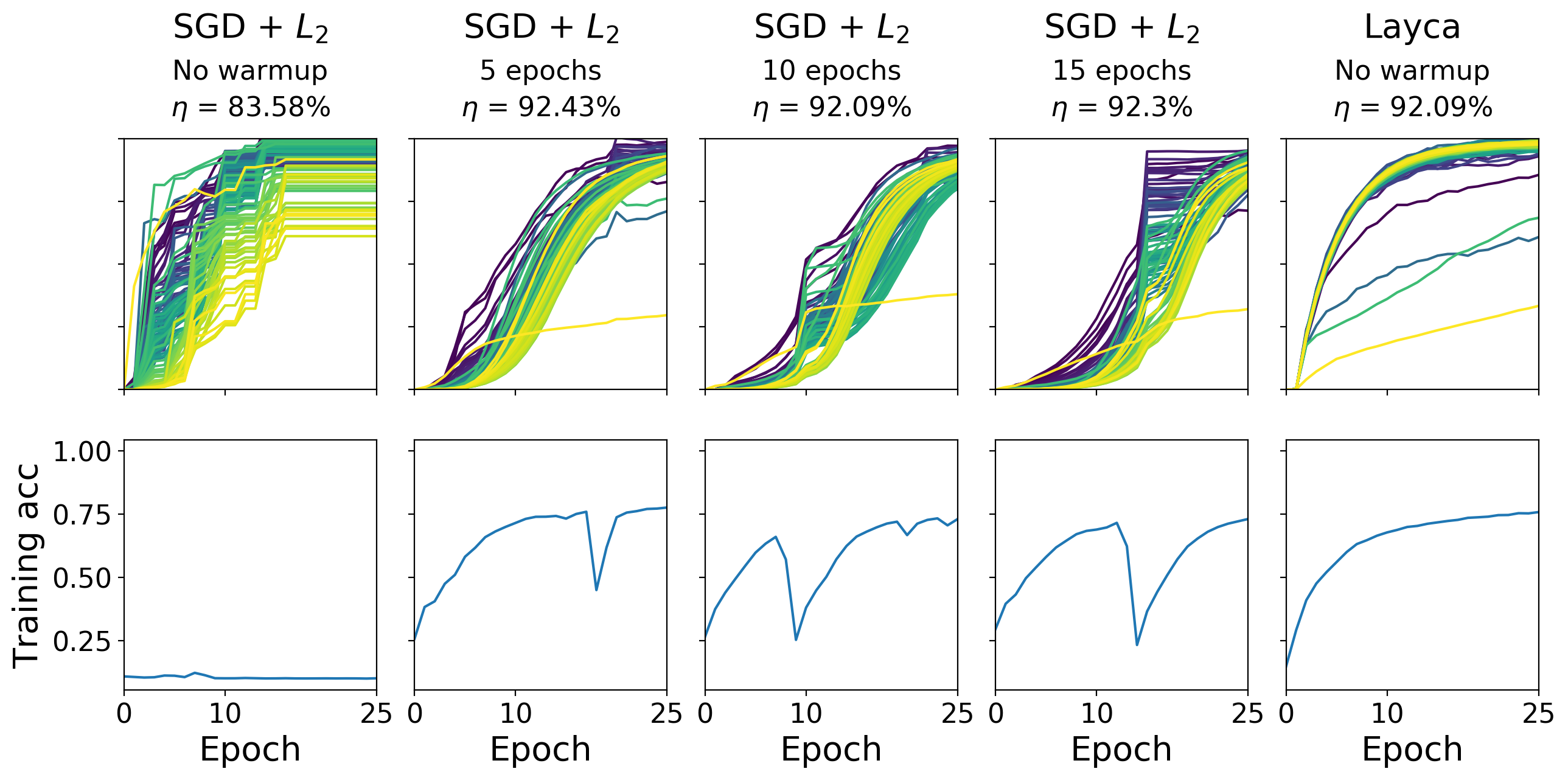}
\end{center}
\caption{Layer rotation and training curves obtained when using different warmup durations (0,5,10 or 15 epochs) during the training of ResNet-110 on CIFAR-10 with high learning rates ($3^{-1}$). The curves are shown for the 25 first epochs only -out of 200. $\eta$ is the final test accuracy. We observe that SGD generates unstable layer rotations that translate into a stagnation or a decrease of the training accuracy. Using warmups drastically reduces these instabilities. Layca doesn't generate instabilities and reaches high generalization performance (and large layer rotations) without the need for warmups.}
\label{fig:warmup_analysis}
\end{figure}

Figure \ref{fig:warmup_analysis} displays the layer rotation and training curves when training with a high learning rate ($3^{-1}$) and different warmup durations (0,5,10 or 15 epochs of warmup). We observe that without warmup, SGD generates unstable layer rotations and training accuracy doesn't improve before the 25th epoch. Using warmups brings significant improvements: a $75\%$ training accuracy is reached after 25 epochs, with only some instabilities in the training curves -that again are synchronized with abrupt layer rotations. Finally, we also use Layca for training (with a $3^{-3}$ learning rate). We observe that it doesn't suffer from SGD's instabilities in terms of layer rotation. Hence, Layca achieves large layer rotations and good generalization performance without the need for warmups.

\subsection{Analysis of adaptive gradient methods} \label{sec:ADGanalysis}
The recent years have seen the rise of adaptive gradient methods in the context of machine learning (\textit{e.g.} RMSProp \citep{Tieleman2012}, Adagrad \citep{Duchi2011}, Adam \citep{Kingma2015}). The key element distinguishing adaptive gradient methods from their non-adaptative equivalents is a parameter-level tuning of the learning rate based on the statistics of each parameter's partial derivative. Initially introduced for improving training speed, \citep{Wilson2017} observed that these methods also had a considerable impact on generalization. Since these methods affect the rate at which individual parameters change, they might also influence the rate at which layers change (and thus layer rotations). 

\begin{figure}[ht]
\begin{center}
\begin{subfigure}[t]{0.7\linewidth}
\includegraphics[width=1.\linewidth]{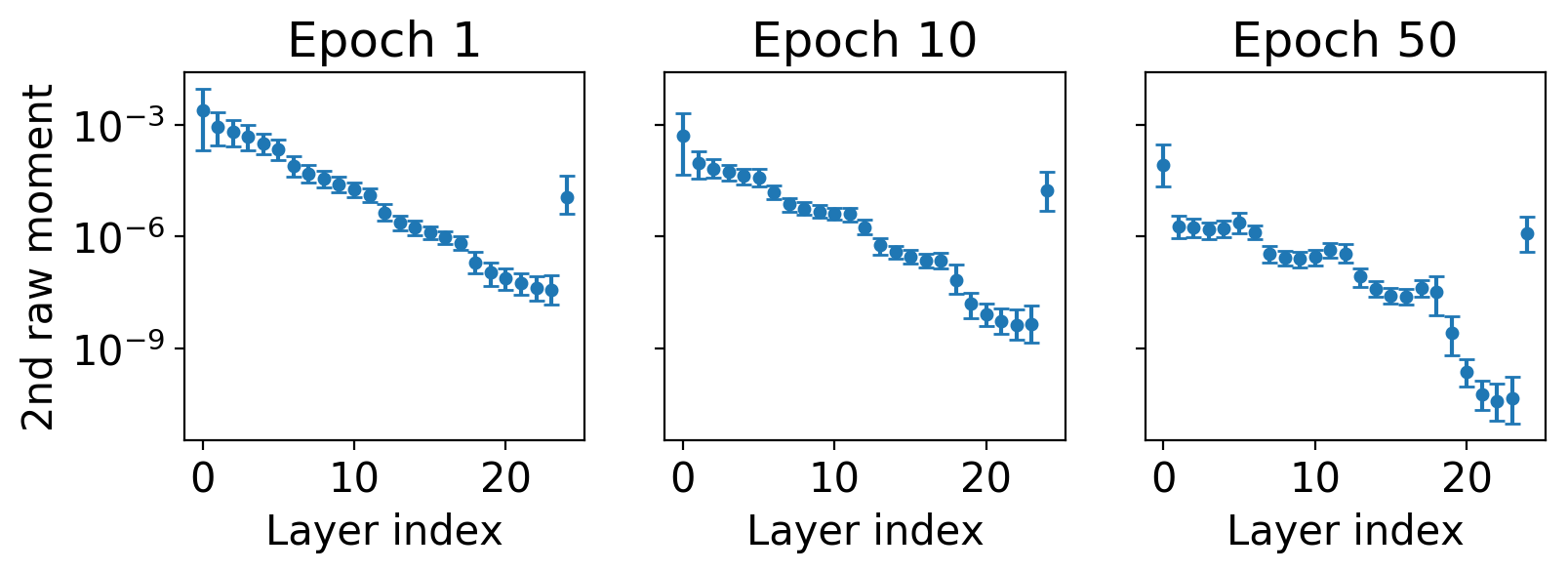}
\end{subfigure}
\end{center}
\caption{Adam's parameter-wise estimates of the second raw moment (uncentered variance) of the gradient during training on C10-CNN1, represented for each layer separately through their $10^{th}$, $50^{th}$ and $90^{th}$ percentiles. The results provide evidence that the parameter-level statistics used by adaptive gradient methods vary mostly between layers and negligibly inside layers.}
\label{fig:adam_analysis}
\end{figure}

We first observe to what extent the parameter-level learning rates of Adam vary across layers. We monitor Adam's estimate of the second raw moment, which is used for determining the parameter-level learning rates, when training on the C10-CNN1 task. The estimate is computed by:
$$v_t = \beta _2\cdot v_{t-1} + (1-\beta _2) \cdot g_t ^2 $$
where $g_t$ and $v_t$ are vectors containing respectively the gradient and the estimates of the second raw moment at training step $t$, and $\beta _2$ is a parameter specifying the decay rate of the moment estimate. While our experiment focuses on Adam, the other adaptive methods (RMSprop, Adagrad) also use statistics of the squared gradients to compute parameter-level learning rates. Figure \ref{fig:adam_analysis} displays the $10^{th}$, $50^{th}$ and $90^{th}$ percentiles of the moment estimations, for each layer separately, as measured at the end of epochs 1, 10 and 50. The conclusion is clear: the estimates vary much more across layers than inside layers. This suggests that adaptive gradient methods might have a drastic impact on layer rotations.

\subsubsection{Adaptive gradient methods can reach SGD's generalization ability with Layca}
\begin{figure}[!h]
\begin{center}
\includegraphics[width=.9\linewidth]{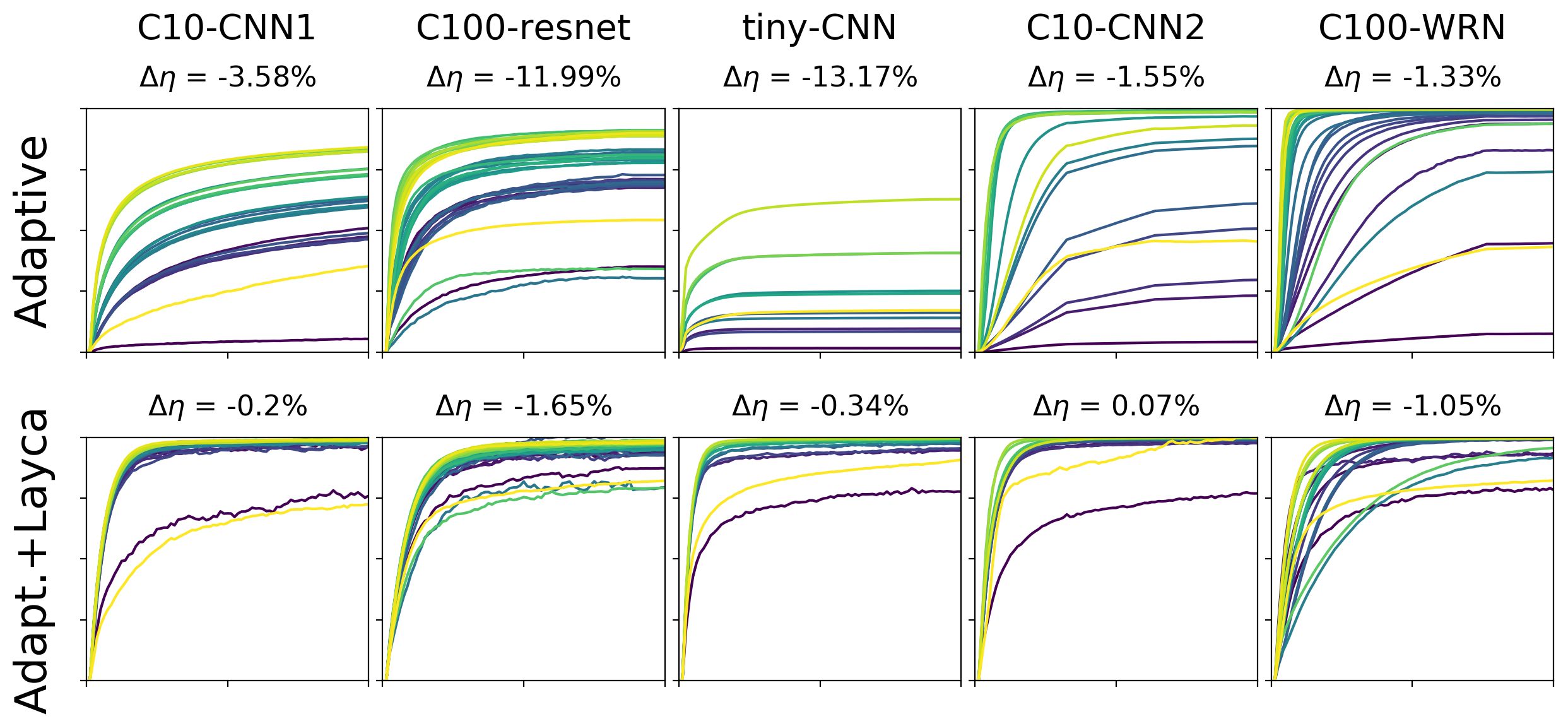} 
\end{center}
\caption{Layer rotation curves and the corresponding test accuracies generated by adaptive gradient methods (RMSProp, Adam, Adagrad, RMSProp+$L_2$ and Adam+$L_2$ respectively for each task/column) without ($1^{st}$ line) and with ($2^{nd}$ line) control of layer rotation with Layca. Colour code, axes and $\Delta\eta$ computation are the same as in Figure \ref{fig:exploration_curves}. In the first line, we observe an overall worse generalization ability compared to Layca's optimal configuration (cfr. $5^{th}$ column of Figure \ref{fig:exploration_curves}) -despite extensive learning rate tuning, together with small and/or non-uniform layer rotations (in accordance with our rule of thumb). When Layca is used on top of adaptive methods to control layer rotation (second line), adaptive methods can reach test accuracies on par with SGD + weight decay.} 
\label{fig:AGM_analysis_curves}
\end{figure}

Since adaptive gradient methods probably affect layer rotations, we will verify if their influence on generalization is coherent with our rule of thumb. Figure \ref{fig:AGM_analysis_curves} ($1^{st}$ line) provides the layer rotation curves and test accuracies obtained when using adaptive gradient methods to train the 5 tasks described in Table \ref{tab:experiments}. We observe an overall worse generalization ability compared to Layca's optimal configuration and small and/or non-uniform layer rotations. We also observe that the layer rotations of adaptive gradient methods are considerably different from the ones induced by SGD (cfr. Figure \ref{fig:lr_analysis}). For example, adaptive gradient methods seem to induce larger rotations of the last layers' weights, while SGD usually favors rotation of the first layers' weights. Could these differences explain the impact of parameter-level adaptivity on generalization? In Figure \ref{fig:AGM_analysis_curves} ($2^{nd}$ line), we show that when Layca is used on top of adaptive methods (to control layer rotation), adaptive methods can reach test accuracies on par with SGD + weight decay. Our observations thus suggest that adaptive gradient methods' poor generalization properties are due to their impact on layer rotations. Moreover, the results again provide supplementary evidence for our rule of thumb.

\subsubsection{SGD can achieve adaptive gradient methods' training speed with Layca}
\begin{figure}[!h]
\begin{center}
\includegraphics[width=.99\linewidth]{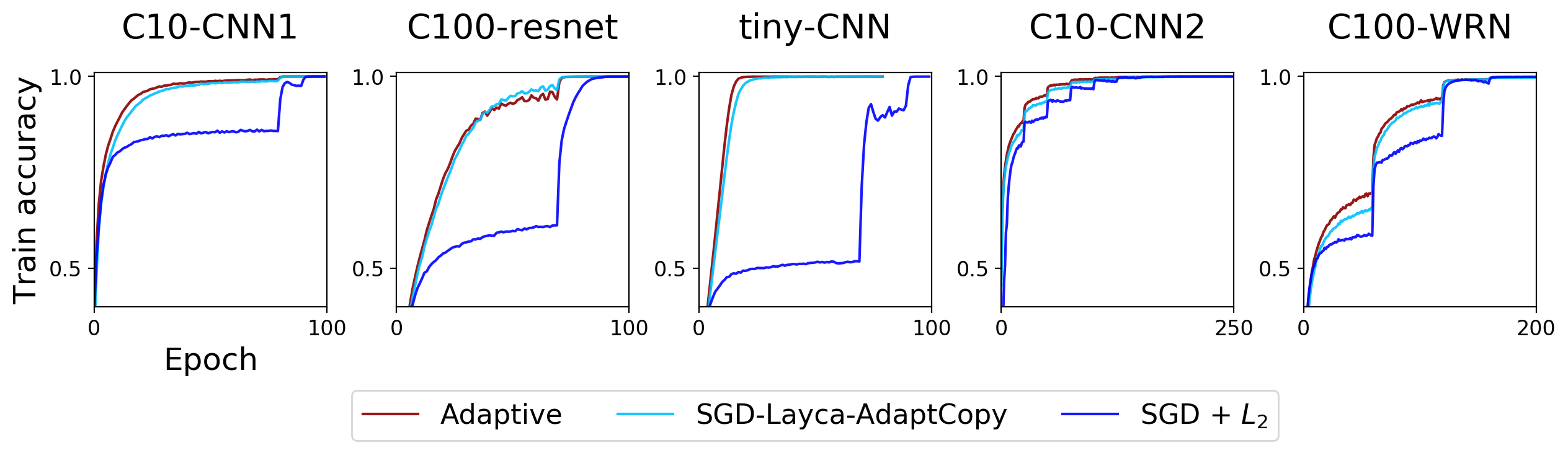} 
\end{center}
\caption[Caption for LOF]{Training curves for the 5 tasks of Table \ref{tab:experiments} with adaptive gradient methods (RMSProp, Adam, Adagrad, RMSProp+$L_2$ and Adam+$L_2$ respectively for each task/column), SGD+weight decay and SGD-Layca-AdaptCopy. During training with SGD-Layca-AdaptCopy, Layca is used to reproduce the layer rotations generated by an adaptive gradient method on the same task\protect\footnotemark. We observe that this training procedure (which doesn't perform parameter-level adaptivity) achieves the same improvements in training speed as adaptive gradient methods.}
\label{fig:AGM_analysis_histories}
\end{figure}

We've seen that the negative impact of adaptive gradient methods on generalization was largely due to their influence on layer rotations. Could layer rotations also explain their positive impact on training speed? To test this hypothesis, we recorded the layer rotation rates emerging from training with adaptive gradient methods, and reproduced them during SGD training with the help of Layca. We then observe if this SGD-Layca optimization procedure (that doesn't perform parameter-level adaptivity) could achieve the improved training speed of adaptive gradient methods. Figure \ref{fig:AGM_analysis_histories} shows the training curves during training of the 5 tasks of Table \ref{tab:experiments} with adaptive gradient methods, SGD+weight decay and SGD-Layca-AdaptCopy (which copies the layer rotation rates of adaptive gradient methods). While adaptive gradient methods train significantly faster than SGD+weight decay, we observe that their training curves are nearly indistinguishable from SGD-Layca-AdaptCopy. Our study thus suggests that adaptive gradient methods impact on both generalization and training speed is due to their influence on layer rotations. 

\footnotetext{When copying the layer rotations of Adam with SGD-Layca-AdaptCopy we also integrate Adam's momentum scheme.}

\section{Related work}
The idea that neurons from different layers potentially train at different rates was motivated by the works on vanishing and exploding gradients \citep{Bengio1994,Hochreiter1998,Glorot2010}. These pioneering works revealed that the norm of gradients is affected by its propagation through the layers, potentially leading to training difficulties. Based on this observation, several other works also designed and studied tools for controlling deep neural network training on a per-layer basis \citep{Yu2017,Ginsburg2018,Bernstein2020}. While these don't conduct a systematic study of layer-level training's relationship with generalization or training speed, they show that layer-level control leads to more stable training, reduced hyperparameter tuning and ultimately better generalization performance. \textit{Interestingly, \citep{Liu2021} recently showed that controlling rotation of weight vectors at the \textit{neuron}-level could improve a network's performance even further.} This indicates that training differences can also occur amongst the neurons of a layer, and further supports our initial hypothesis which states that emergent neuron-level training processes play a crucial role in deep neural networks.

\section{On the interpretation of layer rotations} \label{sec:future}
The previous sections of this chapter demonstrate the remarkable consistency and explanatory power of layer rotation's relationship with generalization. This suggests that layer rotation controls fundamental aspects of deep neural network training. Whether these aspects relate to the clustering abilities and mechanisms studied in Chapters \ref{chap:implicitAbility} and \ref{chap:implicitMechanism} relies on the hypothesis that layer rotation captures the extent by which hidden neurons have been able to ``train" during the network's training process. In this section, we provide an additional experiment to support the link between all these concepts.

We use a toy experiment to visualize how layer rotation affects the features learned by hidden neurons. We train a single hidden layer MLP (with 784 hidden neurons) on a reduced MNIST dataset (1000 samples per class, to increase overparameterization). This toy network has the advantage of having intermediate features that are easily visualized: the weights associated to hidden neurons live in the same space as the input images. Starting from an identical initialization, we train the network with four different learning rates using Layca, leading to four different layer rotation configurations that all reach $100\%$ training accuracy but different generalization abilities (in accordance with our rule of thumb). 

\begin{figure}[!h]
\begin{center}
\includegraphics[width=1.\linewidth]{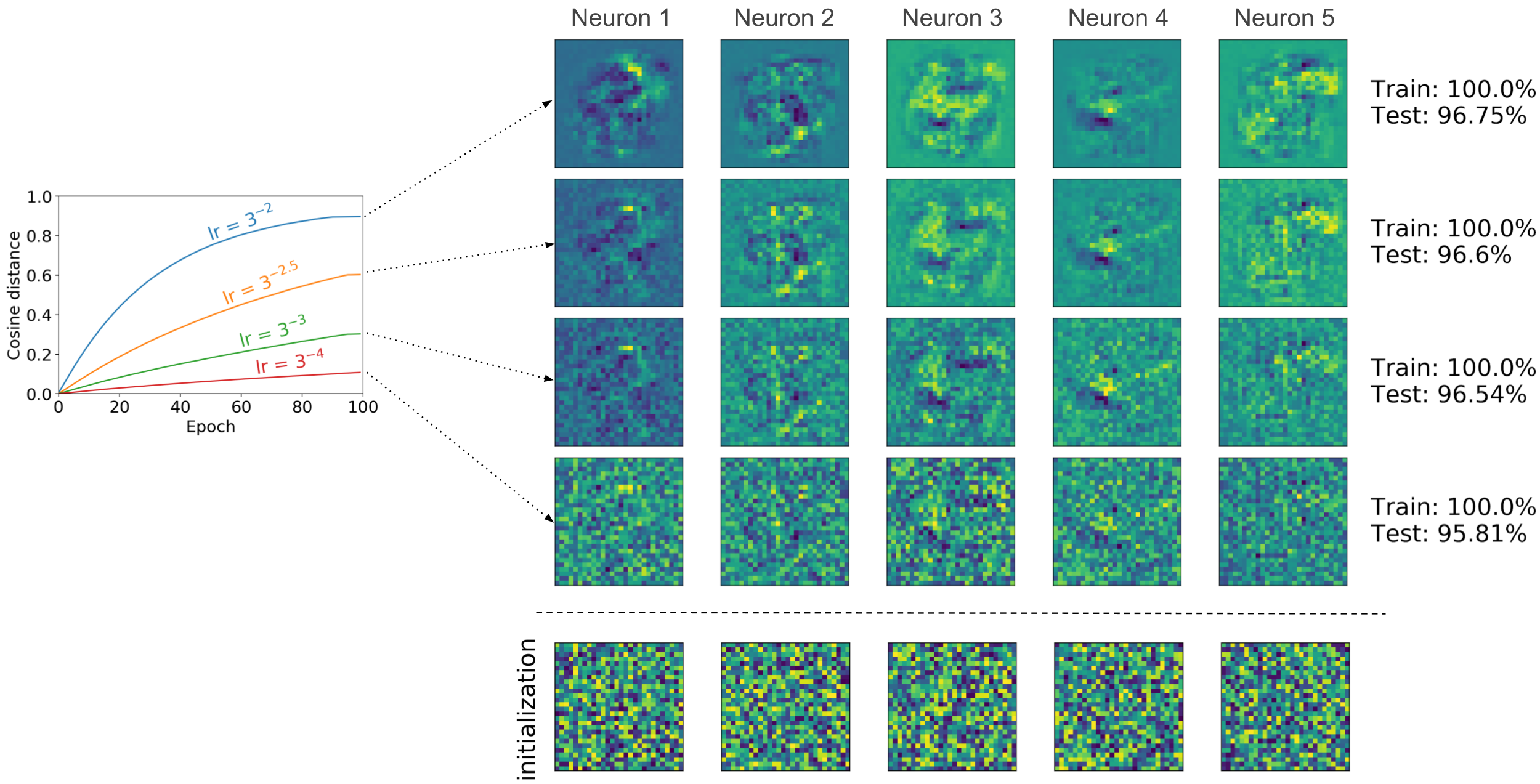} 
\end{center}
\caption{A single hidden layer MLP is trained on a reduced MNIST dataset (1000 examples per class). Starting from an identical initialization, the network is trained with four different learning rates using Layca, leading to four different layer rotation configurations that all reach $100\%$ training accuracy but different generalization abilities (in accordance with our rule of thumb). The learned intermediate features (associated to 5 randomly selected neurons) are visualized for the different layer rotation configurations. The results suggest that layer rotation does not affect which features are learned, but rather to what extent they are learned during the training process.}
\label{fig:feature_quality}
\end{figure}

Figure \ref{fig:feature_quality} displays the features obtained by the different layer rotation configurations (for 5 randomly selected hidden neurons). This visualization unveils an important phenomenon: \textbf{layer rotation does not seem to affect \textit{which} features are learned, but rather \textit{to what extent} they are learned during the training process.} The larger the layer rotation, the more prominent the features -and the less retrievable the initialization. Ultimately, for a layer rotation close to 1, the final weights of the network got rid of all remnants of the initialization. This experiment thus supports our initial hypothesis: fully accomplishing emergent neuron-level training processes is not necessary for reaching $100\%$ training accuracy, but doing so anyway leads to better clustering and generalization abilities. 

\chapter{Discussion and perspectives} \label{chap:discussion}
Our work argues that the natural clustering prior plays a key role in deep learning and generalization. Empirical evidence supporting this hypothesis has been presented in Chapters \ref{chap:implicitAbility},\ref{chap:implicitMechanism} and \ref{chap:implicitHyperparam}. This chapter takes a step back and attempts to see the bigger picture behind our work. More precisely, we (i) discuss to what extent this work effectively validates our hypothesis (Section \ref{sec:towardsTheory}), (ii) predict a rebirth of clustering algorithms for training deep neural networks (Section \ref{sec:rebirthClustering}) and (iii) discuss the societal impact of deep learning research (Section \ref{sec:social}).

\section{Towards validating our hypothesis} \label{sec:towardsTheory}
In order to validate our hypothesis, this work provides a collection of intuitions and experiments. In this section, we detail how future work could improve the generality of our experiments, provide complementary empirical evidence for our hypothesis and incorporate mathematical formalisms.

\subsubsection{Generalizing our experiments}
We ask machine learning models to be able to generalize to a large variety of contexts. So should the empirical claims that support a theory. The experiments presented in this work involve a restricted set of deep learning setups. Using other datasets, architectures, training algorithms and hyperparameters could lead to different or contradicting conclusions. This limitation is particularly strong in Chapter \ref{chap:implicitMechanism}, which studies implicit clustering mechanisms in a setup involving synthetic data and simple neural network architectures. Additionally, the study of large-scale datasets (e.g. ImageNet \citep{Deng2009}) and different modalities (natural language and sounds) would also greatly improve our work.

\subsubsection{Providing complementary empirical evidence}
Identifying and studying implicit phenomena is a difficult endeavour. Conducting many complementary experimental studies are key to (i) offer different perspectives that will help better characterize the phenomenon and (ii) mitigate the risk of misinterpretation due to unrelated confounding factors.

In order to complement our work, a first path worth investigating is the study of densely annotated datasets such as Broden introduced by \citet{Bau}. Our work makes already use of datasets with two levels of class labels for studying implicit clustering. We interpret the subclasses of such datasets as single clusters, but these could in fact be composed of multiple clusters themselves. Moreover, class labels are attributed to the whole image, while the associated objects/concepts are relevant to only a part of it. Datasets like Broden mitigate these two drawbacks by providing many levels of class labels as well as pixel-wise annotations. This holds the potential for improved measures of clustering abilities and more precise studies of clustering mechanisms in natural image datasets.

Another path of investigation consists in further evaluating the explanatory power of our hypothesis. How well does it explain a variety of phenomena? Our work demonstrates the potential of implicit clustering to explain the benefits of pre-training, the coherent gradients hypothesis, neuron interpretability, the benefits of large learning rates, weight decay and others. However, all these explanations remain partial, and many other phenomena are not addressed (e.g. ``double descent" curves \citep{Belkin2019}, the lottery ticket hypothesis \citep{Frankle2018} or the benefits of skip connections \citep{Balduzzi2015}). Conducting in depth studies of all these phenomena under the light of implicit clustering could further evaluate our hypothesis.

\subsubsection{Formalizing intuitions}
The amount of equations and mathematical symbols is remarkably low in this thesis. Our work is indeed mainly composed of informal arguments and intuitions. Being able to translate our claims into mathematical language would help increase their precision and falsifiability. Moreover, mathematics greatly facilitate the exploration of ideas through deductive reasoning \citep{Wigner1960}. Formalizing the intuitions and results presented in Chapter \ref{chap:implicitMechanism} might be a good place to start, given the simplicity of the associated experimental setup. Additionally, \citet{Berner2021} provides a nice overview of several mathematical studies of deep learning, which could provide inspiration for this difficult endeavour.

\section{A rebirth of clustering algorithms} \label{sec:rebirthClustering}
Backpropagation, the algorithm behind SGD, has been the most popular algorithm for training deep neural networks for at least two decades. And for a good reason: it outperforms all other alternatives by a large margin on standard datasets. And yet, there is still a long way to go to reach human-level generalization abilities (cfr. Section \ref{sec:poorGeneralization}). This begs the question: should we continue building on backpropagation and SGD, or explore novel algorithms? If natural data exhibit a clustered structure, as the natural clustering prior states, aren't clustering algorithms a more natural choice for training deep neural networks? \citet{Coates2012} trains deep neural networks with K-means clustering, but their method did not match SGD's performance. Isn't this contradictory with our hypothesis?

While they clearly lack SGD's capabilities on standard classification datasets, clustering algorithms have achieved some successes on natural image-related tasks in the past \citep{Zoran2011,Coates2011a}. Moreover, identifying the right priors appears to be especially crucial for clustering algorithms \citep{Estivill-Castro2002}. We believe that a better characterization of the natural clustering prior in terms of the shape of clusters, their relative density, the distance between them or their relationship with class labels could lead to improved clustering algorithms. For example, many clustering algorithms assume that all clusters contain approximately the same amount of training examples. However, recent works suggest that the ability of deep neural networks to ``memorize" atypical and poorly represented sub-populations is key for their performance on natural image classification tasks \citep{Feldman2020,Jiang2021}. Additionally, many clustering algorithms assume spherical cluster shapes while many aspects of natural images vary in specific, anisotropic ways (e.g. scaling, rotation, translation of objects and object parts).

Hence, the current supremacy of SGD over clustering algorithms might not be the final picture. Because natural clustering-related priors would be more easily integrated into their design, our work predicts a rebirth of clustering algorithms in the coming years, bringing us one step closer to human-level generalization abilities.

\section{The societal impact of deep learning research} \label{sec:social}
Human societies are complex systems. Understanding how they will react to new technologies is a daunting task. Yet, the well-being of billions of individuals can be at stake. The growing integration of deep learning technologies in the industry raises these difficult questions. Because these were an integral part of my own research experience, and encouraged by the NeurIPS conference's call for a ``Broader impact" section\footnote{NeurIPS' call for a ``Broader impact" section: \textit{In order to provide a balanced perspective, authors are required to include a statement of the potential broader impact of their work, including its ethical aspects and future societal consequences. Authors should take care to discuss both positive and negative outcomes.}}, I shortly discuss my personal take on the societal impact of deep learning research.

The ethical concerns around deep learning are numerous. Deep learning technology can be used for autonomous military drones \citep{ScottShane2018}, generating misinformation automatically \citep{AlecRadford2019}, racial discrimination \citep{Kickuchiyo2019}, mass manipulation on social networks \citep{Coombe2020} and many others. Moreover, deep learning confers power to the data owners. The centralization of data in big tech companies leads to rising social, economic and political inequalities \citep{YuvalNoahHarari2018}. Of course, many positive opportunities also emerge from deep learning such as healthcare applications \citep{Panesar2019}, scientific progress \citep{Jumper2021a} and tackling climate change \citep{Rolnick2019}). But from my very limited perspective, I feel that the negative outcomes largely outnumber the positive ones in terms of practical impact.

As researchers and engineers, it might be appealing to leave these ethical considerations to political institutions. They possess the power to regulate technologies, and are expected to preserve common good. But reality doesn't necessarily match expectations. The technological progress could be too fast for ankylosed political systems. Big tech companies could heavily limit a government's room for manoeuvre when profit is at stake. As Professor Harari claims, many political decisions could in fact be in the hands of the scientists and engineers that develop today's technologies \citep{Harari2016}. 

What are scientists and engineers expected to do then? This remains an open question. Several brave individuals decide to quit the field entirely (e.g., \citet{Amadeo2018,Yuan}). Conferences organize workshops around the topic (e.g., \citet{Chowdhury2021,Li2021}). Partnerships with civil society and media organizations are built (e.g. Partnership On AI\footnote{\url{https://partnershiponai.org/}}). In my humble opinion, throwing deep learning technology into a competitive ecosystem is bound to raise inequalities and harm common good. Hence, I believe that designing tools to facilitate cooperation of citizens around shared goals, at small and large scale, has a big potential for positive change. These tools could for example take the form of web applications that facilitate knowledge sharing and collective decision making or that connect people with similar goals. I believe there is a lot of room for improvement around collaborative software design, and that this endeavour constitutes a promising path towards a society where deep learning researchers can pursue their quest with a peace of mind.

\cleardoublepage  
\phantomsection 
\renewcommand{\thechapter}{$\star$}
\renewcommand{\thesection}{$\star$}
\addcontentsline{toc}{chapter}{\protect\numberline{}Conclusion} 
\chapter*{Conclusion}
Deep learning gained a lot of popularity for the plethora of applications it enables. But the open questions and mysteries behind these successes are at least as fascinating. They challenge our understanding of generalization, which constitutes the most fundamental aspect of machine learning. Moreover, they exhibit connections with the human brain, one of the universe's greatest mysteries. In this thesis, we propose a novel path towards a better understanding of deep learning.

Our work builds on the idea that generalization is strongly influenced by the priors integrated into a learning system's design. We propose a natural clustering prior for supervised image classification problems, and study (i) to what extent this prior is integrated into deep learning systems and (ii) if its integration influences generalization. 

We provide a collection of experiments supporting the occurrence of an implicit clustering ability, mechanism and hyperparameter in deep learning. Moreover, we demonstrate empirically that these components consistently influence the generalization abilities of deep neural networks. We further highlight many connections between our observations and previous work on neuron interpretability, the early phase of training, pre-training, the coherent gradients hypothesis and others.

Overall, our work reveals that the natural clustering prior offers a promising path towards understanding the generalization abilities of deep learning systems. Additionally, it unveils a path towards new clustering-based training algorithms that could push the generalization abilities of such learning systems even further. With these exciting perspectives in mind, we look forward to the future developments of the fascinating field of deep learning. 

\cleardoublepage  
\phantomsection 
\renewcommand{\thechapter}{$\star$}
\renewcommand{\thesection}{$\star$}
\addcontentsline{toc}{chapter}{\protect\numberline{}\bibname} 
\bibliography{chapters/references}

\cleardoublepage  
\phantomsection 
\renewcommand{\thechapter}{$\star$}
\renewcommand{\thesection}{$\star$}
\addcontentsline{toc}{chapter}{\protect\numberline{}Publications} 
\chapter*{Publications}
This thesis wraps up a series of works that have been previously published in several conference venues:
\begin{itemize}
\item Simon Carbonnelle, C. De Vleeschouwer. Intraclass clustering: an implicit learning ability that regularizes DNNs, \textit{ICLR} 2021.
\item Simon Carbonnelle, C. De Vleeschouwer. Experimental study of the neuron-level mechanisms emerging from backpropagation, \textit{ESANN} 2019.
\item Simon Carbonnelle, C. De Vleeschouwer. Layer rotation: a surprisingly simple indicator of generalization in deep networks, \textit{Workshop Deep Phenomena, ICML} 2019.
\end{itemize}

My journey as a PhD student started with a collaboration with Claire Gosse and Marie Van Reybroeck, two researchers in speech and language pathology. This collaboration initiated an ongoing research project that studies the interactions between handwriting, spelling and dyslexia through the use of digital tablets and signal processing tools. My work in this context also contributed to the following two publications:
\begin{itemize}
\item C. Gosse, S. Carbonnelle, C. De Vleeschouwer, M. Van Reybroeck. Specifying the graphic characteristics of words that influence children's handwriting, \textit{Reading and Writing}, 31 (5), 1181-1207, 2018.
\item C. Gosse, S. Carbonnelle, C. De Vleeschouwer, M. Van Reybroeck. The influence of graphic complexities of words on the handwriting of children of 2nd grade, \textit{SIG WRITING, 16th international conference of the EARLI special interest group on writing}, Liverpool, 2016.
\end{itemize}

%

\end{document}